\documentclass[10pt]{article} 
\usepackage[preprint]{tmlr}

\usepackage{amsmath,amsfonts,bm}

\def\eqref#1{equation~\ref{#1}}

\def\1{\bm{1}}

\DeclareMathAlphabet{\mathsfit}{\encodingdefault}{\sfdefault}{m}{sl}
\SetMathAlphabet{\mathsfit}{bold}{\encodingdefault}{\sfdefault}{bx}{n}

\usepackage{hyperref}
\usepackage{url}

\def\openreview{\url{https://openreview.net/forum?id=XXXX}} 

\usepackage{helpers/style}
\title{\papertitle}

\begin{document}

\maketitle

\begin{abstract}
\acrfull{tsad} is a fundamentally challenging task: it is unsupervised, requires high local precision capabilities, and suffers from limited labelled data.
Recent work has explored adapting \acrfullpl{llm} to time-series via cross-modal transfer.
However, existing approaches are poorly suited to Multivariate TSAD (\acrshort{mtsad}), as tokenisation often obscures fine-grained temporal and variables structures, and most methods optimise auxiliary objectives such as forecasting or reconstruction.
We posit that these surrogate objectives become unreliable as model capacity increases.
Highly expressive models risk reconstructing or forecasting anomalies, undermining the foundations of both strategies.
Instead, we propose to use \acrshortpl{llm} purely as feature extractors and directly analyse their induced feature spaces.
In this work, investigate the ability of \acrshortpl{llm} to provide useful representations for \acrshortpl{mtsad} in a zero-shot setting through classification.
To this end, we study the separability of the extracted features, under the assumption that anomalous features should be isolated from normal ones.
We analyse their temporal and contextual discriminative ability using simple step-wise tokenisation schemes, with one time-step per token, which preserve fine-grained local information.
Their usability for \acrshort{mtsad} is assessed using an unsupervised Isolation Forest, enabling a direct evaluation of the resulting representation quality.
We further investigate whether the observed performance can be attributed to the language-related pre-training of the \acrshortpl{llm} or arises independently of it.
Our results provide empirical insight into the ability of \acrshortpl{llm} to represent multivariate time-series for anomaly detection and highlight broader applications of cross-modal foundation models.
\end{abstract}

\section{Introduction}
\label{sec:intro}

\makeatletter
\let\thefootnote\relax\footnotetext{\textit{The code will be made publicly available.}}
\makeatother

Time-series analysis is inherently challenging due to the high heterogeneity and domain dependence of available datasets, compounded by their limited sizes.
These constraints hinder the development of general-purpose \glspl{tsfm} capable of addressing core tasks such as classification, forecasting, and anomaly detection.

Recent work has explored the use of \glspl{llm} for time-series processing~\citep{Zhang2024LargeLM,Jiang2024EmpoweringTS}.
Trained on large-scale textual corpora, \glspl{llm} possess strong sequential modelling capabilities and have demonstrated notable cross-modal adaptability~\citep{Lu2022FrozenPT}.
Leveraging pre-trained \glspl{llm} as processors for time-series data is therefore an appealing direction, benefiting from the maturity and scale of language modelling research.

A central challenge lies in the mismatch between time-series data and textual representations.
Unlike natural language, time-series consist of continuous numerical values with precise temporal and inter-variable relationships.
Prior work has identified several directions to address this mismatch, including learned projections, model adaptation, and specialised tokenisers~\citep{Spathis2023TheFS}; however, only a few strategies allow the use of \glspl{llm} for time-series analysis in a zero-shot manner.
The popular direct tokenisation of \citet{Gruver2023LargeLM} converts numbers to text characters but compromises numerical precision and can fragment values across multiple tokens, disrupting temporal dependencies.
These issues are exacerbated in multivariate settings, where serialisation destroys the multivariate structure, hinders dependencies between variables, and drastically inflates token count~\citep{Chatzigeorgakidis2024MultiCastZM}.

A deeper issue lies in how existing approaches prioritise general representation learning, overlooking the specific demands of \gls{tsad}.
In particular, \gls{tsad} poses additional difficulties, as it requires high local sensitivity and is typically formulated in an unsupervised setting due to the absence of anomalies in the training data~\citep{Darban2022DeepLF,Mejri2024UnsupervisedAD}.
Common processing strategies, particularly tokenisation schemes such as patching or quantisation, can obscure fine-grained temporal patterns and attenuate anomalies.
Furthermore, most methods rely on auxiliary objectives, notably forecasting or reconstruction;
while effective for broad evaluation, these objectives are not aligned with anomaly detection: sufficiently expressive models risk reconstructing or forecasting anomalous patterns as faithfully as normal data, thereby suppressing residual scores and leaving anomalies undetected by design~\citep{Gong2019MemorizingNT,Yoon2025MOMEMTOPM}.
Consequently, higher model capacity and reconstruction fidelity do not guarantee superior anomaly detection; on the contrary, excessive reconstruction capability can actively degrade performance by absorbing anomalous signals.

To circumvent these limitations, we consider anomaly detection through the lens of \gls{occ}~\citep{Scholkopf2001EstimatingTS}, where normal patterns define the data distribution and anomalies correspond to deviations from it.
This perspective naturally favours approaches that operate directly on feature representations to estimate abnormality~\citep{Hermary2026ASTERLP}.
In this context, we investigate features extracted from pre-trained foundation models such as \glspl{llm}:
as these models continue to scale, assessing their ability to transfer to new modalities in a zero-shot setting becomes increasingly relevant.

In this direction, we propose to evaluate the extent to which \gls{llm}-induced representations isolate anomalous from normal patterns.
We therefore investigate simple, zero-shot-compatible tokenisation strategies, with an emphasis on step-wise tokenisation where each time-step is mapped to a single token.
This design limits the effect of local information erasure, making it particularly suitable for detecting point anomalies while not compromising the detection of other anomaly types,
and allowing for a focus on the temporal processing of \glspl{llm}.

Our contributions are summarised as follows:
\begin{itemize}
    \item We discuss the limitations of current foundation model approaches applied to \gls{mtsad} and propose investigating language model features as an alternative approach, treating anomaly detection as a central objective.
    \item We propose a unified evaluation protocol designed for imbalanced, unsupervised anomaly detection settings, based on an isolation prior for anomalies.
    \item We provide an extensive empirical study across multiple \gls{llm} architectures, tokenisation methods, datasets and context sizes, offering insight into the zero-shot applicability of \glspl{llm} for \gls{mtsad}. We further investigate whether language knowledge is leveraged through trivial tokenisation strategies.
\end{itemize}

\section{Related Work}
\label{sec:related}

The emergence of \glspl{tsfm} marks a significant shift from task-specific architectures toward unified frameworks capable of addressing core evaluation tasks such as classification, short- and long-term forecasting, and anomaly detection~\citep{Liang2024FoundationMF,Meyer2024BenchmarkingTS}.
Recent advancements, exemplified by UniTS~\citep{Gao2024UniTSAU}, demonstrate the feasibility of training models across these diverse tasks simultaneously, allowing for fine-tuning on anomaly detection datasets or direct zero-shot application to novel tasks.
However, many current \glspl{tsfm} remain constrained by a focus on proxy objectives, such as reconstruction or forecasting which are typically limited approaches in anomaly detection~\citep{Hermary2025RemovingGB}.

Parallel to this, the sequential nature of time-series data has led to the adoption of \glspl{llm} as powerful backbones for temporal analysis~\citep{Zhang2024LargeLM}.
This direction is driven by the inherent multi-modal capabilities of \glspl{llm} and their proficiency in processing sequential dependencies~\citep{Jiang2024EmpoweringTS}.
Most research in this area focuses on forecasting-based objectives or fine-tuning strategies designed to bridge the modality gap~\citep{Bian2024MultiPatchPA,Hu2025ContextAlignmentAA,Sun2023TESTTP,Masserano2024EnhancingFM}.

In the sub-field of time-series anomaly detection, \glspl{llm} are increasingly utilised to provide detection capabilities, but also natural language explainability.
Architectures like TriP-LLM~\citep{Yu2025TriPLLMAT} and MADLLM~\citep{Tao2025MADLLMMA} train specialised adapters to bridge the modality and domain gaps, while AnomalyLLM~\citep{Liu2024LargeLMKD} explores knowledge distillation from LLMs.

Despite these advancements, the direct zero-shot application of \glspl{llm} for \gls{tsad} remains an emergent field, characterised by a wide array of diverse strategies.
While some approaches rely on powerful closed-source models and elaborate prompting~\citep{Xu2025CanML,Dong2024CanLS,Liu2026ASO}, others, like LLMAD~\citep{Liu2024LargeLM}, integrate domain knowledge and databases, converting time-series into tabular or text formats for few-shot explanations.
Additional attempts have been made to obtain anomaly positions directly from the \gls{llm} via prompting~\citep{RussellGilbert2024AADLLMAA}.
However, SigLLM~\citep{Alnegheimish2024CanLL} demonstrates that using \glspl{llm} in a forecasting-based strategy yields far better results than direct questioning.

Alternatively, research in image-based modalities leverages the maturity of \glspl{vlm} by transforming time-series into plots or frequency maps~\citep{He2025HarnessingVM,Zhuang2024SeeIT}.
While these visual foundations offer an alternative for cross-modal analysis, this work focuses on the efficiency and granularity of direct time-series-to-token conversions for multivariate time-series, in zero-shot setting.

\begin{figure*}
    \centering
    \resizebox{1\linewidth}{!}{\input{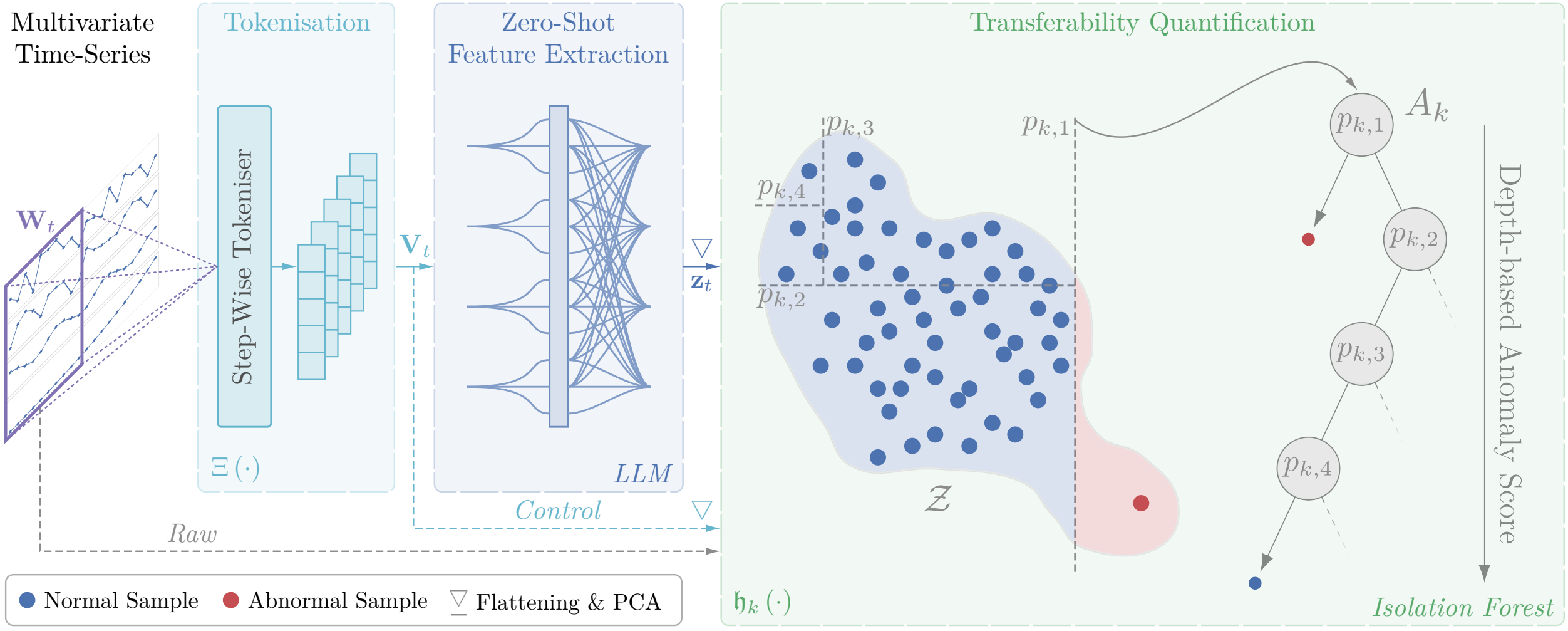}}
    \caption{
We adopt a windowed approach, assigning one anomaly score per window.
Each window is tokenised step-wise (one token per time-step), processed by a pre-trained \gls{llm}, and aggregated into a fixed-size feature vector.
An Isolation Forest is then fitted on these features to produce anomaly scores.
The resulting performance reflects the suitability of the feature space for anomaly detection and, consequently, the representations transferability to \gls{mtsad}.
The \textit{Raw} path corresponds to applying the Isolation Forest directly on the dataset; the \textit{Control} one on the tokenised dataset.
}
    \label{fig:pipeline}
\end{figure*}

\section{Methodology}
\label{sec:method}

We formalise the \gls{tsad} problem in Section~\ref{subsec:formul}
We then detail the tokenisation methods in Section~\ref{subsec:tokenmethod} and the quantification of transferability in Section~\ref{subsec:transferability}, as illustrated in Figure~\ref{fig:pipeline}.
Finally, we describe standard adaptations of \glspl{tsfm} for anomaly detection, used as baselines, in Section~\ref{subsec:baselines}.

\subsection{Problem Formulation}
\label{subsec:formul}
Let $\fset{X}:=\{\fvm{x}_t\in\mathds{R}^D\}_{t=1}^T$ be a multivariate time-series of length $T$ with $D$ variables, with corresponding labels $\fset{Y}:=\{y_t\in\{0,1\}\}_{t=1}^T$, indicating normal and anomalous measurements.
The goal of \gls{tsad} is to find a function $\ffunc{f}$ that assigns an anomaly score $s_t=\ffunc{f}(\cdot)$ to each measurement.
Since anomalies are context-dependent, $\ffunc{f}$ operates on a observation \textit{window} of the past $L$ measurements, \ie,
$s_t=\ffunc{f}([\fvm{x}_{t-L+1},\dots,\fvm{x}_t])$, where $\ffunc{f}:\mathds{R}^{L\times D}\to\mathds{R}$.
We thus process the time-series as a set of overlapping windows $\fset{W}:=\{\fvm{W}_t\}_{t=L}^T$, with $\fvm{W}_t:=[\fvm{x}_{t-L+1},\dots,\fvm{x}_t]$.
A decision threshold $\tau\in\mathds{R}$ is applied to $s_t$ to obtain predicted labels $\Hat{y}_t=\mathds{I}[s_t\ge\tau]\in\{0,1\}$, which we compare to $y_t$ during testing to assess model performance.

\subsection{Step-Wise Tokenisation Methods}
\label{subsec:tokenmethod}

In an attempt to quantify the zero-shot transferability of sequential knowledge in \glspl{llm} to \gls{mtsad}, we explore simple, least-disruptive methods of encoding time-series for \gls{llm} input.
Two main challenges arise: (i) a mismatch between the dimensionality of multivariate time-series and the \gls{llm} embedding, and (ii) a more fundamental modality gap between continuous signals and discrete text.

Since subtle local deviations can be important in \gls{mtsad}, we prioritise encodings that preserve full temporal detail.
We therefore consider step-wise tokenisation functions $\Xi : \mathds{R}^{D} \rightarrow \mathds{R}^M$ that map each time-step to a token $\fvm{v}_t = \Xi(\fvm{x}_t)$ of dimension $M$ (\gls{llm} embedding size).
We evaluate three alignment strategies.

First, following \citet{Zhou2023OneFA}, we employ Padding, which zero-pads each time-step vector $\fvm{x}_t$.
Second, we utilise Interpolation, which rescales the input by in-filling synthetic dimensions between the original features.
These approaches are inherently non-disruptive since $M > D$.

Finally, to weigh the modality gap against potential information loss, Closest Vocabulary quantises each time-step by mapping interpolated features to their nearest token embeddings via cosine similarity.
Unlike learned mappings~\citep{Sun2023TESTTP}, this approach is non-parametric but destructive, as it discretises continuous values into a fixed vocabulary.

\subsection{Quantifying Anomaly Separability}
\label{subsec:transferability}

We use pre-trained \glspl{llm} as feature extractors and assess their zero-shot applicability to \gls{mtsad} by quantifying the isolation of anomalous representations.
The obtained token sequence $\fvm{V}_t := (\fvm{v}_{t-L+1},\dots,\fvm{v}_t)$ from a window $\fvm{W}_t$ is passed through the \gls{llm}; the processed sequence is finally flattened to obtain the feature vector $\fvm{z}_t \in \mathds{R}^{LM}$.
After obtaining the set $\mathcal{Z} := \{\fvm{z}_t\}_{t=L}^{T}$ of all window features, we assess the separability of features from normal and anomalous windows.

As no prior knowledge about the nature or structure of anomalies is assumed, we adopt a generic outlier-based perspective: anomalies are treated as observations that deviate from the distribution of normal data.
Under this assumption, we use the Isolation Forest~\citep{Liu2008IsolationF} algorithm to evaluate the suitability of the induced feature space for anomaly detection.

Isolation Forest is an unsupervised method that isolates samples by partitioning the feature space.
Let $\mathcal{F} = \{A_1, \dots, A_K\}$ denote an Isolation Forest composed of $K$ binary trees, each of depth $G\in\mathds{N}$.
Each tree $A_k$ is constructed by recursively partitioning the feature space using $G$ random splits ($\{p_{k,i}\}_{i=1}^G$) along randomly selected dimensions.
For the feature vector $\fvm{z}_t$, whose dimensionality is reduced in practice with Principal Component Analysis, its path length $\ffunc{h}_k(\fvm{z}_t)$ in tree $A_k$ is defined as the number of edges from the root to the terminal node containing $\fvm{z}_t$.
Its expected path length over the forest, $\mathds{E}_{\mathcal{F}}\left[\ffunc{h}(\fvm{z}_t)\right]$, is used to define the anomaly score as:
\begin{equation}    
s_t = 2^{-\alpha\mathbb{E}_{\mathcal{F}}\left[\ffunc{h}(\fvm{z}_t)\right]},
\end{equation}
where $\alpha$ is a normalisation constant based on the tree depth.

In this formulation, smaller values of $\mathds{E}_{\mathcal{F}}\left[\ffunc{h}(\fvm{z}_t)\right]$ (\ie, shorter paths) correspond to higher anomaly scores, which stems from the core idea of Isolation Forest that anomalous points are rare and structurally distinct, and are therefore isolated more quickly through fewer random splits.

Due to its low parameterisation and computational efficiency, Isolation Forest is particularly well suited for evaluating fixed feature representations without any additional training or optimisation.

\subsection{Forecasting \& Reconstruction Baselines}
\label{subsec:baselines}

We identify two main categories of \gls{tsfm} usage.
The first corresponds to task-specific foundation architectures, designed to address time-series challenges such as forecasting, classification, or anomaly detection.
These models are typically developed, trained, and evaluated for a specific task and dataset.
The second category consists of pre-trained models, trained on large-scale, generic objectives and subsequently transferred to new datasets and tasks.

For \gls{tsad}, both approaches generally rely on proxy objectives, most commonly reconstruction or forecasting.
In the reconstruction setting, the anomaly score $\hat{y}_t$ is derived from the \gls{mse} between the input window $\fvm{W}_t$ and its reconstruction $\Hat{\fvm{W}}_t$:

\begin{equation}
    s_t = \frac{1}{LD}  \sum_{j=1}^{L}\sum_{i=1}^{D} \left( \fvm{W}_t^{[i,j]} - \Hat{\fvm{W}}_t^{[i,j]} \right)^2.
\end{equation}

In the forecasting setting, models are given a window $\fvm{W}_t$ and predict the next $H\in\mathds{N}$ time-steps for each of the $D$ variables.
These models typically output a predictive distribution in the form of $Q\in\mathds{N}$ quantile forecasts $\hat{\fvm{H}}_{t,q}$ ($\hat{\fvm{H}}\in\mathds{R}^{H\times D}$, and $q\in [0, 1]$)
To exploit this distributional output, the PinBall loss ($\ffunc{L}_{\text{PinBall}}$), a native scoring rule for quantile regression, is usually employed:

\begin{equation}
    \ell(a, b, q) = \begin{cases} 
        q \cdot (a - b) & \text{if } a \geq b \\ 
        (1 - q) \cdot (b - a) & \text{if } a < b 
    \end{cases},
\end{equation}

\begin{equation}
    \ffunc{L}_{\text{PinBall}}(\fvm{H}_{t,q},\, \hat{\fvm{H}}_{t,q},\, q) = 
    \frac{1}{FD} \sum_{j=1}^{H} \sum_{i=1}^{D}\, 
    \ell\!\left(\fvm{H}_{t,q}^{[i,j]},\, \hat{\fvm{H}}_{t,q}^{[i,j]},\, q\right).
\end{equation}

Intuitively, observation windows are considered anomalous if they lie in the tails of the predicted distribution, \ie, if they are unlikely under the model’s forecast.
The anomaly score is calculated by aggregating the deviations across forecast steps and quantiles, resulting in a scalar value that increases with the level of surprise:

\begin{equation}
    s_t = \frac{1}{Q}\sum_{\forall q}\ffunc{L}_{\text{PinBall}}(\fvm{H}_{t,q},\, \hat{\fvm{H}}_{t,q},\, q).
\end{equation}

\section{Experiments}
\label{sec:xp}

\textit{\textbf{Protocol.}}
Evaluating models based solely on the label of the last time-step of each window $\fvm{W}_t \in \mathds{R}^{L\times D}$ can be overly restrictive.
As the window size increases, anomalies may occur earlier within the window, making this formulation unnecessarily difficult for this study.
To relax this constraint, we define the window label as $y_t = \max\left(\left\{y_i,\,\forall i \in \llbracket t -L + 1,\,t\rrbracket\right\}\right)$, \ie, a window is labelled anomalous if any of its time steps is anomalous.
This choice is a standard procedure in \gls{tsad}~\citep{Zhou2023LabelFreeMT} and only slightly alters the proportion of anomalous samples as a function of $L$.
To ensure consistent sample counts across experiments, we adopt a fixed striding strategy (set to the largest window size, 128), which has the additional effect of making each time-step contribute to at most one window.
Adjusted statistics are provided in the supplementary material, but we observed that varying these settings has limited impact on the results (complementary experiments with a unit stride are provided in supplementary).

Evaluation is performed exclusively on the test set, as the training data contains no anomalies, but no label is required by the Isolation Forest.
To account for stochasticity in \gls{pca} and Isolation Forest algorithms, all results are averaged over five random seeds for the pre-trained and random experiments.
Results of AnomalyTransformer, TimesNet, and GPT4TS baselines are also averaged across five seeds.

\begin{table}[t]
    \centering
    \caption{
Trained and pre-trained \gls{tsfm} baselines results with reconstruction (\gls{mse}) and forecasting ($\ffunc{L}_{\text{PinBall}}$) proxy objectives. AnomalyTransformer is included as a baseline for comparison. Among the remaining results, the best and second-best are shown in \textbf{bold} and \underline{underlined}, respectively.
}
    \label{tab:baselines}
    \resizebox{1\linewidth}{!}{
\renewcommand{\arraystretch}{1.2}%
\setlength{\tabcolsep}{3pt}%
\begin{tabular}{cc|ccccccccccccc}
\toprule
\multicolumn{3}{c}{\multirow[c]{2}{*}{\textbf{Model}}} & \multicolumn{3}{c}{\textbf{PSM}} & \multicolumn{3}{c}{\textbf{PUMP}} & \multicolumn{3}{c}{\textbf{SWaT}} & \multicolumn{3}{c}{\textbf{WaDi}} \\
\cmidrule(lr){4-6}
\cmidrule(lr){7-9}
\cmidrule(lr){10-12}
\cmidrule(lr){13-15}
 \multicolumn{3}{c}{}& \textbf{AUPR} & \textbf{AUROC} & \textbf{F$_1$} & \textbf{AUPR} & \textbf{AUROC} & \textbf{F$_1$} & \textbf{AUPR} & \textbf{AUROC} & \textbf{F$_1$} & \textbf{AUPR} & \textbf{AUROC} & \textbf{F$_1$} \\
\midrule\midrule
 \multirow[c]{2}{*}{\raisebox{-.9\height}{\rotatebox{90}{Trained}}}& \multirow[c]{2}{*}{\raisebox{-.9\height}{\rotatebox{90}{MSE\textsuperscript{$\dagger$}}}} & AnomalyTransformer & 68.12\stdv{00.40} & 78.97\stdv{00.64} & 64.00\stdv{00.85} & 22.00\stdv{00.07} & 79.56\stdv{00.06} & 49.09\stdv{00.75} & 69.97\stdv{01.27} & 80.88\stdv{00.10} & 71.69\stdv{02.15} & 29.27\stdv{00.01} & 53.03\stdv{00.10} & 14.99\stdv{00.14} \rule[-2ex]{0pt}{4ex} \\
\cline{3-15}
\rule{0pt}{3ex}&&GPT4TS & 52.76\stdv{00.11} & 62.05\stdv{00.07} & 48.65\stdv{00.23} & 07.35\stdv{00.00} & 21.02\stdv{00.03} & 18.66\stdv{00.05} & \textbf{10.92}\stdv{00.02} & 22.12\stdv{00.02} & \textbf{10.33}\stdv{00.13} & \underline{41.37}\stdv{00.00} & 55.82\stdv{00.01} & \underline{14.07}\stdv{00.01} \\
 &&TimesNet & 51.83\stdv{00.10} & 61.55\stdv{00.06} & 47.72\stdv{00.26} & 07.38\stdv{00.01} & 21.15\stdv{00.06} & 18.83\stdv{00.17} & \underline{10.72}\stdv{00.02} & 21.83\stdv{00.07} & \underline{09.39}\stdv{01.46} & \textbf{41.38}\stdv{00.01} & \underline{55.95}\stdv{00.07} & 14.05\stdv{00.01} \rule[-2ex]{0pt}{4ex} \\

\cline{1-15}

\rule{0pt}{3ex}\multirow[c]{7}{*}{\raisebox{0\height}{\rotatebox{90}{Pre-Trained}}}&
 \multirow[c]{4}{*}{\raisebox{0\height}{\rotatebox{90}{$\ffunc{L}_{\text{PinBall}}$}}}
 &Chronos ($F=16$) & 44.17 & 57.55 & 44.58 & 09.04 & 27.75 & 12.90 & 09.63 & 26.94 & 02.04 & 06.68 & 48.62 & 13.14 \\
&&Chronos ($F=32$) & 43.84 & 59.86 & \textbf{53.37} & 09.51 & 27.84 & 16.88 & 09.44 & 26.96 & 05.49 & 06.92 & 50.40 & 12.88 \\
&&Chronos ($F=64$) & 44.29 & 59.16 & \underline{53.10} & 10.07 & 31.99 & 20.08 & 10.10 & \underline{27.29} & 03.26 & 07.60 & 53.61 & 13.55 \\
&&Chronos ($F=128$) & 46.72 & 60.65 & 52.13 & 10.26 & 36.74 & \textbf{21.75} & 10.49 & \textbf{27.54} & 06.90 & 09.38 & \textbf{56.06} & \textbf{15.58}  \rule[-1ex]{0pt}{4ex} \\
\cline{3-15}
\rule{0pt}{3ex}&
 \multirow[c]{4}{*}{\raisebox{.5\height}{\rotatebox{90}{MSE}}}&MOMENT (Small)  & \underline{53.36} & \underline{65.11} & 47.71 & \textbf{12.50} & \underline{43.85} & 20.43 & 09.44 & 25.32 & 02.32 & 11.31 & 49.01 & 12.78 \\
 &&MOMENT (Base)  & \textbf{53.61} & \textbf{65.31} & 48.39 & \underline{12.44} & 43.57 & \underline{20.80} & 09.39 & 25.25 & 02.39 & 11.11 & 49.21 & 12.39 \\
 &&MOMENT (Large)  & 52.98 & 65.01 & 48.96 & 12.23 & \textbf{46.03} & 20.57 & 08.99 & 25.31 & 02.10 & 08.10 & 48.66 & 12.16 \\
\bottomrule
\multicolumn{15}{p{1.1\linewidth}}{\textsuperscript{$\dagger$}AnomalyTransformer has a scoring tailored for AD which comes only partially from the reconstruction error.}
\end{tabular}
}
\end{table}

\textbf{\textit{Datasets.}}
We experimented with four common multi-variate time\hyp{}series anomaly detection datasets.
Specifically, they are: Pooled Server Metrics (PSM)~\citep{Abdulaal2021PracticalAT}; PUMP Fillage (PUMP)~\citep{Feng2021TimeSA}; Secure Water Treatment (SWaT)~\citep{Mathur2016SWaTAW}; Water Distribution (WaDi)~\citep{Ahmed2017WADIAW}.
Although these datasets primarily cover the industrial domain, they represent core MTSAD benchmarks and encompass diverse types of anomalies.
Nevertheless, given the limitations and heterogeneity of MTSAD data~\citep{Wu2023CurrentTS}, we acknowledge that evaluating additional datasets would provide a more comprehensive assessment of the applicability of LLM-derived representations to MTSAD, and defer this to future work.
The statistics of the datasets can be found in the supplementary material.

\textit{\textbf{Metrics.}} To evaluate the anomaly detection performance and thus the models transferability to \gls{mtsad}, we consider three main metrics: the \gls{auroc}, the Area Under the \gls{aupr} and the \gls{f1} with a threshold set by Youden's Index.
We report \gls{f1} as an indication of in-situ performance, while acknowledging that selecting the threshold based on evaluation across the full test set introduces label dependence.
Youden's Index nevertheless provides a consistent, balanced criterion for inter-model comparison, avoiding \gls{f1}-maximising threshold sweeps or anomaly-ratio-based thresholding, both of which are common in \gls{tsad} studies~\citep{Wu2022TimesNetT2}.

\textit{\textbf{Model Architectures.}} We chose to use four open source \gls{llm} architectures with available weights in our pipeline:
GPT2~\citep{Radford2019LanguageMA}, Qwen2 (0.5B)~\citep{Yang2024Qwen2TR}, Phi-3~\citep{Abdin2024Phi3TR} and Mistral-7B~\citep{Jiang2023Mistral7}.
In Section~\ref{subsec:multimodal}, we study two additional Qwen versions: Qwen2 (7B)~\citep{Yang2024Qwen2TR} and Qwen2-VL  (7B)~\citep{Wang2024Qwen2VLEV}.
These architectures were selected for their diverse origins, likely distinct pre-training corpora, ease of integration into our current systems, and comparable release timelines.

We complement our study with four \gls{tsfm} baselines outside of our pipeline: TimesNet~\citep{Wu2022TimesNetT2}, GPT4TS~\citep{Zhou2023OneFA}, 
Chronos-2~\citep{Ansari2025Chronos2FU} and MOMENT~\citep{Goswami2024MOMENTAF}.
TimesNet and GPT4TS are reconstruction-based methods, the latter being \gls{llm}-based, that are specifically trained on each dataset (Section~\ref{subsec:tsfmbaseline}).
On the other hand, Chronos-2 is already pre-trained on a forecasting task and can be used in a zero-shot setting with a forecasting strategy (Section~\ref{subsec:tsfmbaseline}), or as a feature extractor (Section~\ref{subsec:results}).
Similarly, MOMENT is pre-trained on a reconstruction objective and can be used in a zero-shot setting with a reconstruction error scoring strategy (Section~\ref{subsec:tsfmbaseline}).
Additionally, we trained AnomalyTransformer~\citep{Xu2021AnomalyTT} to provide ground results from a method that is entirely designed and optimised to detect anomalies in time-series.

Model version identifiers are omitted hereafter for conciseness; for all figures, $\Delta(a, b):= a -b$ and positive results imply better average performance than the baseline.

\subsection{Performance of \glspl{tsfm}}
\label{subsec:tsfmbaseline}

We report results for standard \gls{tsfm} approaches applied to \gls{mtsad} in Table~\ref{tab:baselines}.
TimesNet and GPT4TS are trained in an unsupervised manner using a reconstruction objective, with anomaly scores derived from the reconstruction error (\gls{mse}).
In contrast, Chronos is used as a pre-trained model, and anomaly scores are computed from the forecasting error using the Pinball loss ($\ffunc{L}_{\text{PinBall}}$).
Results are reported for forecasting horizons of 16, 32, 64, and 128.

TimesNet and GPT4TS yield nearly identical performance, consistent with recent \gls{tsad} benchmarks~\citep{Qiu2025TABUB}, and perform similarly to other baselines across most datasets.
Chronos and MOMENT largely surpass TimesNet and GPT4TS on PUMP, and achieve the majority of the top \gls{aupr} and \gls{auroc} results on other datasets.
However, \gls{auroc} scores stay close to or below 50\% across most setting, and noticeably behind AnomalyTransformer on three out of the four studied datasets.
These results suggest that standard forecasting and reconstruction objectives in \glspl{tsfm}, trained or zero-shot, are suboptimal for robust anomaly detection, creating an opportunity to investigate alternative strategies including \gls{llm}-driven approaches.

Evaluating MOMENT across its Small, Base, and Large variants (roughly scaled by a factor of three in parameter count) reveals marginal but overall slightly declining performance differences.
This suggests that increased model capacity does not necessarily improve performance in \gls{tsad}, aligning with our concern that higher reconstruction capability may even degrade detection quality.

\begin{table}[t]
    \centering
    \caption{
Results of the studied \glspl{llm} under the proposed framework.
For conciseness, only results for the best-performing window size are reported.
For each metric, average over window-sizes and overall average across window sizes and datasets are also reported for both pre-trained and randomised models.
}
    \label{tab:llm-best}
    \resizebox{1\linewidth}{!}{
\renewcommand{\arraystretch}{1.2}%
\setlength{\tabcolsep}{3pt}%
\newcommand{\rand}[1]{*#1}
\begin{tabular}{cccccccccccccc}
\toprule
 \multicolumn{2}{c}{\multirow[c]{2}{*}{\textbf{Model}}} & \multicolumn{3}{c}{\textbf{PSM}} & \multicolumn{3}{c}{\textbf{PUMP}} & \multicolumn{3}{c}{\textbf{SWaT}} & \multicolumn{3}{c}{\textbf{WaDi}} \\
\cmidrule(lr){3-5}
\cmidrule(lr){6-8}
\cmidrule(lr){9-11}
\cmidrule(lr){12-14}
 & & \textbf{AUPR} & \textbf{AUROC} & \textbf{F$_1$} & \textbf{AUPR} & \textbf{AUROC} & \textbf{F$_1$} & \textbf{AUPR} & \textbf{AUROC} & \textbf{F$_1$} & \textbf{AUPR} & \textbf{AUROC} & \textbf{F$_1$} \\
\midrule\midrule
\multirow[c]{4}{*}{GPT} & $L=32$ & 32.92\stdv{0.41} & 54.08\stdv{0.94} & 35.81\stdv{4.44} & 17.14\stdv{1.08} & 67.92\stdv{1.27} & 32.98\stdv{2.14} & 53.10\stdv{6.01} & 67.71\stdv{0.76} & 55.55\stdv{7.40} & 15.22\stdv{6.46} & 43.17\stdv{2.77} & 12.53\stdv{0.16} \\
 & \multirow[c]{2}{*}{$\mathds{E}_L$} & 33.70 & 54.86 & 37.13 & 16.44 & 66.86 & 31.78 & 44.74 & 65.79 & 49.47 & 12.83 & 41.44 & 12.14 \\
  & & \rand{36.13} & \rand{56.84} & \rand{39.24} & \rand{18.72} & \rand{73.40} & \rand{38.44} & \rand{46.66} & \rand{67.07} & \rand{48.91} & \rand{16.74} & \rand{43.23} & \rand{12.37} \\
\cmidrule(lr){9-11}
\cmidrule(lr){12-14}
 & \multirow[c]{1}{*}{$\mathds{E}_{L, \fset{X}}$} &&-&&&-&& 26.93 & 57.24 & 32.63 
 & \rand{29.56} & \rand{60.13} & \rand{34.74} \\
\midrule
\multirow[c]{4}{*}{Mistral} & $L=4$ & 32.88\stdv{0.89} & 56.78\stdv{0.92} & 37.38\stdv{5.91} & 18.47\stdv{0.78} & 74.92\stdv{1.36} & 38.51\stdv{0.99} & 26.36\stdv{10.52} & 64.59\stdv{3.06} & 36.03\stdv{4.84} & 21.40\stdv{3.10} & 45.40\stdv{1.33} & 11.96\stdv{0.32} \\
 & \multirow[c]{2}{*}{$\mathds{E}_{L}$} & 34.99 & 54.54 & 35.31 & 19.18 & 74.43 & 39.62 & 27.49 & 63.41 & 35.23 & 21.08 & 45.34 & 12.44 \\
 & & \rand{36.03} & \rand{55.46} & \rand{39.79} & \rand{17.60} & \rand{72.22} & \rand{36.93} & \rand{48.66} & \rand{68.00} & \rand{49.25} & \rand{23.30} & \rand{44.86} & \rand{11.99} \\
\cmidrule(lr){9-11}
\cmidrule(lr){12-14}
 & \multirow[c]{1}{*}{$\mathds{E}_{L, \fset{X}}$} &&-&&&-&& \multicolumn{1}{c}{25.68} & 59.43 & 30.65 & \multicolumn{1}{c}{\rand{31.40}} & \rand{60.14} & \rand{34.49} \\
\midrule
\multirow[c]{4}{*}{Phi} & $L=16$ & 29.44\stdv{0.54} & 53.60\stdv{0.42} & 43.36\stdv{2.19} & 16.03\stdv{1.17} & 68.41\stdv{2.31} & 32.96\stdv{2.44} & 47.78\stdv{12.76} & 67.37\stdv{1.41} & 50.06\stdv{10.85} & 16.31\stdv{5.60} & 41.39\stdv{1.81} & 11.46\stdv{0.27} \\
 & \multirow[c]{2}{*}{$\mathds{E}_{L}$} & 30.14 & 51.89 & 43.71 & 15.73 & 66.86 & 31.82 & 42.35 & 65.64 & 46.62 & 15.90 & 41.24 & 11.88 \\
 & & \rand{35.48} & \rand{55.03} & \rand{39.61} & \rand{17.28} & \rand{71.46} & \rand{36.70} & \rand{46.88} & \rand{67.57} & \rand{46.85} & \rand{23.95} & \rand{44.66} & \rand{11.87} \\
\cmidrule(lr){9-11}
\cmidrule(lr){12-14}
 & \multirow[c]{1}{*}{$\mathds{E}_{L, \fset{X}}$} &&-&&&-&& 26.03 & 56.41 & 33.51
 & \rand{30.90} & \rand{59.68} & \rand{33.76}\\
\midrule
\multirow[c]{4}{*}{Qwen} & $L=8$ & 50.72\stdv{0.64} & 67.80\stdv{0.58} & 49.21\stdv{1.19} & 17.94\stdv{0.43} & 73.91\stdv{0.53} & 36.15\stdv{0.74} & 30.71\stdv{12.67} & 65.32\stdv{3.50} & 39.47\stdv{8.41} & 13.93\stdv{4.75} & 43.09\stdv{2.27} & 11.74\stdv{0.15} \\
 & \multirow[c]{2}{*}{$\mathds{E}_{L}$} & 50.58 & 65.51 & 48.78 & 17.70 & 72.22 & 36.77 & 27.06 & 63.91 & 37.22 & 18.33 & 44.18 & 12.31 \\
 & & \rand{38.25} & \rand{56.25} & \rand{35.36} & \rand{18.32} & \rand{73.36} & \rand{38.24} & \rand{38.69} & \rand{65.82} & \rand{39.92} & \rand{20.62} & \rand{44.58} & \rand{12.24} \\
\cmidrule(lr){9-11}
\cmidrule(lr){12-14}
 & \multirow[c]{1}{*}{$\mathds{E}_{L, \fset{X}}$} &&-&&&-&& 28.42 & 61.46 & 33.77 & \rand{28.97} & \rand{60.00} & \rand{31.44} \\
\bottomrule
\multicolumn{14}{p{1.3\linewidth}}{*Results of the randomised models. $\mathds{E}_L$: Metric average across window sizes. $\mathds{E}_{L, \fset{X}}$: Metric average across window sizes and datasets.}
\end{tabular}
}
\end{table}

\subsection{Tokenisation Strategies}
\label{subsec:tokenisation}

\begin{figure}[t]
\begin{minipage}[t]{.49\linewidth}
    \centering
    \includegraphics[width=\linewidth]{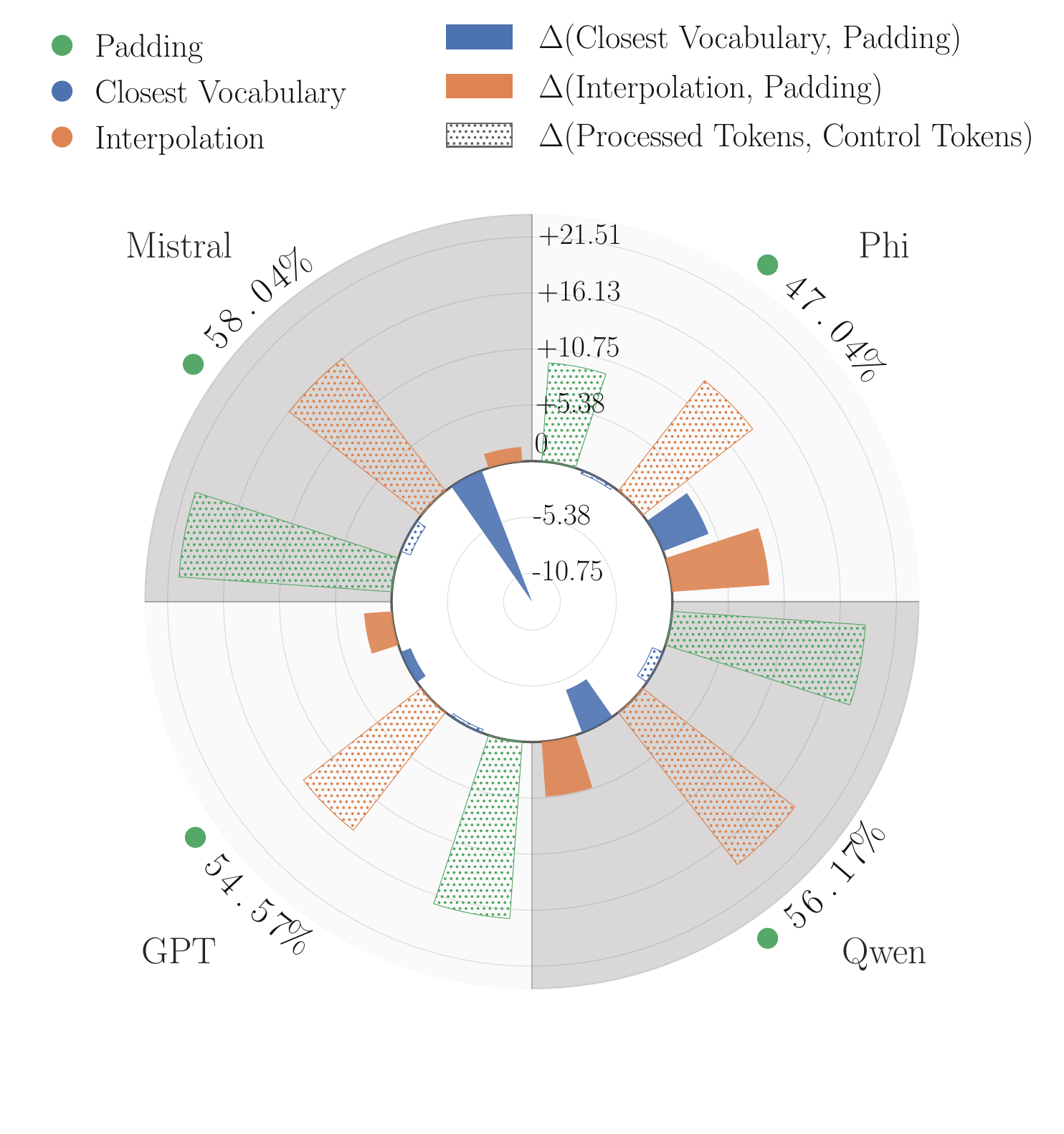}\\[-2.5em]
\caption{Comparison of 3 step-wise tokenisation methods.
We expose the \gls{auroc} result differences between the \textit{Control} path and after the \glspl{llm} (dotted bars).
For plain bars, results with Padding serve as a baseline and study the differences between those and the ones achieved with Closest Vocabulary and Interpolation.
Padding average scores are indicated on the slices sides.
}
    \label{fig:tokenisations}
\end{minipage}
\hfill
\begin{minipage}[t]{.49\linewidth}
    \centering
     \includegraphics[width=\linewidth]{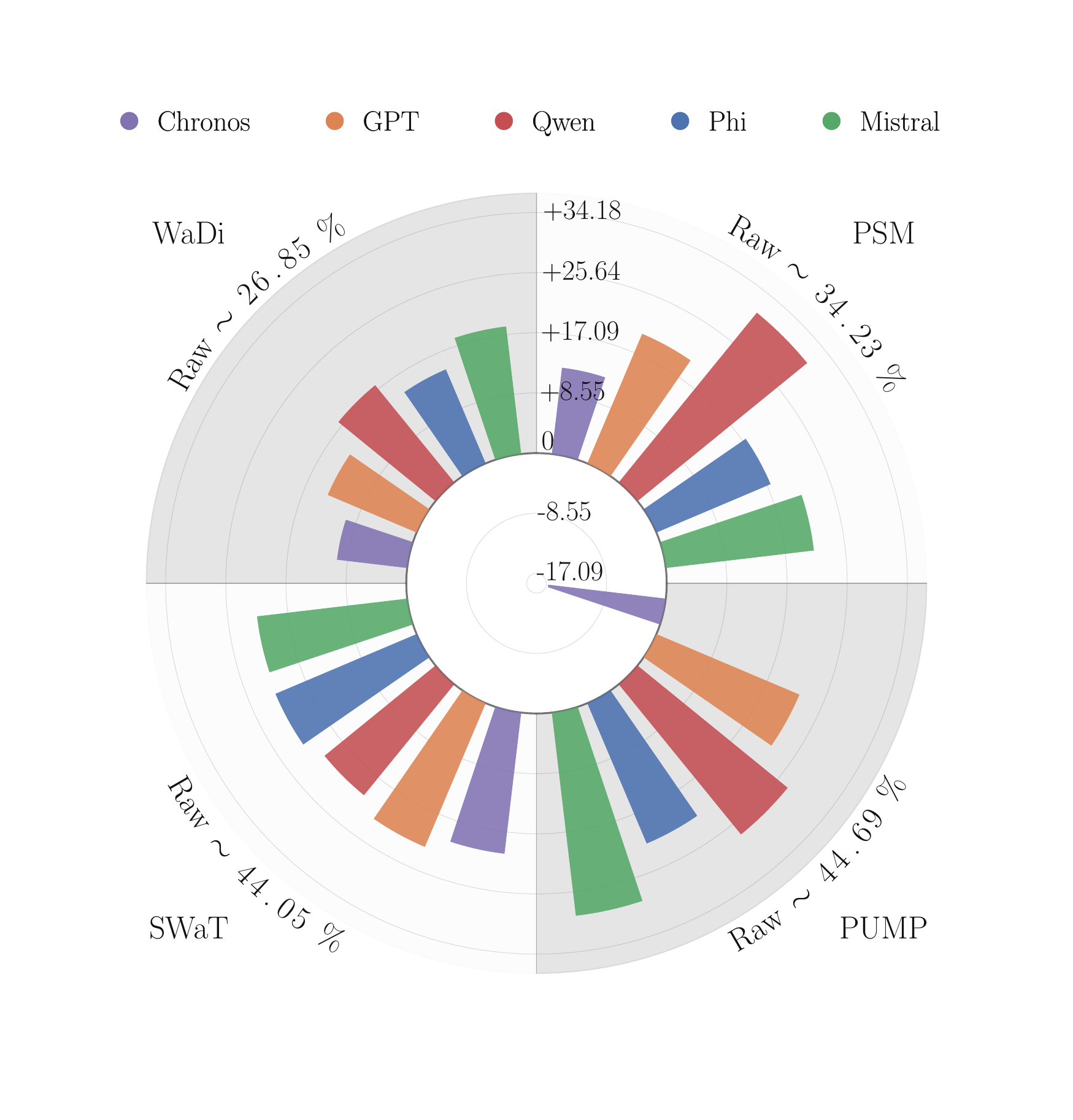}\\[-2.5em]
\caption{\gls{auroc} results of the studied \glspl{llm} on different datasets, with Interpolation.
We show the results difference with the baseline (\ie, \textit{Raw} path) and the results achieved by each \gls{llm}, \ie, $\Delta$(LLM, \textit{Raw}).
The average baseline results are indicated on the slices sides and are expected results on the $0$ ring.
}
    \label{fig:results_datasets}
\end{minipage}
\end{figure}

We explore three tokenisation strategies: Padding, Interpolation, and Closest Vocabulary.
For each method, results are averaged across models and datasets, and reported relative to a baseline (Figure~\ref{fig:tokenisations}).

We first compare the performance of an Isolation Forest applied directly to the tokens (\textit{Control} as a baseline) with the performance obtained after \gls{llm} processing (dotted bars).
For Padding and Interpolation, \gls{llm} processing yields substantial improvements, with gains of up to $\approx\,$20 \gls{auroc} points for Mistral and Qwen.
In contrast, no improvement is observed for Closest Vocabulary, where performance even slightly degrades.
This suggests that the benefit of \glspl{llm} emerges most when tokenisation preserves sufficient information, allowing them to meaningfully enrich the input representations.

To compare tokenisation strategies directly, we then take Padding as a baseline and report relative differences (after \gls{llm} processing).
Closest Vocabulary consistently underperforms, with drops of up to 10 \gls{auroc} points for Mistral.
Interpolation achieves the best results overall, although the gains remain modest for GPT, Mistral, and Qwen (up to $\approx\,$5 points) and are more pronounced for Phi (up to $\approx\,$10 points).

Overall, these results indicate that tokenisation has a significant impact on performance, with even simple design choices leading to substantial variations.

\subsection{Zero-Shot \gls{llm} Features Evaluation}
\label{subsec:results}

\gls{auroc} results are summarised in Figure~\ref{fig:results_datasets}.
To analyse them, we focus on score differences relative to a baseline obtained by applying Isolation Forest directly on the raw datasets (\textit{Raw} path in Figure~\ref{fig:pipeline}).
Results are aggregated as in previous sections.
A brief view of quantitative results can be found in Table~\ref{tab:llm-best}, and complete detailed results are provided in the supplementary.

We additionally include Chronos as a \gls{tsfm} baseline.
Unlike the previous setup in Section~\ref{subsec:tsfmbaseline}, we use it here as a feature extractor rather than relying on forecasting error: the pre-trained model processes raw time-series directly, and Isolation Forest is applied on the resulting embeddings.
As it operates natively on continuous inputs, no tokenisation is required.

Across datasets, Chronos features consistently yields the lowest performance compared to LLM-based approaches, and even falls below the \textit{Raw} baseline on PUMP.
However, its feature-based formulation still improves substantially over its forecasting-based use on SWaT, with gains exceeding 40 \gls{auroc} points on average, and achieving similar results on PUMP at $L=128$.

Overall, all \gls{llm}-based methods improve over the \textit{Raw} baseline with consistent gains across window sizes and datasets, showcasing the benefits of \gls{llm} processing.
Phi approaches the performance of the trained GPT4TS baseline, while GPT, Qwen, and Mistral surpass MOMENT on PSM, PUMP and SWaT.

\gls{aupr} and \gls{f1} exhibit strongly correlated patterns; visualisations for these metrics, as well as detailed results per window size, dataset, and tokenisation are provided in the supplementary material.
While directly comparing \gls{tsfm} baseline scores with Isolation Forest scores obtained from \gls{llm} features is not strictly fair, since the latter uses all the test data to construct the forest, the competitive performance of the \gls{llm}-based approach demonstrates the viability of these features for anomaly detection.
More broadly, these results support using \glspl{llm} as zero-shot feature extractors, providing an alternative to adapting standard \gls{tsfm} architectures for anomaly detection through proxy objectives such as forecasting error, while allowing the resulting feature space to be exploited by other fully unsupervised anomaly detection algorithms.

\subsection{Impact of Scale and Multi-Modality-Awareness}
\label{subsec:multimodal}

We assess whether multi-modal pre-training improves transferability to time-series data by comparing Qwen-VL (7B) with its text-only counterpart Qwen (7B), alongside the smaller baseline Qwen (0.5B).

\begin{figure}[t]
\begin{minipage}[t]{.49\linewidth}
    \centering
     \includegraphics[width=.97\linewidth]{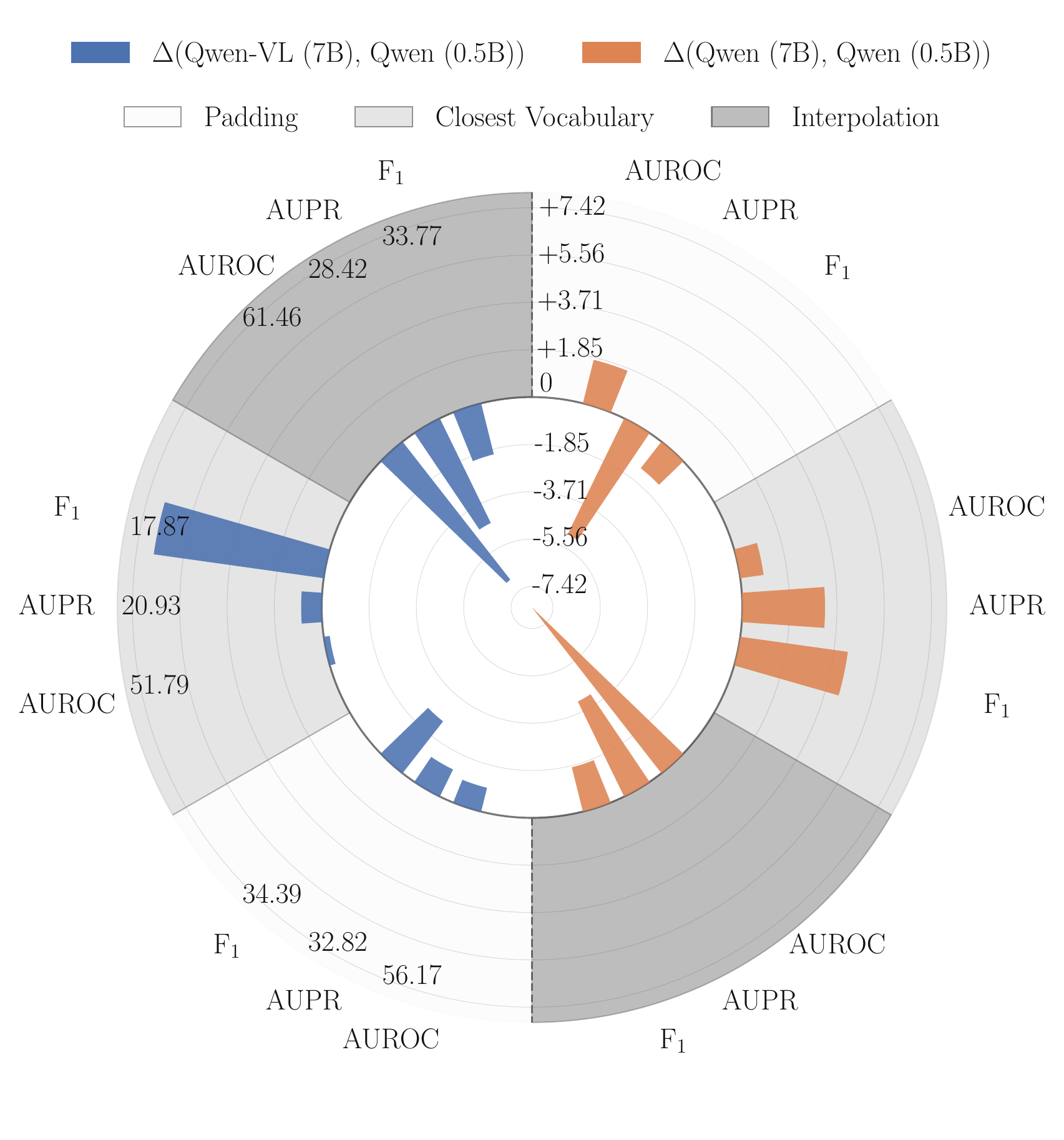}\\[-1.5em]
\caption{Results achieved by Qwen (0.5B), Qwen (7B) and Qwen-VL (7B), on the 3 tokenisation methods studied.
Qwen (0.5B) results, with the respective tokenisation method, serve as a baseline and are shown on the left-hand side of the plot.
Qwen (0.5B) baseline scores are indicated below the metrics on the left half.
}
    \label{fig:multimodal}
\end{minipage}
\hfill
\begin{minipage}[t]{.49\linewidth}
    \centering
     \includegraphics[width=\linewidth]{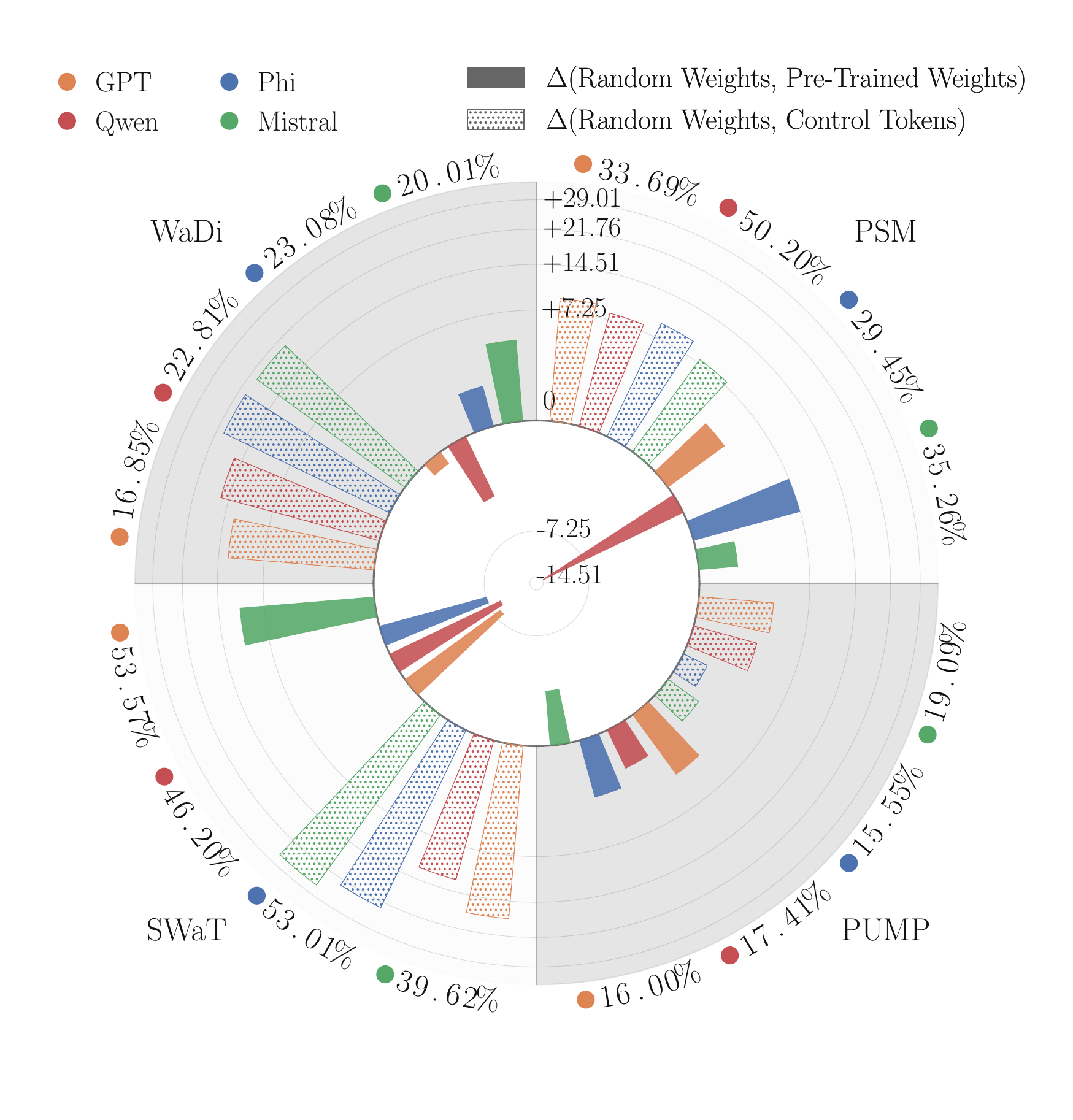}\\[-1.5em]
\caption{Comparison of \gls{aupr} obtained by applying Isolation Forest to features from randomly initialised models (Random Weights) against two baselines: features from pre-trained models (Pre-Trained Weights, solid bars) and without feature extraction (\textit{Control} Tokens, dotted bars).
Average scores with pre-trained features are shown on the slices sides.
Radial scale is gamma corrected ($\gamma=0.5$).
}
    \label{fig:randominit_aupr}
\end{minipage}
\end{figure}

Results in Figure~\ref{fig:multimodal} show that, for Padding and Interpolation tokenisations, the smaller Qwen (0.5B) consistently outperforms both larger models by 7 to 8 \gls{auroc} points.
This suggests that neither increased scale nor multi-modal pre-training provides benefits in this zero-shot setting under these tokenisation schemes.
In contrast, larger models perform better with Closest Vocabulary tokenisation despite similar vocabulary sizes ($\approx\,$150k), likely due to their higher embedding dimensionality (896 $\rightarrow$ 3584) which increases vocabulary capacity and may reduce information loss during discretisation.

However, the performance of Qwen (0.5B) remains consistently superior to both larger versions across all metrics with the two other tokenisation methods, achieving its best average scores and greater gaps with Interpolation.
Overall, these results indicate that tokenisation has a stronger effect on performance than either model scale or multi-modal pre-training in this setting.

\subsection{Inductive Bias and Knowledge Transfer}

In the previous sections, we showed that features extracted from \glspl{llm} consistently improve anomaly detection performance.
However, it remains unclear whether these gains arise from (i) the attention-based architectures themselves, whose inductive bias may naturally produce representations that isolate anomalous time-series, or (ii) knowledge transferred from language pre-training.

To investigate this, Figure~\ref{fig:randominit_aupr} compares the average \gls{aupr} of each model, randomly initialised over five different seeds, with results on \textit{Control} tokens (dotted bars) and the corresponding pre-trained models (solid bars) with the Interpolation tokenisation.
The Isolation Forest is always used with the same random state across all features.
Average quantitative results are reported in Table~\ref{tab:llm-best} (see \textit{*scores}), and full results, including with the other two tokenisation strategies are provided in the supplementary.

Randomly initialised models consistently outperform \textit{Control} tokens classification (positive dotted bars), achieving gains of up to 29 points for Mistral on SWaT.
Surprisingly, pre-training does not consistently improve performance (negative plain bars).
Frequently even, pre-trained models perform worse than their randomly initialised counterparts.
Only Qwen improves on three of the four datasets, while the other models improve on at most one, supporting Qwen's overall superiority (Figure~\ref{fig:results_datasets}).
Despite using random weights, the deterministic sequence of operations seems sufficient to increase feature dissociation.

To further isolate the impact of the architecture, we compared the results against those from a random linear projection and a randomly initialized 3-layer \gls{mlp}.
Random \glspl{llm} significantly outperformed both baselines (Figure~\ref{fig:wilcoxon}; full linear and \gls{mlp} results are provided in the supplementary).
Combined with the consistent performance observed across different \gls{llm} architectures, these findings suggest that this property is an intrinsic feature of Transformer-based \gls{nlp} models rather than being tied to specific model designs.

These results suggest that pre-training alone is insufficient to guarantee effective zero-shot cross-modal feature extraction, and non-trivial tokenisation strategies or task-specific fine-tuning are required to exploit the knowledge encoded in \glspl{llm}.
Overall, the architectures themselves seem to provide most of the zero-shot feature extraction capability, whereas effectively leveraging pre-trained knowledge, in a general setting, requires more appropriate tokenisation or task-specific fine-tuning.
Similar patterns are observed with \gls{auroc} and \gls{f1} metrics; figures are provided in the supplementary material.

\begin{figure}
    \centering
    \includegraphics[width=.95\linewidth]{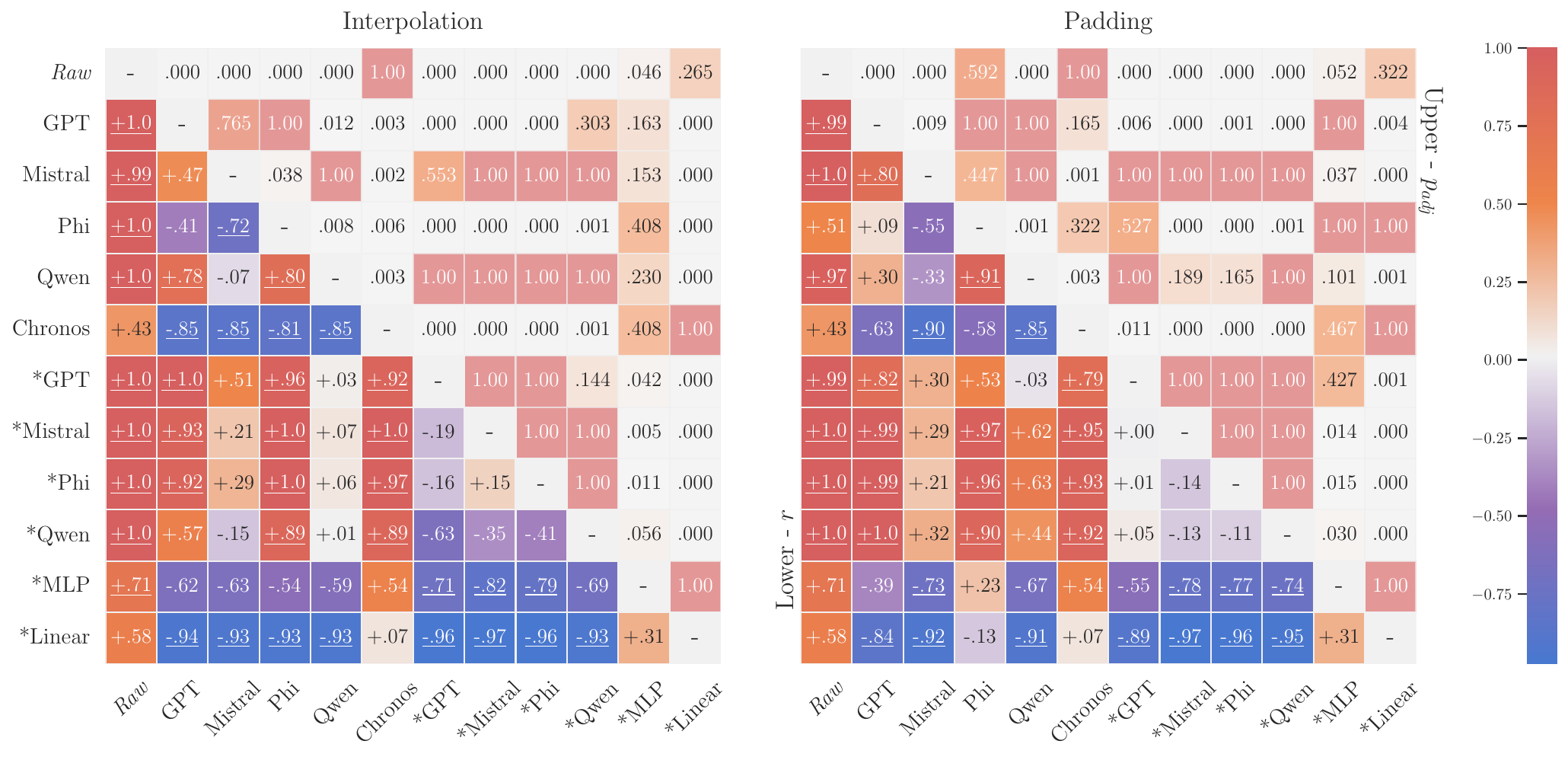}
    \caption{Pairwise \gls{auroc} comparisons between models across datasets and window sizes, for Interpolation and Padding tokenisations.
    \textit{Lower Triangle:} Effect size $r \in [-1,1]$.
    A positive score implies the row model outperforms the column model; statistically significant tests ($p_{adj}<0.05$) are \underline{underlined}.
    \textit{Upper Triangle: } Corresponding adjusted $p$-value ($p_{adj}\in[0,1]$) of the paired test.  
    *Denotes the randomised models.}
    \label{fig:wilcoxon}
\end{figure}

\subsection{Pairwise Statistical Analysis}

To assess whether performance differences between the considered \glspl{llm} are statistically significant, we perform a two-sided paired Wilcoxon signed-rank test~\citep{Wilcoxon1945IndividualCB} on their \gls{auroc} scores.
For each pair of \glspl{llm}, the test is performed over the six window sizes and four datasets, yielding $24$ paired measurements per comparison.
We report the Holm-adjusted $p$-value ($p_{\mathrm{adj}}$) to account for multiple pairwise comparisons, together with the signed rank-biserial correlation $r$ as an effect-size measure.
The $p_{\mathrm{adj}}$ indicates whether the paired performance differences provide evidence against the null hypothesis of a zero-centred difference, while $|r|$ quantifies the magnitude of the rank-based difference between the two \glspl{llm}.

This procedure is repeated independently for each adapter configuration.
We discuss the results for Interpolation and Padding, shown in Figure~\ref{fig:wilcoxon}; results for Closest Vocabulary are mostly statistically insignificant and are provided in the supplementary.

For Interpolation, we observe a significant improvement over the \textit{Raw} tokens for all pre-trained \glspl{llm}.
Qwen achieves significantly higher \gls{auroc} than all other \glspl{llm}, except Mistral, for which no significant difference is observed ($p_{\mathrm{adj}}=1.00$).
Chronos features consistently underperform the \gls{llm}-based representations.
Randomised \glspl{llm} also exhibit consistently positive effects, significantly outperforming \textit{Raw}, GPT, Phi, randomised \gls{mlp}, and randomised Linear.
The performance between the pre-trained rand randomised versions of Qwen and Mistral is, however, statistically inconclusive.

The Padding results exhibit a similar behaviour for the randomised models.
However, the differences between pre-trained \glspl{llm} are substantially less pronounced, with fewer statistically significant comparisons.
When comparing Padding and Interpolation relative to their respective \textit{Raw} baselines, Phi is the only pre-trained \gls{llm} for which Padding leads to a lower performance than Interpolation.

Overall, the strong performance of the randomised models under both tokenisations complements the previous observations on the limited ability of simplistic tokenisation strategies to exploit \gls{llm} knowledge.
Furthermore, the loss of statistical significance in the differences between pre-trained models when switching from Interpolation to Padding, not observed for randomised models, further suggests that bridging the cross-domain gap between time-series and \gls{nlp} representations is non-trivial.

\section{Conclusion}
\label{sec:concl}

In this work, we investigated the use of \gls{llm} representations for time-series anomaly detection in a zero-shot setting.
We proposed an evaluation pipeline to assess the suitability of \gls{llm} representations to \gls{mtsad} by framing anomaly detection as an unsupervised classification problem and quantifying the isolation of anomalous representations.
Our study centres on evaluating time-series feature representation; we extract high-dimensional embeddings from pre-trained \glspl{llm} and evaluate different steps chosen to minimise the loss of local temporal information.
By passing these tokenised sequences through the \gls{llm}, we generate feature representations that are then analysed via an Isolation Forest for anomaly detection.
Ultimately, the resulting anomaly scores serve as a robust proxy for evaluating the quality of the alignment and the suitability of the representations for \gls{mtsad}.

Our results show that even in a zero-shot setting, several \glspl{llm} extract useful feature representations for anomaly detection.
When combined with shallow unsupervised classifiers, these representations can outperform recent off-the-shelf time-series foundation models based on forecasting error.
However, our analysis also indicates that simple tokenisation strategies are generally insufficient to effectively exploit the knowledge encoded in pre-trained \glspl{llm} for \gls{mtsad}.
In fact, we established that in this setting randomly initialised \glspl{llm} often achieve significantly strong performance, suggesting that the architectural components of \glspl{llm} provide greater benefits than any knowledge acquired through language pre-training.

This study aims to support a research direction in \gls{mtsad} with foundation models.
Framed around the specificities of \gls{tsad}, it highlights the potential of \glspl{llm} as feature extractors for time-series data.
In particular, it provides clues that \glspl{llm} can serve as effective representation refiners in this setting.
A promising line of research lies in improving tokenisation strategies to better bridge the domain gap and reduce reliance on fine-tuning, while preserving the local structure of the data that is crucial for anomaly detection.

Future work will extend this study to more language models and model versions, a broader range of datasets, and unsupervised classification methods in order to better account for the diversity of anomaly types and feature space geometries beyond the isolation assumption.

\subsubsection*{Acknowledgments}

This research was funded by the Luxembourg National Research Fund (FNR), grant reference \texttt{DEFENCE22/17813724/AUREA}.

\bibliography{bib}

@article{Wolf2019HuggingFacesTS,
  author       = {Thomas Wolf and
                  Lysandre Debut and
                  Victor Sanh and
                  Julien Chaumond and
                  Clement Delangue and
                  Anthony Moi and
                  Pierric Cistac and
                  Tim Rault and
                  R{\'{e}}mi Louf and
                  Morgan Funtowicz and
                  Jamie Brew},
  title        = {HuggingFace's Transformers: State-of-the-art Natural Language Processing},
  journal      = {CoRR},
  volume       = {abs/1910.03771},
  year         = {2019}
}

@techreport{Radford2019LanguageMA,
  title={Language models are unsupervised multitask learners},
  author={Radford, Alec and Wu, Jeffrey and others},
  journal={OpenAI Blog},
  year={2019}
}

@article{Wang2024Qwen2VLEV,
  author       = {Peng Wang and
                  Shuai Bai and
                  Sinan Tan and
                  Shijie Wang and
                  Zhihao Fan and
                  Jinze Bai and
                  Keqin Chen and
                  Xuejing Liu and
                  Jialin Wang and
                  Wenbin Ge and
                  Yang Fan and
                  Kai Dang and
                  Mengfei Du and
                  Xuancheng Ren and
                  Rui Men and
                  Dayiheng Liu and
                  Chang Zhou and
                  Jingren Zhou and
                  Junyang Lin},
  title        = {Qwen2-VL: Enhancing Vision-Language Model's Perception of the
                  World at Any Resolution},
  journal      = {CoRR},
  volume       = {abs/2409.12191},
  year         = {2024}
}

@article{Yang2024Qwen2TR,
  author       = {An Yang and
                  Baosong Yang and
                  Binyuan Hui and
                  Bo Zheng and
                  Bowen Yu and
                  Chang Zhou and
                  Chengpeng Li and
                  Chengyuan Li and
                  Dayiheng Liu and
                  Fei Huang and
                  Guanting Dong and
                  Haoran Wei and
                  Huan Lin and
                  Jialong Tang and
                  Jialin Wang and
                  Jian Yang and
                  Jianhong Tu and
                  Jianwei Zhang and
                  Jianxin Ma and
                  Jianxin Yang and
                  Jin Xu and
                  Jingren Zhou and
                  Jinze Bai and
                  Jinzheng He and
                  Junyang Lin and
                  Kai Dang and
                  Keming Lu and
                  Keqin Chen and
                  Kexin Yang and
                  Mei Li and
                  Mingfeng Xue and
                  Na Ni and
                  Pei Zhang and
                  Peng Wang and
                  Ru Peng and
                  Rui Men and
                  Ruize Gao and
                  Runji Lin and
                  Shijie Wang and
                  Shuai Bai and
                  Sinan Tan and
                  Tianhang Zhu and
                  Tianhao Li and
                  Tianyu Liu and
                  Wenbin Ge and
                  Xiaodong Deng and
                  Xiaohuan Zhou and
                  Xingzhang Ren and
                  Xinyu Zhang and
                  Xipin Wei and
                  Xuancheng Ren and
                  Xuejing Liu and
                  Yang Fan and
                  Yang Yao and
                  Yichang Zhang and
                  Yu Wan and
                  Yunfei Chu and
                  Yuqiong Liu and
                  Zeyu Cui and
                  Zhenru Zhang and
                  Zhifang Guo and
                  Zhihao Fan},
  title        = {Qwen2 Technical Report},
  journal      = {CoRR},
  volume       = {abs/2407.10671},
  year         = {2024}
}

@article{Ansari2025Chronos2FU,
  author       = {Abdul Fatir Ansari and
                  Oleksandr Shchur and
                  Jaris K{\"{u}}ken and
                  Andreas Auer and
                  Boran Han and
                  Pedro Mercado and
                  Syama Sundar Rangapuram and
                  Huibin Shen and
                  Lorenzo Stella and
                  Xiyuan Zhang and
                  Mononito Goswami and
                  Shubham Kapoor and
                  Danielle C. Maddix and
                  Pablo Guerron and
                  Tony Hu and
                  Junming Yin and
                  Nick Erickson and
                  Prateek Mutalik Desai and
                  Hao Wang and
                  Huzefa Rangwala and
                  George Karypis and
                  Yuyang Wang and
                  Michael Bohlke{-}Schneider},
  title        = {Chronos-2: From Univariate to Universal Forecasting},
  journal      = {CoRR},
  volume       = {abs/2510.15821},
  year         = {2025}
}

@article{Abdin2024Phi3TR,
  author       = {Microsoft},
  title        = {Phi-3 Technical Report: {A} Highly Capable Language Model Locally
                  on Your Phone},
  journal      = {CoRR},
  volume       = {abs/2404.14219},
  year         = {2024}
}

@article{Jiang2023Mistral7,
  author       = {Albert Q. Jiang and
                  Alexandre Sablayrolles and
                  Arthur Mensch and
                  Chris Bamford and
                  Devendra Singh Chaplot and
                  Diego de Las Casas and
                  Florian Bressand and
                  Gianna Lengyel and
                  Guillaume Lample and
                  Lucile Saulnier and
                  L{\'{e}}lio Renard Lavaud and
                  Marie{-}Anne Lachaux and
                  Pierre Stock and
                  Teven Le Scao and
                  Thibaut Lavril and
                  Thomas Wang and
                  Timoth{\'{e}}e Lacroix and
                  William El Sayed},
  title        = {Mistral 7B},
  journal      = {CoRR},
  volume       = {abs/2310.06825},
  year         = {2023}
}

@inproceedings{Abdulaal2021PracticalAT,
  author       = {Ahmed Abdulaal and
                  Zhuanghua Liu and
                  Tomer Lancewicki},
  title        = {Practical Approach to Asynchronous Multivariate Time Series Anomaly
                  Detection and Localization},
  booktitle    = {{KDD}},
  pages        = {2485--2494},
  publisher    = {{ACM}},
  year         = {2021}
}

@inproceedings{Mathur2016SWaTAW,
  author       = {Aditya P. Mathur and
                  Nils Ole Tippenhauer},
  title        = {SWaT: a water treatment testbed for research and training on {ICS}
                  security},
  booktitle    = {CySWATER@CPSWeek},
  pages        = {31--36},
  publisher    = {{IEEE} Computer Society},
  year         = {2016}
}

@inproceedings{Feng2021TimeSA,
  author       = {Cheng Feng and
                  Pengwei Tian},
  title        = {Time Series Anomaly Detection for Cyber-physical Systems via Neural
                  System Identification and Bayesian Filtering},
  booktitle    = {{KDD}},
  pages        = {2858--2867},
  publisher    = {{ACM}},
  year         = {2021}
}

@inproceedings{Ahmed2017WADIAW,
  author       = {Chuadhry Mujeeb Ahmed and
                  Venkata Reddy Palleti and
                  Aditya P. Mathur},
  title        = {{WADI:} a water distribution testbed for research in the design of
                  secure cyber physical systems},
  booktitle    = {CySWATER@CPSWeek},
  pages        = {25--28},
  publisher    = {{ACM}},
  year         = {2017}
}

@article{Spathis2023TheFS,
  author       = {Dimitris Spathis and
                  Fahim Kawsar},
  title        = {\emph{The first step is the hardest}: pitfalls of representing and
                  tokenizing temporal data for large language models},
  journal      = {J. Am. Medical Informatics Assoc.},
  volume       = {31},
  number       = {9},
  pages        = {2151--2158},
  year         = {2024}
}

@inproceedings{Masserano2024EnhancingFM,
  author       = {Luca Masserano and
                  Abdul Fatir Ansari and
                  Boran Han and
                  Xiyuan Zhang and
                  Christos Faloutsos and
                  Michael W. Mahoney and
                  Andrew Gordon Wilson and
                  Youngsuk Park and
                  Syama Sundar Rangapuram and
                  Danielle C. Maddix and
                  Bernie Wang},
  title        = {Enhancing Foundation Models for Time Series Forecasting via Wavelet-based
                  Tokenization},
  booktitle    = {{ICML}},
  series       = {Proceedings of Machine Learning Research},
  publisher    = {{PMLR} / OpenReview.net},
  year         = {2025}
}

@inproceedings{Liang2024FoundationMF,
  author       = {Yuxuan Liang and
                  Haomin Wen and
                  Yuqi Nie and
                  Yushan Jiang and
                  Ming Jin and
                  Dongjin Song and
                  Shirui Pan and
                  Qingsong Wen},
  title        = {Foundation Models for Time Series Analysis: {A} Tutorial and Survey},
  booktitle    = {{KDD}},
  pages        = {6555--6565},
  publisher    = {{ACM}},
  year         = {2024}
}

@article{Meyer2024BenchmarkingTS,
  author       = {Marcel Meyer and
                  David Zapata Gonzalez and
                  Sascha Kaltenpoth and
                  Oliver M{\"{u}}ller},
  title        = {Benchmarking Time Series Foundation Models for Short-Term Household
                  Electricity Load Forecasting},
  journal      = {{IEEE} Access},
  volume       = {13},
  pages        = {218141--218153},
  year         = {2025}
}

@inproceedings{Zhang2024LargeLM,
  author       = {Xiyuan Zhang and
                  Ranak Roy Chowdhury and
                  Rajesh K. Gupta and
                  Jingbo Shang},
  title        = {Large Language Models for Time Series: {A} Survey},
  booktitle    = {{IJCAI}},
  pages        = {8335--8343},
  publisher    = {ijcai.org},
  year         = {2024}
}

@article{Liu2026ASO,
  author       = {Xingyue Liu and
                  Feizhong Zhou and
                  Hanguang Xiao and
                  Zhipeng Li and
                  Shuai Liu and
                  Lingling Qian},
  title        = {A survey On large language models for medical time series},
  journal      = {Expert Syst. Appl.},
  volume       = {311},
  pages        = {131364},
  year         = {2026}
}

@article{Dong2024CanLS,
  author       = {Manqing Dong and
                  Hao Huang and
                  Longbing Cao},
  title        = {Can LLMs Serve As Time Series Anomaly Detectors?},
  journal      = {CoRR},
  volume       = {abs/2408.03475},
  year         = {2024},
}

@inproceedings{Jiang2024EmpoweringTS,
  author       = {Yushan Jiang and
                  Zijie Pan and
                  Xikun Zhang and
                  Sahil Garg and
                  Anderson Schneider and
                  Yuriy Nevmyvaka and
                  Dongjin Song},
  title        = {Empowering Time Series Analysis with Large Language Models: {A} Survey},
  booktitle    = {{IJCAI}},
  pages        = {8095--8103},
  publisher    = {ijcai.org},
  year         = {2024}
}

@inproceedings{Xu2025CanML,
  author       = {Xiongxiao Xu and
                  Haoran Wang and
                  Yueqing Liang and
                  Philip S. Yu and
                  Yue Zhao and
                  Kai Shu},
  title        = {Can Multimodal LLMs Perform Time Series Anomaly Detection?},
  booktitle    = {{WWW}},
  pages        = {5392--5403},
  publisher    = {{ACM}},
  year         = {2026}
}

@inproceedings{Gruver2023LargeLM,
  author       = {Nate Gruver and
                  Marc Finzi and
                  Shikai Qiu and
                  Andrew Gordon Wilson},
  title        = {Large Language Models Are Zero-Shot Time Series Forecasters},
  booktitle    = {NeurIPS},
  year         = {2023}
}

@inproceedings{Liu2024LargeLM,
  author       = {Jun Liu and
                  Chaoyun Zhang and
                  Jiaxu Qian and
                  Minghua Ma and
                  Si Qin and
                  Chetan Bansal and
                  Qingwei Lin and
                  Saravan Rajmohan and
                  Dongmei Zhang},
  title        = {Large Language Models can Deliver Accurate and Interpretable Time
                  Series Anomaly Detection},
  booktitle    = {{KDD} {(2)}},
  pages        = {4623--4634},
  publisher    = {{ACM}},
  year         = {2025}
}

@inproceedings{RussellGilbert2024AADLLMAA,
  author       = {Alicia Russell{-}Gilbert and
                  Alexander Sommers and
                  Andrew Thompson and
                  Logan Cummins and
                  Sudip Mittal and
                  Shahram Rahimi and
                  Maria Seale and
                  Joseph Jabour and
                  Thomas Arnold and
                  Joshua Church},
  title        = {{AAD-LLM:} Adaptive Anomaly Detection Using Large Language Models},
  booktitle    = {{IEEE} Big Data},
  pages        = {4194--4203},
  publisher    = {{IEEE}},
  year         = {2024}
}

@article{Yu2025TriPLLMAT,
  author       = {Yuan{-}Cheng Yu and
                  Yen{-}Chieh Ouyang and
                  Chun{-}An Lin},
  title        = {TriP-LLM: {A} Tri-Branch Patch-Wise Large Language Model Framework
                  for Time-Series Anomaly Detection},
  journal      = {{IEEE} Access},
  volume       = {13},
  pages        = {168643--168653},
  year         = {2025}
}

@inproceedings{Alnegheimish2024CanLL,
  author       = {Sarah Alnegheimish and
                  Linh Nguyen and
                  Laure Berti{-}{\'{E}}quille and
                  Kalyan Veeramachaneni},
  title        = {Can Large Language Models be Anomaly Detectors for Time Series?},
  booktitle    = {{DSAA}},
  pages        = {1--10},
  publisher    = {{IEEE}},
  year         = {2024}
}

@inproceedings{He2025HarnessingVM,
  author       = {Zelin He and
                  Sarah Alnegheimish and
                  Matthew Reimherr},
  title        = {Harnessing Vision-Language Models for Time Series Anomaly Detection},
  booktitle    = {{AAAI}},
  pages        = {21690--21698},
  publisher    = {{AAAI} Press},
  year         = {2026}
}

@inproceedings{Liu2024LargeLMKD,
  author       = {Chen Liu and
                  Shibo He and
                  Qihang Zhou and
                  Shizhong Li and
                  Wenchao Meng},
  title        = {Large Language Model Guided Knowledge Distillation for Time Series
                  Anomaly Detection},
  booktitle    = {{IJCAI}},
  pages        = {2162--2170},
  publisher    = {ijcai.org},
  year         = {2024}
}

@article{Zhuang2024SeeIT,
  author       = {Jiaxin Zhuang and
                  Leon Yan and
                  Zhenwei Zhang and
                  Ruiqi Wang and
                  Jiawei Zhang and
                  Yuantao Gu},
  title        = {See it, Think it, Sorted: Large Multimodal Models are Few-shot Time
                  Series Anomaly Analyzers},
  journal      = {CoRR},
  volume       = {abs/2411.02465},
  year         = {2024}
}

@inproceedings{Tao2025MADLLMMA,
  author       = {Wei Tao and
                  Xiaoyang Qu and
                  Kai Lu and
                  Jiguang Wan and
                  Guokuan Li and
                  Jianzong Wang},
  title        = {{MADLLM:} Multivariate Anomaly Detection via Pre-trained LLMs},
  booktitle    = {{ICME}},
  pages        = {1--6},
  publisher    = {{IEEE}},
  year         = {2025}
}

@inproceedings{Hu2025ContextAlignmentAA,
  author       = {Yuxiao Hu and
                  Qian Li and
                  Dongxiao Zhang and
                  Jinyue Yan and
                  Yuntian Chen},
  title        = {Context-Alignment: Activating and Enhancing LLMs Capabilities in Time
                  Series},
  booktitle    = {{ICLR}},
  publisher    = {OpenReview.net},
  year         = {2025}
}

@inproceedings{Bian2024MultiPatchPA,
  author       = {Yuxuan Bian and
                  Xuan Ju and
                  Jiangtong Li and
                  Zhijian Xu and
                  Dawei Cheng and
                  Qiang Xu},
  title        = {Multi-Patch Prediction: Adapting Language Models for Time Series Representation
                  Learning},
  booktitle    = {{ICML}},
  series       = {Proceedings of Machine Learning Research},
  pages        = {3889--3912},
  publisher    = {{PMLR} / OpenReview.net},
  year         = {2024}
}

@inproceedings{Hermary2026ASTERLP,
  title={ASTER: Latent Pseudo-Anomaly Generation for Unsupervised Time-Series Anomaly Detection},
  author={Romain Hermary and Samet Hicsonmez and Dan Pineau and Abd El Rahman Shabayek and Djamila Aouada},
  year={2026},
  booktitle = {{International Conference on Pattern Recognition}},
}

@inproceedings{Chatzigeorgakidis2024MultiCastZM,
  author       = {Georgios Chatzigeorgakidis and
                  Konstantinos Lentzos and
                  Dimitrios Skoutas},
  title        = {MultiCast: Zero-Shot Multivariate Time Series Forecasting Using LLMs},
  booktitle    = {{ICDEW}},
  pages        = {119--127},
  publisher    = {{IEEE}},
  year         = {2024}
}

@article{Qiu2025TABUB,
  author       = {Xiangfei Qiu and
                  Zhe Li and
                  Wanghui Qiu and
                  Shiyan Hu and
                  Lekui Zhou and
                  Xingjian Wu and
                  Zhengyu Li and
                  Chenjuan Guo and
                  Aoying Zhou and
                  Zhenli Sheng and
                  Jilin Hu and
                  Christian S. Jensen and
                  Bin Yang},
  title        = {{TAB:} Unified Benchmarking of Time Series Anomaly Detection Methods},
  journal      = {Proc. {VLDB} Endow.},
  volume       = {18},
  number       = {9},
  pages        = {2775--2789},
  year         = {2025}
}

@article{Darban2022DeepLF,
  author       = {Zahra Zamanzadeh Darban and
                  Geoffrey I. Webb and
                  Shirui Pan and
                  Charu Aggarwal and
                  Mahsa Salehi},
  title        = {Deep Learning for Time Series Anomaly Detection: {A} Survey},
  journal      = {{ACM} Comput. Surv.},
  volume       = {57},
  number       = {1},
  pages        = {15:1--15:42},
  year         = {2025}
}

@inproceedings{Sun2023TESTTP,
  author       = {Chenxi Sun and
                  Hongyan Li and
                  Yaliang Li and
                  Shenda Hong},
  title        = {{TEST:} Text Prototype Aligned Embedding to Activate LLM's Ability
                  for Time Series},
  booktitle    = {{ICLR}},
  publisher    = {OpenReview.net},
  year         = {2024}
}

@inproceedings{Lu2022FrozenPT,
  author       = {Kevin Lu and
                  Aditya Grover and
                  Pieter Abbeel and
                  Igor Mordatch},
  title        = {Frozen Pretrained Transformers as Universal Computation Engines},
  booktitle    = {{AAAI}},
  pages        = {7628--7636},
  publisher    = {{AAAI} Press},
  year         = {2022}
}

@inproceedings{Wu2022TimesNetT2,
  author       = {Haixu Wu and
                  Tengge Hu and
                  Yong Liu and
                  Hang Zhou and
                  Jianmin Wang and
                  Mingsheng Long},
  title        = {TimesNet: Temporal 2D-Variation Modeling for General Time Series Analysis},
  booktitle    = {{ICLR}},
  publisher    = {OpenReview.net},
  year         = {2023}
}

@inproceedings{Zhou2023OneFA,
  author       = {Tian Zhou and
                  Peisong Niu and
                  Xue Wang and
                  Liang Sun and
                  Rong Jin},
  title        = {One Fits All: Power General Time Series Analysis by Pretrained {LM}},
  booktitle    = {NeurIPS},
  year         = {2023}
}

@inproceedings{Liu2008IsolationF,
  author       = {Fei Tony Liu and
                  Kai Ming Ting and
                  Zhi{-}Hua Zhou},
  title        = {Isolation Forest},
  booktitle    = {{ICDM}},
  pages        = {413--422},
  publisher    = {{IEEE} Computer Society},
  year         = {2008}
}

@inproceedings{Gao2024UniTSAU,
  author       = {Shanghua Gao and
                  Teddy Koker and
                  Owen Queen and
                  Tom Hartvigsen and
                  Theodoros Tsiligkaridis and
                  Marinka Zitnik},
  title        = {UniTS: {A} Unified Multi-Task Time Series Model},
  booktitle    = {NeurIPS},
  year         = {2024}
}

@article{Scholkopf2001EstimatingTS,
  author       = {Bernhard Sch{\"{o}}lkopf and
                  John C. Platt and
                  John Shawe{-}Taylor and
                  Alexander J. Smola and
                  Robert C. Williamson},
  title        = {Estimating the Support of a High-Dimensional Distribution},
  journal      = {Neural Comput.},
  volume       = {13},
  number       = {7},
  pages        = {1443--1471},
  year         = {2001}
}

@inproceedings{Hermary2025RemovingGB,
  author       = {Romain Hermary and
                  Vincent Gaudilli{\`{e}}re and
                  Abd El Rahman Shabayek and
                  Djamila Aouada},
  title        = {Removing Geometric Bias in One-Class Anomaly Detection with Adaptive
                  Feature Perturbation},
  booktitle    = {{WACV}},
  pages        = {6612--6622},
  publisher    = {{IEEE}},
  year         = {2025}
}

@article{Mejri2024UnsupervisedAD,
  author       = {Nesryne Mejri and
                  Laura Lopez{-}Fuentes and
                  Kankana Roy and
                  Pavel Chernakov and
                  Enjie Ghorbel and
                  Djamila Aouada},
  title        = {Unsupervised anomaly detection in time-series: An extensive evaluation
                  and analysis of state-of-the-art methods},
  journal      = {Expert Syst. Appl.},
  volume       = {256},
  pages        = {124922},
  year         = {2024}
}

@inproceedings{Goswami2024MOMENTAF,
  author       = {Mononito Goswami and
                  Konrad Szafer and
                  Arjun Choudhry and
                  Yifu Cai and
                  Shuo Li and
                  Artur Dubrawski},
  title        = {{MOMENT:} {A} Family of Open Time-series Foundation Models},
  booktitle    = {{ICML}},
  series       = {Proceedings of Machine Learning Research},
  volume       = {235},
  pages        = {16115--16152},
  publisher    = {{PMLR} / OpenReview.net},
  year         = {2024}
}

@inproceedings{Xu2021AnomalyTT,
  author       = {Jiehui Xu and
                  Haixu Wu and
                  Jianmin Wang and
                  Mingsheng Long},
  title        = {Anomaly Transformer: Time Series Anomaly Detection with Association
                  Discrepancy},
  booktitle    = {{ICLR}},
  publisher    = {OpenReview.net},
  year         = {2022}
}

@article{Wilcoxon1945IndividualCB,
  author  = {Wilcoxon, Frank},
  title   = {Individual Comparisons by Ranking Methods},
  journal = {Biometrics Bulletin},
  volume  = {1},
  number  = {6},
  pages   = {80--83},
  year    = {1945},
  doi     = {10.2307/3001968}
}

@article{Zhou2023LabelFreeMT,
  author       = {Qihang Zhou and
                  Shibo He and
                  Haoyu Liu and
                  Jiming Chen and
                  Wenchao Meng},
  title        = {Label-Free Multivariate Time Series Anomaly Detection},
  journal      = {{IEEE} Trans. Knowl. Data Eng.},
  volume       = {36},
  number       = {7},
  pages        = {3166--3179},
  year         = {2024}
}

@inproceedings{Gong2019MemorizingNT,
  author       = {Dong Gong and
                  Lingqiao Liu and
                  Vuong Le and
                  Budhaditya Saha and
                  Moussa Reda Mansour and
                  Svetha Venkatesh and
                  Anton van den Hengel},
  title        = {Memorizing Normality to Detect Anomaly: Memory-Augmented Deep Autoencoder
                  for Unsupervised Anomaly Detection},
  booktitle    = {{ICCV}},
  pages        = {1705--1714},
  publisher    = {{IEEE}},
  year         = {2019}
}

@article{Yoon2025MOMEMTOPM,
    title={Patch-based Memory Gate Model in Time Series Foundation Model},
    author={Samuel Yoon and Jongwon Kim and Juyoung Ha and Young Myoung Ko},
    journal={Transactions on Machine Learning Research},
    issn={2835-8856},
    year={2026},
    note={}
}

@article{Wu2023CurrentTS,
  author       = {Renjie Wu and
                  Eamonn J. Keogh},
  title        = {Current Time Series Anomaly Detection Benchmarks are Flawed and are
                  Creating the Illusion of Progress},
  journal      = {{IEEE} Trans. Knowl. Data Eng.},
  volume       = {35},
  number       = {3},
  pages        = {2421--2429},
  year         = {2023}
}
\bibliographystyle{tmlr}

\clearpage
\appendices
\section{Parameters}
\subsection{PCA \& Isolation Forest}

We used the Principle Component Analysis (PCA) algorithm to reduce all analysed vectors to the same dimensionality, before fitting the Isolation Forest.
We set the target dimension to 512, and retained on average close to 98\% of explained variance; we thus reduced the complexity of the study without compromising the results.

We set the number of trees in the Isolation Forest to 100.

\subsection{Trained Baselines}

We trained three TSFM baselines, TimesNet~\citep{Wu2022TimesNetT2},  GPT4TS~\citep{Zhou2023OneFA} and AnomalyTransformer~\citep{Xu2021AnomalyTT}, on four anomaly detection datasets and presented the results in the main paper.
All of them were trained on the full train sets, with parameters matching the published configurations, over 5 different seeds, with a window size set to a 100 time-steps.
Both GPT4TS and TimesNet were trained for 10 epochs, while 3 were done for AnomalyTransoformer.
The \texttt{projection depth} (or \texttt{d\_model}) parameter of the models are reported in Table~\ref{tab:llms}.

Reported results follow the same evaluation protocol with a stride of 128.
The models have open-source implementations, and we included them in our codebase to ensure that all experiments are done in the same environment.

\subsection{Pre-Trained Baselines}
A window size of 128 was used for MOMENT~\citep{Goswami2024MOMENTAF} and Chronos~\citep{Ansari2025Chronos2FU}.
Only one evaluation with a fixed seed (42) was required, since there is no randomised process.
Reported results follow the same evaluation protocol with a stride of 128.

\subsection{Environment}

\textit{Code, requirements, and the Docker image will be made publicly available.} Experiments were conducted on NVIDIA RTX A6000 (550.163.01) GPUs and an AMD EPYC 7742 64-Core processor.

\section{Models Details}

Table~\ref{tab:llms} summarises the model details.
As mentioned in the previous section, TimesNet and GPT4TS where reimplemented.
For all other models in the study, we took advantage of the HuggingFace platform~\citep{Wolf2019HuggingFacesTS} and its vast amount of shared architectures and pre-trained weights.
The versions used are also depicted in the table.
 
\section{Datasets \& Evaluation Setting}

\begin{table*}[t]
    \centering
    \caption{Details of the models used in the study.}
    \label{tab:llms}
    \resizebox{.95\linewidth}{!}{
\renewcommand{\arraystretch}{1.2}%
\setlength{\tabcolsep}{9pt}%
\begin{tabular}{cccccc}
\toprule
 \multirow{2}{*}{\textbf{Architecture}}
 & \multirow{2}{*}{\shortstack{\textbf{HuggingFace}~\citep{Wolf2019HuggingFacesTS}\\\textbf{Version}}}
 & \multirow{2}{*}{\textbf{\shortstack{Number of \\Parameters}}}
 & \multirow{2}{*}{\shortstack{\textbf{Embedding}\\\textbf{Size}}}
 & \multirow{2}{*}{\textbf{\shortstack{Multimodal\\Training}}}
 & \multirow{2}{*}{\shortstack{\textbf{Vocabulary}\\\textbf{Size}}} \\
 \\
\midrule\midrule
AnomalyTransformer~\citep{Xu2021AnomalyTT} &-& 5,002,387 &512& \XSolidBrush&-\\
GPT4TS~\citep{Zhou2023OneFA}& - & $\approx\,$81,960,828& 768 &\XSolidBrush&-\\
TimesNet~\citep{Wu2022TimesNetT2} & - & $\approx\,$28,150,972 & 
128 &\XSolidBrush&-\\
Chronos~\citep{Ansari2025Chronos2FU} & \texttt{amazon/chronos-2} & 119,477,664 & $D\times P \times 768$\textsuperscript{$\dagger$} &\XSolidBrush&-\\
 MOMENT (Small)~\citep{Goswami2024MOMENTAF} & \texttt{AutonLab/MOMENT-1-small} & 35,341,512 & 512 &\XSolidBrush&-\\
 MOMENT (Base)~\citep{Goswami2024MOMENTAF} & \texttt{AutonLab/MOMENT-1-base} & 109,641,608 & 768 &\XSolidBrush&-\\
 MOMENT (Large)~\citep{Goswami2024MOMENTAF} & \texttt{AutonLab/MOMENT-1-large} & 341,248,520 & 1024 &\XSolidBrush&-\\
 GPT~\citep{Radford2019LanguageMA} & \texttt{openai-community/gpt2} & 124,439,808 & 768 &\XSolidBrush&50257\\
 Qwen~\citep{Yang2024Qwen2TR} & \texttt{Qwen/Qwen2-0.5B} & 494,032,768 & 896 &\XSolidBrush&151936\\
 Qwen~(7B)~\citep{Yang2024Qwen2TR} & \texttt{Qwen/Qwen2-7B} & 7,070,619,136 & 3584 & \XSolidBrush&152064\\
 Qwen-VL~(7B)~\citep{Wang2024Qwen2VLEV}& \texttt{Qwen/Qwen2-VL-7B} & 7,070,619,136 & 3584 & \Checkmark&152064\\
 Mistral~\citep{Jiang2023Mistral7}& \texttt{mistralai/Mistral-7B-v0.1} & 
7,110,660,096 & 3072 &\XSolidBrush&32000\\
 Phi~\citep{Abdin2024Phi3TR} & \texttt{microsoft/Phi-3-mini-4k-instruct} & 3,722,578,944 & 4096 &\XSolidBrush&32064\\
 \bottomrule
 
\multicolumn{6}{p{1.5\linewidth}}{\textsuperscript{$\dagger$} When used as feature extractor, Chronos has an output size that depends on the number dimension of the dataset, $D$, and the number of patches $P$ which defaults to $L/16$ ($L$ being the window size).}
\end{tabular}
}
\end{table*}

\begin{table*}[t]
    \centering
    \caption{Detailed statistics of the time-series anomaly detection datasets.}
    \label{tab:datasets}
    \resizebox{.95\linewidth}{!}{
\renewcommand{\arraystretch}{1.2}%
\setlength{\tabcolsep}{9pt}%
\begin{tabular}{ccccccccc}
\toprule
\multirow{3}{*}{\textbf{Dataset}} & \multirow{3}{*}{\textbf{Features}} & \multirow{3}{*}{\textbf{Train Set}} & \multirow{3}{*}{\textbf{Test Set}} & \multirow{3}{*}{\textbf{Anomaly \%}} & \multicolumn{4}{c}{\textbf{Anomalous Sequences}}\\
\cline{6-9}
& & & & & \multicolumn{3}{c}{Length} & \multirow{2}{*}{Num.} \\
\cline{6-8}
& & & && Min. & Max. & Avg. &\\
\midrule\midrule
{SWaT}~\citep{Abdulaal2021PracticalAT} & {51} & {495,000} & {449,919} &  {12.14}  &{101}&{35,900}&{1,560.60} & {35}\\

{WaDi}~\citep{Ahmed2017WADIAW}  & {123} & {784,537} & {172,801} & {5.77} &{88}&{1,741}&{712.64} & {14} \\

{PUMP}~\citep{Feng2021TimeSA} & {44} & {76,901} & {143,401} &  {10.05}   &{42}&{8,391}&{2,401.33}&{6}\\

{PSM}~\citep{Abdulaal2021PracticalAT}& {25}  & {132,481} & {87,841} & {27.76} &{1}&{8,861}&{338.625}&{72}\\
\bottomrule
\end{tabular}

}
\end{table*}

\begin{table*}[t]
    \centering
    \caption{Effective number of samples and anomaly ratio changes with the application of maximum value on the window targets and strides.}
    \label{tab:datasets_stride}
    \resizebox{.7\linewidth}{!}{
\newcommand{\gc}[1]{\cellcolor{gray!20}#1}%
\renewcommand{\arraystretch}{1.2}%
\setlength{\tabcolsep}{9pt}
\newlength{\rowheight}%
\setlength{\rowheight}{\fpeval{2.5 * \arraystretch}ex}%
\newcommand{\vcline}[1]{\multirow{1}{*}{\rule{.1pt}{\rowheight * #1}}}%
\begin{tabular}{ccccccccccc}
\toprule
\multirow{2}{*}{\textbf{Dataset}} & \multirow{2}{*}{\textbf{Stride}}& \multirow{2}{*}{\textbf{\shortstack{Test Size\\(\# Windows)}}} & \multicolumn{7}{c}{\textbf{Anomaly \% \wrt{} Window Size}} \\
\cline{4-10}
&& &1&4&8&16&32&64&128\\
\midrule\midrule
\multirow{7}{*}{PSM}&1&87841&\gc{27.76}&28.00&28.31&28.87&29.91&31.71&35.06\\
&4&21961&\vcline{5}&27.99&\vcline{1}&\vcline{2}&\vcline{3}&\vcline{4}&\vcline{5}\\
&8&10981&&\vcline{4}&28.37&&&&\\
&16&5491&&&\vcline{3}&28.88&&&\\
&32&2746&&&&\vcline{2}&30.01&&\\
&64&1373&&&&&\vcline{1}&31.76&\\
&128&687&27.51&27.80&28.38&28.82&30.28&32.61&35.23\\
\midrule
\multirow{7}{*}{PUMP}&1&143401&\gc{10.05}&10.06&10.08&10.11&10.18&10.31&10.58\\
&4&35851&\vcline{5}&10.06&\vcline{1}&\vcline{2}&\vcline{3}&\vcline{4}&\vcline{5}\\
&8&17926&&\vcline{4}&10.08&&&&\\
&16&8963&&&\vcline{3}&10.12&&&\\
&32&4482&&&&\vcline{2}&10.17&&\\
&64&2241&&&&&\vcline{1}&10.22&\\
&128&1121&9.90&9.90&9.90&9.99&10.26&10.26&10.44\\
\midrule
\multirow{7}{*}{SWaT}&1&449919&\gc{12.14}&12.16&12.19&12.26&12.38&12.63&13.10\\
&4&112480&\vcline{5}&12.17&\vcline{1}&\vcline{2}&\vcline{3}&\vcline{4}&\vcline{5}\\
&8&56240&&\vcline{4}&12.20&&&&\\
&16&28120&&&\vcline{3}&12.27&&&\\
&32&14060&&&&\vcline{2}&12.40&&\\
&64&7030&&&&&\vcline{1}&12.66&\\
&128&3515&12.23&12.23&12.26&12.29&12.35&12.55&13.14\\
\midrule
\multirow{7}{*}{WaDi}&1&172801&\gc{5.77}&5.80&5.83&5.90&6.02&6.28&6.80\\
&4&43201&\vcline{5}&5.79&\vcline{1}&\vcline{2}&\vcline{3}&\vcline{4}&\vcline{5}\\
&8&21601&&\vcline{4}&5.83&&&&\\
&16&10801&&&\vcline{3}&5.87&&&\\
&32&5401&&&&\vcline{2}&5.96&&\\
&64&2701&&&&&\vcline{1}&6.26&\\
&128&1351&5.85&5.92&6.00&6.00&6.14&6.37&6.88\\
\bottomrule
\end{tabular}
}
\end{table*}

\begin{table*}[t]
    \centering
    \caption{Effective number of samples and anomalous time-steps window coverage changes with different strides.}
    \label{tab:datasets_stride_coverage}
    \resizebox{.7\linewidth}{!}{
\newcommand{\gc}[1]{\cellcolor{gray!20}#1}%
\renewcommand{\arraystretch}{1.2}%
\setlength{\tabcolsep}{9pt}
\ifdefined\rowheight\else\newlength{\rowheight}\fi
\setlength{\rowheight}{\fpeval{2.5 * \arraystretch}ex}%
\newcommand{\vcline}[1]{\multirow{1}{*}{\rule{.1pt}{\rowheight * #1}}}%
\begin{tabular}{ccccccccccc}
\toprule
\multirow{2}{*}{\textbf{Dataset}} & \multirow{2}{*}{\textbf{Stride}}& \multirow{2}{*}{\textbf{\shortstack{Test Size\\(\# Windows)}}} & \multicolumn{7}{c}{\textbf{Anomaly Window Coverage (\%) \wrt{} Window Size}} \\
\cline{4-10}
&& &1&4&8&16&32&64&128\\
\midrule\midrule
\multirow{7}{*}{PSM}&1&87841&\gc{100.0}&99.12&98.05&96.09&92.77&87.45&79.00\\
&4&21961&\vcline{5}&99.17&\vcline{1}&\vcline{2}&\vcline{3}&\vcline{4}&\vcline{5}\\
&8&10981&&\vcline{4}&97.85&&&&\\
&16&5491&&&\vcline{3}&96.09&&&\\
&32&2746&&&&\vcline{2}&92.48&&\\
&64&1373&&&&&\vcline{1}&87.30&\\
&128&687&100.0&99.21&97.17&95.21&90.77&85.06&78.61\\
\midrule
\multirow{7}{*}{PUMP}&1&143401&\gc{100.0}&99.85&99.71&99.37&98.73&97.46&94.97\\
&4&35851&\vcline{5}&99.90&\vcline{1}&\vcline{2}&\vcline{3}&\vcline{4}&\vcline{5}\\
&8&17926&&\vcline{4}&99.66&&&&\\
&16&8963&&&\vcline{3}&99.27&&&\\
&32&4482&&&&\vcline{2}&98.73&&\\
&64&2241&&&&&\vcline{1}&98.29&\\
&128&1121&100.0&100.0&100.0&99.32&97.71&98.24&96.19\\
\midrule
\multirow{7}{*}{SWaT}&1&449919&\gc{100.0}&99.80&99.56&99.07&98.05&96.14&92.68\\
&4&112480&\vcline{5}&99.80&\vcline{1}&\vcline{2}&\vcline{3}&\vcline{4}&\vcline{5}\\
&8&56240&&\vcline{4}&99.51&&&&\\
&16&28120&&&\vcline{3}&98.93&&&\\
&32&14060&&&&\vcline{2}&97.85&&\\
&64&7030&&&&&\vcline{1}&95.90&\\
&128&3515&100.0&100.0&99.66&99.41&98.58&96.68&92.38\\
\midrule
\multirow{7}{*}{WaDi}&1&172801&\gc{100.0}&99.56&99.02&97.95&95.85&91.89&84.86\\
&4&43201&\vcline{5}&99.66&\vcline{1}&\vcline{2}&\vcline{3}&\vcline{4}&\vcline{5}\\
&8&21601&&\vcline{4}&99.07&&&&\\
&16&10801&&&\vcline{3}&98.34&&&\\
&32&5401&&&&\vcline{2}&96.83&&\\
&64&2701&&&&&\vcline{1}&92.24&\\
&128&1351&100.0&99.71&99.07&98.93&95.56&90.58&83.79\\
\bottomrule
\end{tabular}
}
\end{table*}

\begin{table*}[t]
    \centering
    \caption{Effective number of samples and anomalous segments window centroids changes with different strides.}
    \label{tab:datasets_stride_centroid}
    \resizebox{.7\linewidth}{!}{
\newcommand{\gc}[1]{\cellcolor{gray!20}#1}%
\renewcommand{\arraystretch}{1.2}%
\setlength{\tabcolsep}{9pt}
\ifdefined\rowheight\else\newlength{\rowheight}\fi
\setlength{\rowheight}{\fpeval{2.5 * \arraystretch}ex}%
\newcommand{\vcline}[1]{\multirow{1}{*}{\rule{.1pt}{\rowheight * #1}}}%
\begin{tabular}{ccccccccccc}
\toprule
\multirow{2}{*}{\textbf{Dataset}} & \multirow{2}{*}{\textbf{Stride}}& \multirow{2}{*}{\textbf{\shortstack{Test Size\\(\# Windows)}}} & \multicolumn{7}{c}{\textbf{Anomaly Segment Centroid (\% from left to right) }} \\
\cline{4-10}
&& &1&4&8&16&32&64&128\\
\midrule\midrule
\multirow{7}{*}{PSM}&1&87841&\gc{00.00}&50.00&50.00&50.01&50.01&50.05&50.06\\
&4&21961&\vcline{5}&49.94&\vcline{1}&\vcline{2}&\vcline{3}&\vcline{4}&\vcline{5}\\
&8&10981&&\vcline{4}&49.98&&&&\\
&16&5491&&&\vcline{3}&50.11&&&\\
&32&2746&&&&\vcline{2}&50.16&&\\
&64&1373&&&&&\vcline{1}&50.01&\\
&128&687&00.00&50.17&49.78&50.54&49.87&49.53&51.30\\
\midrule
\multirow{7}{*}{PUMP}&1&143401&\gc{00.00}&50.00&50.00&50.00&50.00&50.00&50.00\\
&4&35851&\vcline{5}&50.02&\vcline{1}&\vcline{2}&\vcline{3}&\vcline{4}&\vcline{5}\\
&8&17926&&\vcline{4}&50.02&&&&\\
&16&8963&&&\vcline{3}&49.99&&&\\
&32&4482&&&&\vcline{2}&49.93&&\\
&64&2241&&&&&\vcline{1}&50.15&\\
&128&1121&00.00&50.00&50.00&49.64&48.82&49.72&50.36\\
\midrule
\multirow{7}{*}{SWaT}&1&449919&\gc{00.00}&50.00&50.00&50.00&50.00&50.00&50.00\\
&4&112480&\vcline{5}&50.01&\vcline{1}&\vcline{2}&\vcline{3}&\vcline{4}&\vcline{5}\\
&8&56240&&\vcline{4}&50.03&&&&\\
&16&28120&&&\vcline{3}&50.01&&&\\
&32&14060&&&&\vcline{2}&50.02&&\\
&64&7030&&&&&\vcline{1}&50.05&\\
&128&3515&00.00&50.00&50.02&50.08&50.28&50.18&49.66\\
\midrule
\multirow{7}{*}{WaDi}&1&172801&\gc{00.00}&50.00&50.00&50.00&50.00&50.00&50.00\\
&4&43201&\vcline{5}&49.99&\vcline{1}&\vcline{2}&\vcline{3}&\vcline{4}&\vcline{5}\\
&8&21601&&\vcline{4}&50.06&&&&\\
&16&10801&&&\vcline{3}&49.93&&&\\
&32&5401&&&&\vcline{2}&49.77&&\\
&64&2701&&&&&\vcline{1}&50.01&\\
&128&1351&00.00&49.79&49.47&50.08&50.58&51.16&50.89\\
\bottomrule
\end{tabular}
}
\end{table*}

\begin{table*}[t]
    \centering
    \caption{Complementary results with a stride of 1.}
    \label{tab:stride1}
    \resizebox{.95\linewidth}{!}{
\renewcommand{\arraystretch}{1.2}%

\begin{tabular}{ccccccccccccc}
\toprule
\multirow[c]{2}{*}{\textbf{Model}} & \multicolumn{3}{c}{\textbf{PSM}} & \multicolumn{3}{c}{\textbf{PUMP}} & \multicolumn{3}{c}{\textbf{SWaT}} & \multicolumn{3}{c}{\textbf{WaDi}} \\
\cline{2-13}
 & \textbf{AUPR} & \textbf{AUROC} & \textbf{F$_1$} & \textbf{AUPR} & \textbf{AUROC} & \textbf{F$_1$} & \textbf{AUPR} & \textbf{AUROC} & \textbf{F$_1$} & \textbf{AUPR} & \textbf{AUROC} & \textbf{F$_1$} \\
\midrule\midrule
AnomalyTransformer & 68.25 & 78.76 & 63.98 & 21.98 & 79.65 & 48.90 & 69.13 & 80.83 & 72.44 & 29.40 & 52.43 & 14.83 \\
  GPT4TS & 52.87 & 61.68 & 49.18 & 07.64 & 20.79 & 21.87 & 10.77 & 21.95 & 10.19 & 40.91 & 55.17 & 14.17 \\
  TimesNet & 52.25 & 61.44 & 48.74 & 07.63 & 20.85 & 21.89 & 10.60 & 21.77 & 10.04 & 40.87 & 55.14 & 14.19 \\
\midrule
MOMENT (Small) & 52.53 & 64.94 & 45.35 & 12.43 & 43.94 & 20.43 & 12.19 & 27.95 & 10.57 & 19.05 & 51.97 & 14.02 \\
 MOMENT (Base) & 66.81 & 70.80 & 54.92 & 15.33 & 44.56 & 26.28 & 12.18 & 28.03 & 05.09 & 18.48 & 52.11 & 14.08 \\
 MOMENT (Large) & 66.36 & 70.68 & 54.81 & 13.85 & 51.08 & 24.05 & 08.88 & 24.22 & 02.91 & 07.97 & 49.03 & 11.27 \\
 \midrule
 
GPT & 33.19 & 54.09 & 33.17 & 15.80 & 65.01 & 29.85 &-&-&-&-&-&-\\
Qwen & 50.81 & 67.08 & 49.60 & 17.47 & 71.14 & 35.85 &-&-&-&-&-&-\\
\bottomrule
\end{tabular}
}
\end{table*}

Detailed statistics of the 4 datasets used in this study can be found in Table~\ref{tab:datasets}.
As mentioned in the main paper, our evaluation uses a target modification strategy ($y_t = \max\left(\left\{y_i,\,\forall i \in \llbracket t -L + 1,\,t\rrbracket\right\}\right)$) that increase the ratio of anomalous windows.

We also use a stride of 128 throughout the whole study, to keep the same number of samples and increase the throughput of results.
We report the effective ratio changes in Table~\ref{tab:datasets_stride}.
The original ratios of the datasets are greyed out and correspond to both a window size and a stride of 1.
The effective ratios from this study are the last row of each datasets, window sizes from 4 to 128.
Although these modifications can change the overall statistics of the datasets anomalies, we observed no significant difference with an evaluation on the complete dataset.
We thus consider this scale-reduced setting representative of the full evaluation case, in terms of model comparisons.
To quantitatively support this simplified protocol, particularly analysing the anomaly distributions over the windows, we provide two complementary statistics: coverage (Table~\ref{tab:datasets_stride_coverage}) and centroid (Table~\ref{tab:datasets_stride_centroid}).
Coverage is the amount of anomalous time-steps in the window, while centroid is the average temporal position of the anomalous segment within the window.
We observe almost no change compared to the windows statistics over the full test set, which supports an equivalence between the protocols.
Additionally, we provide results of 6 baselines and 2 LLMs with a stride of 1 (full test set) in Table~\ref{tab:stride1}.
Results are similar to a stride of 128, used in the main study, and ranking is preserved between models.

\section{Complementary Results}

\subsection{Main Figures: Other Metrics}

\begin{figure}[t]
    \centering
    \begin{subfigure}{0.49\linewidth}
        \centering
        \includegraphics[width=\linewidth]{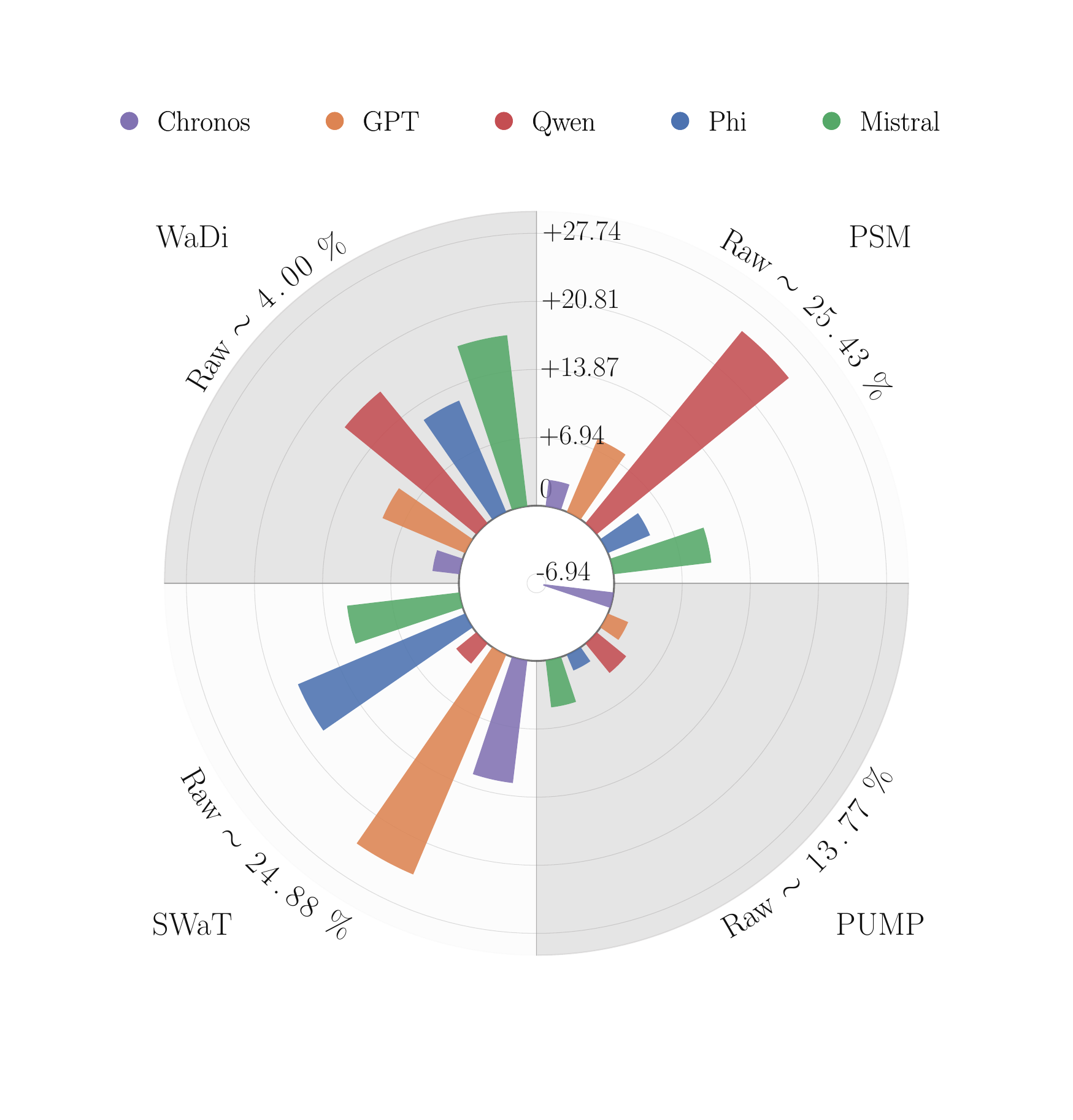}
        \caption{AUPR}
        \label{fig:results_datasets_aupr}
    \end{subfigure}
    \hfill
    \begin{subfigure}{0.49\linewidth}
        \centering
        \includegraphics[width=\linewidth]{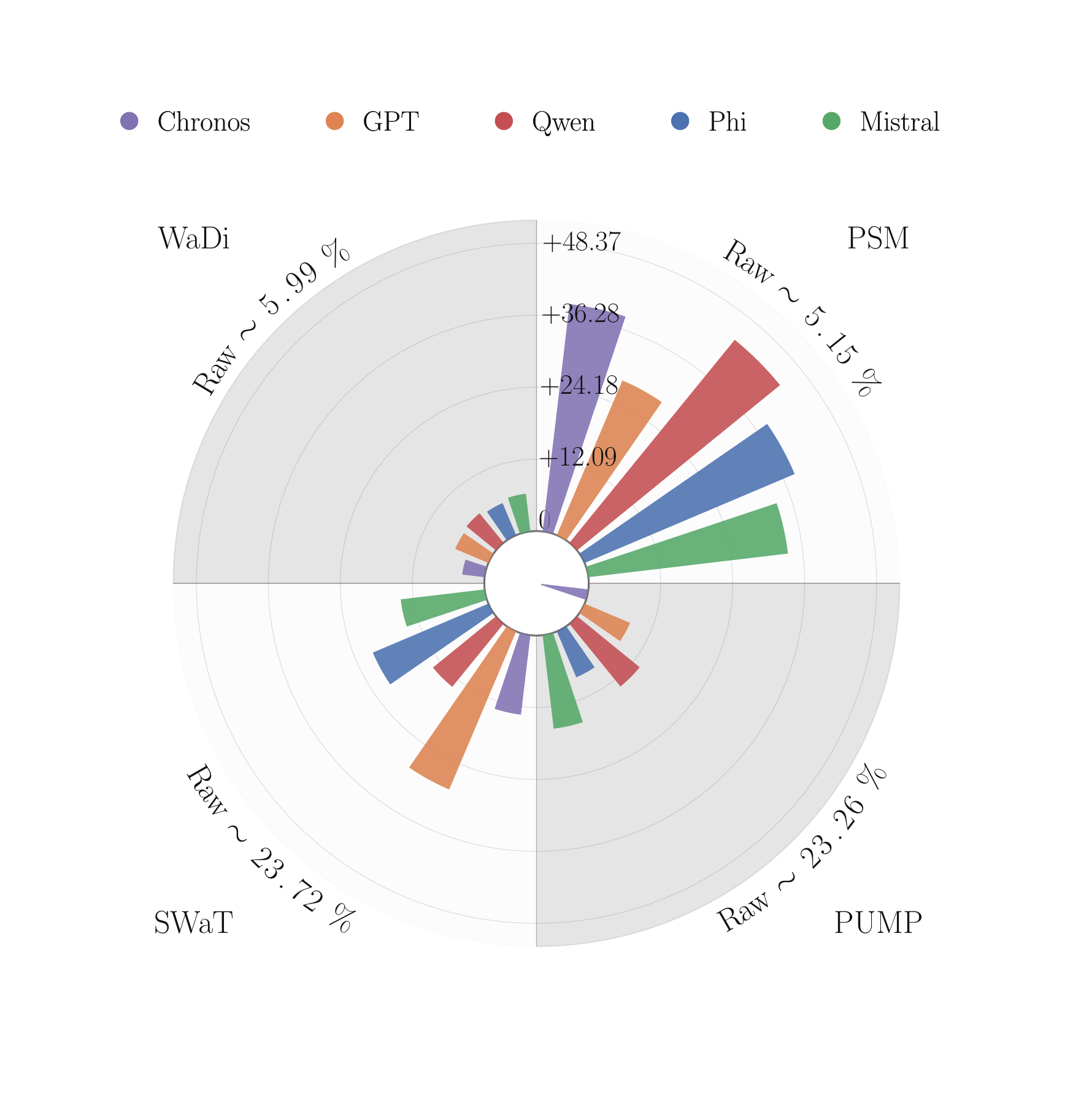}
        \caption{$F_1$}
        \label{fig:results_datasets_f1}
    \end{subfigure}
    \caption{Results of the studied LLMs on different datasets, with Interpolation.}
    \label{fig:results_datasets_pool}
\end{figure}

\begin{figure}[t]
    \centering
    \begin{subfigure}{0.49\linewidth}
        \centering
        \includegraphics[width=\linewidth]{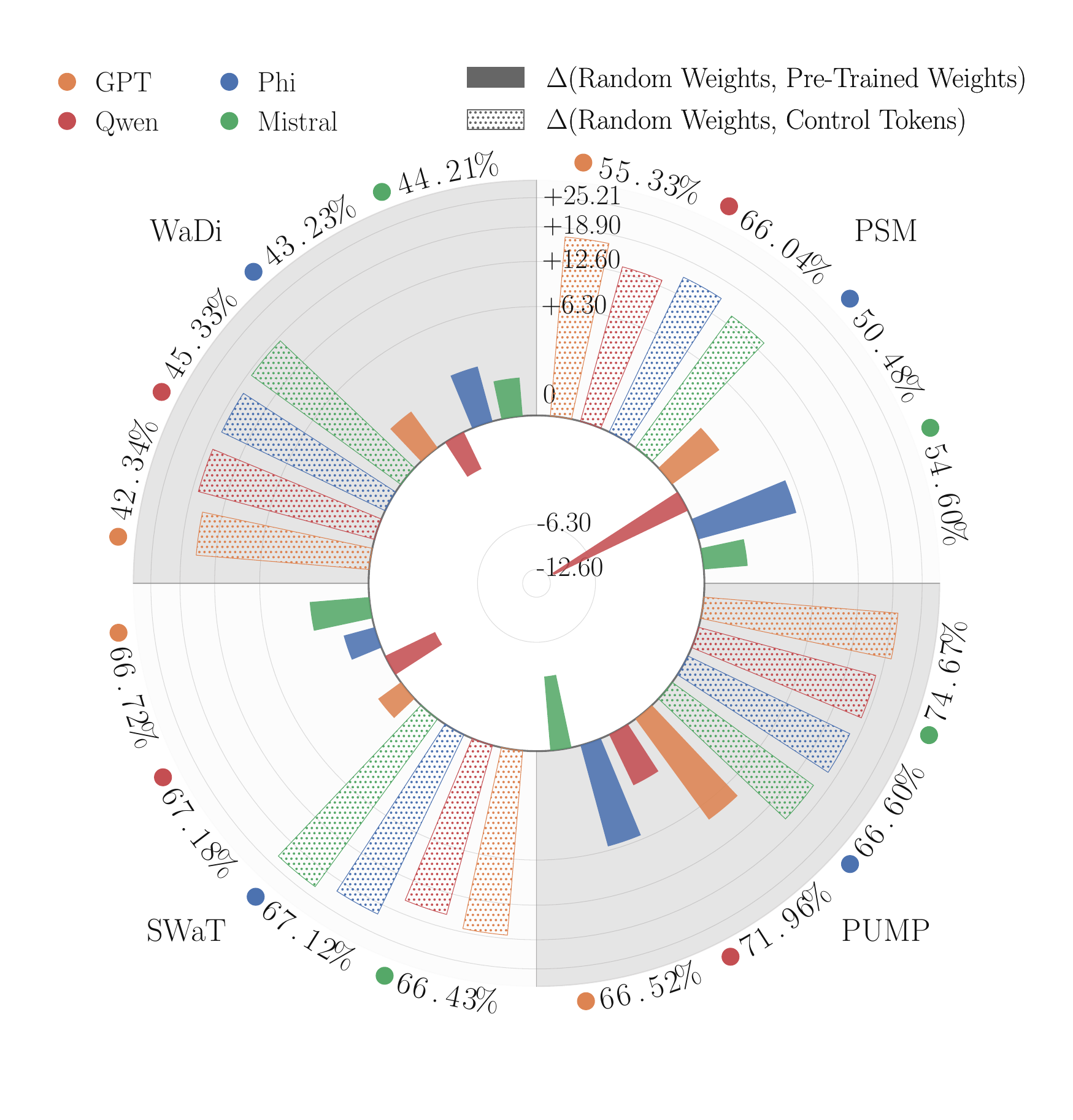}
        \caption{AUROC}
        \label{fig:randominit_auroc}
    \end{subfigure}
    \hfill
    \begin{subfigure}{0.49\linewidth}
        \centering
        \includegraphics[width=\linewidth]{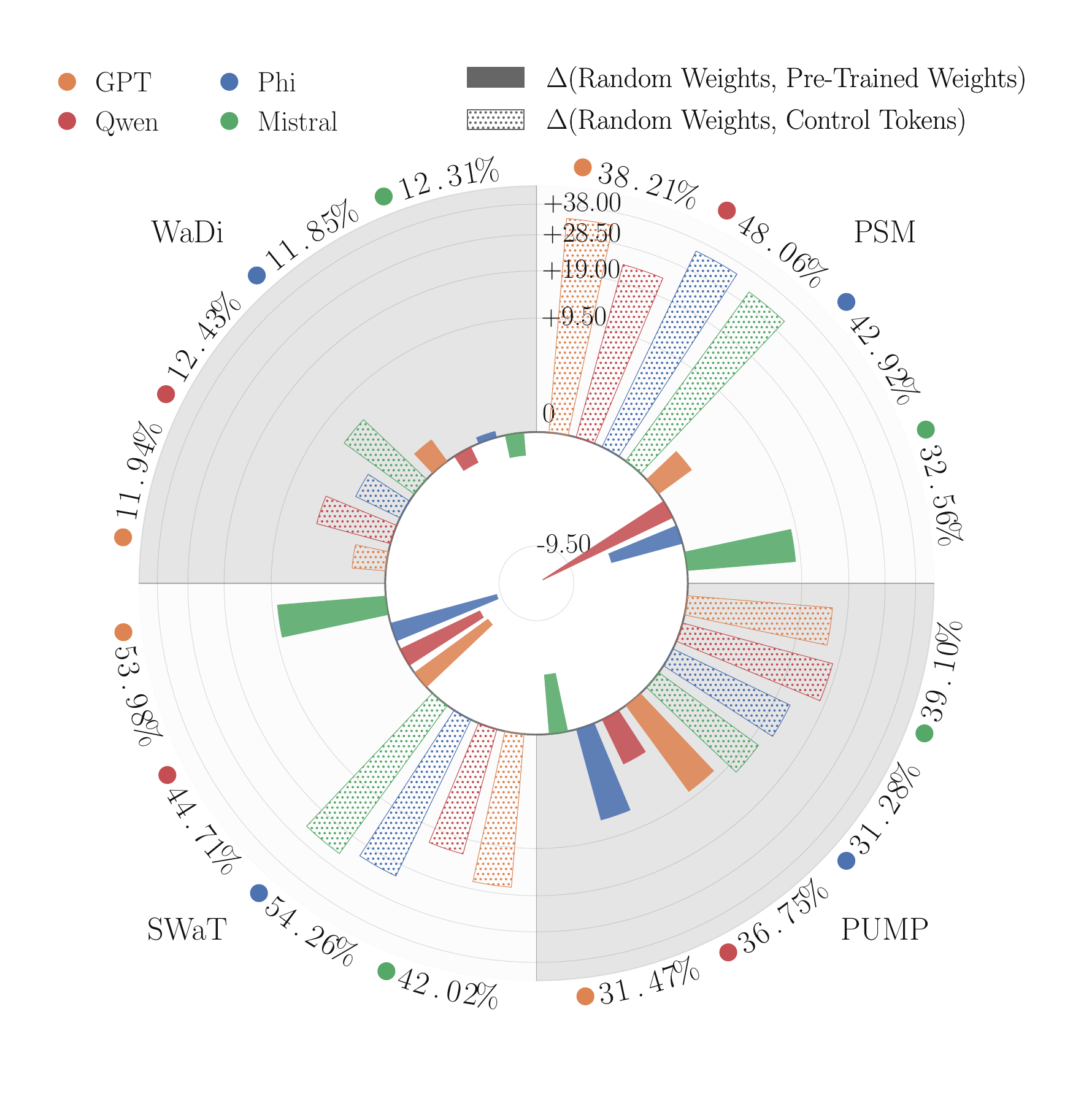}
        \caption{F$_1$}
        \label{fig:randominit_f1}
    \end{subfigure}
    \caption{Study on the impact of inductive bias and knowledge transfer.}
    \label{fig:randominit}
\end{figure}

\begin{figure}
    \centering
    \includegraphics[width=1\linewidth]{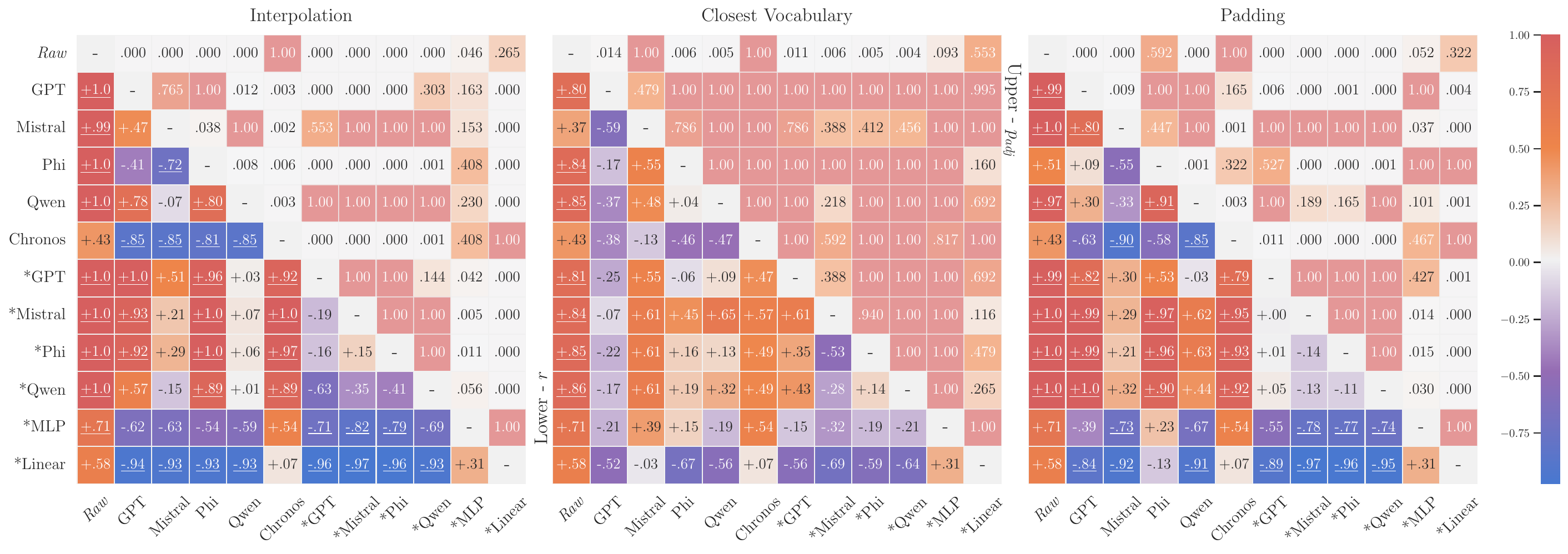}
    \caption{Wilcoxon test study, with Closest Vocabulary tokenisation results.}
    \label{fig:wilcoxon-cv}
\end{figure}

We complement the figures of the main paper focusing on one metric with the two others considered in this study.
Figure~\ref{fig:results_datasets_aupr} and
Figure~\ref{fig:results_datasets_f1} present AUPR and F$_1$ scores for the overall results over the different datasets.
Figure~\ref{fig:randominit_auroc} and
Figure~\ref{fig:randominit_f1} present AUROC and F$_1$ scores for the study on the impact of inductive bias and knowledge transfer.
Figure~\ref{fig:wilcoxon-cv} provides the results of the statistical analysis for the Closest Vocabulary tokenisation, along Interpolation and Padding from the main article.

\subsection{Detailed Results}

The raw results used to generate the main study of pre-trained LLMs, per window size and with standard deviations, can be found from Table~\ref{tab:results-cv} to Table~\ref{tab:results-pool}.
The \textit{Raw} and \textit{Control} results are in Table~\ref{tab:results-raw} to Table~\ref{tab:results-control-pool}.
The results for randomised models are in Table~\ref{tab:results-random-mlp} to Table~\ref{tab:results-random-pool}.

\begin{figure*}[ht]
    \centering
    \includegraphics[width=1\linewidth]{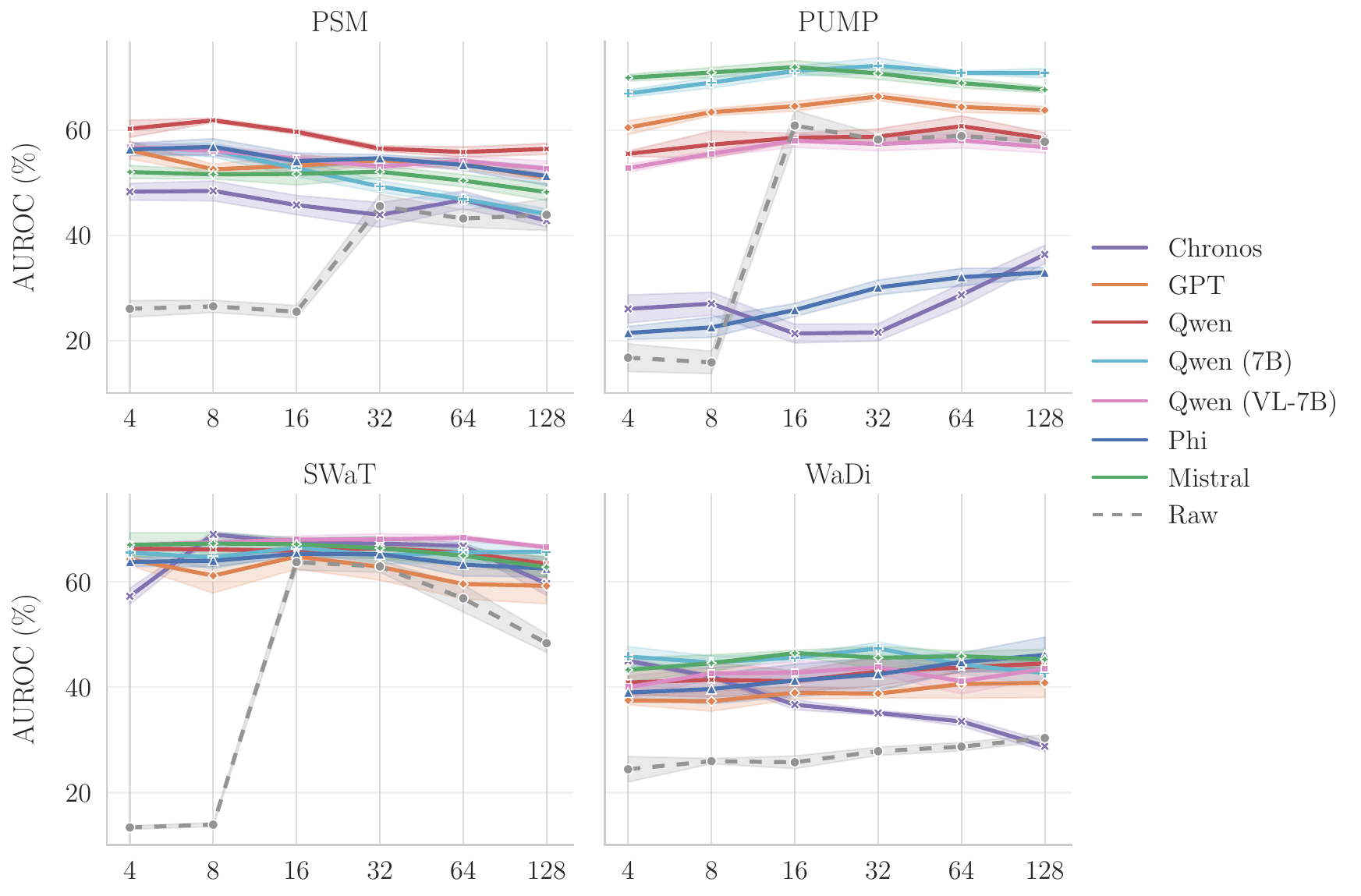}
    \caption{AUROC Padding}
    \label{fig:aurocpad}
\end{figure*}

\begin{figure*}[ht]
    \centering
    \includegraphics[width=1\linewidth]{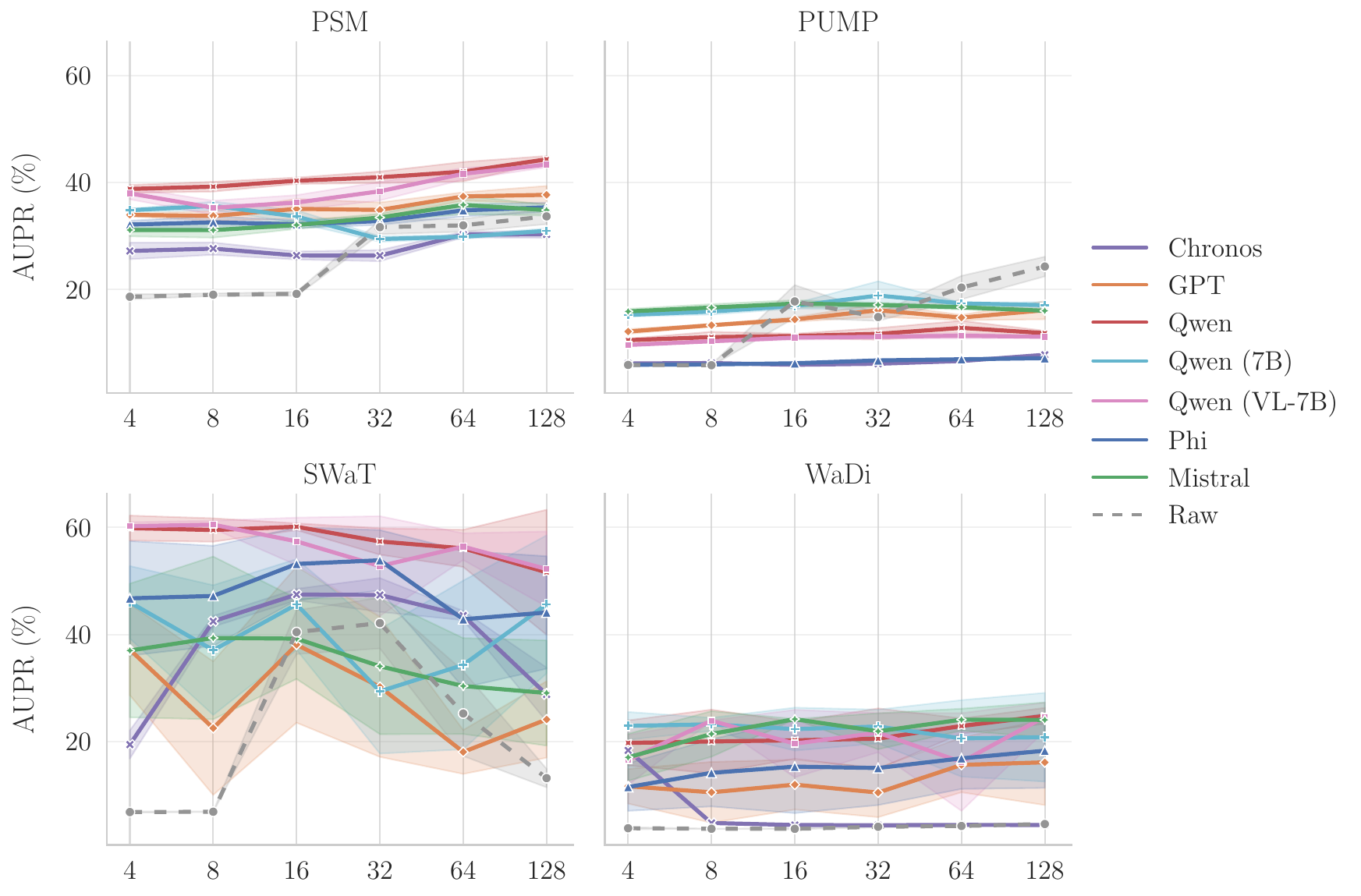}
    \caption{AUPR Padding}
    \label{fig:auprpad}
\end{figure*}

\begin{figure*}[ht]
    \centering
    \includegraphics[width=1\linewidth]{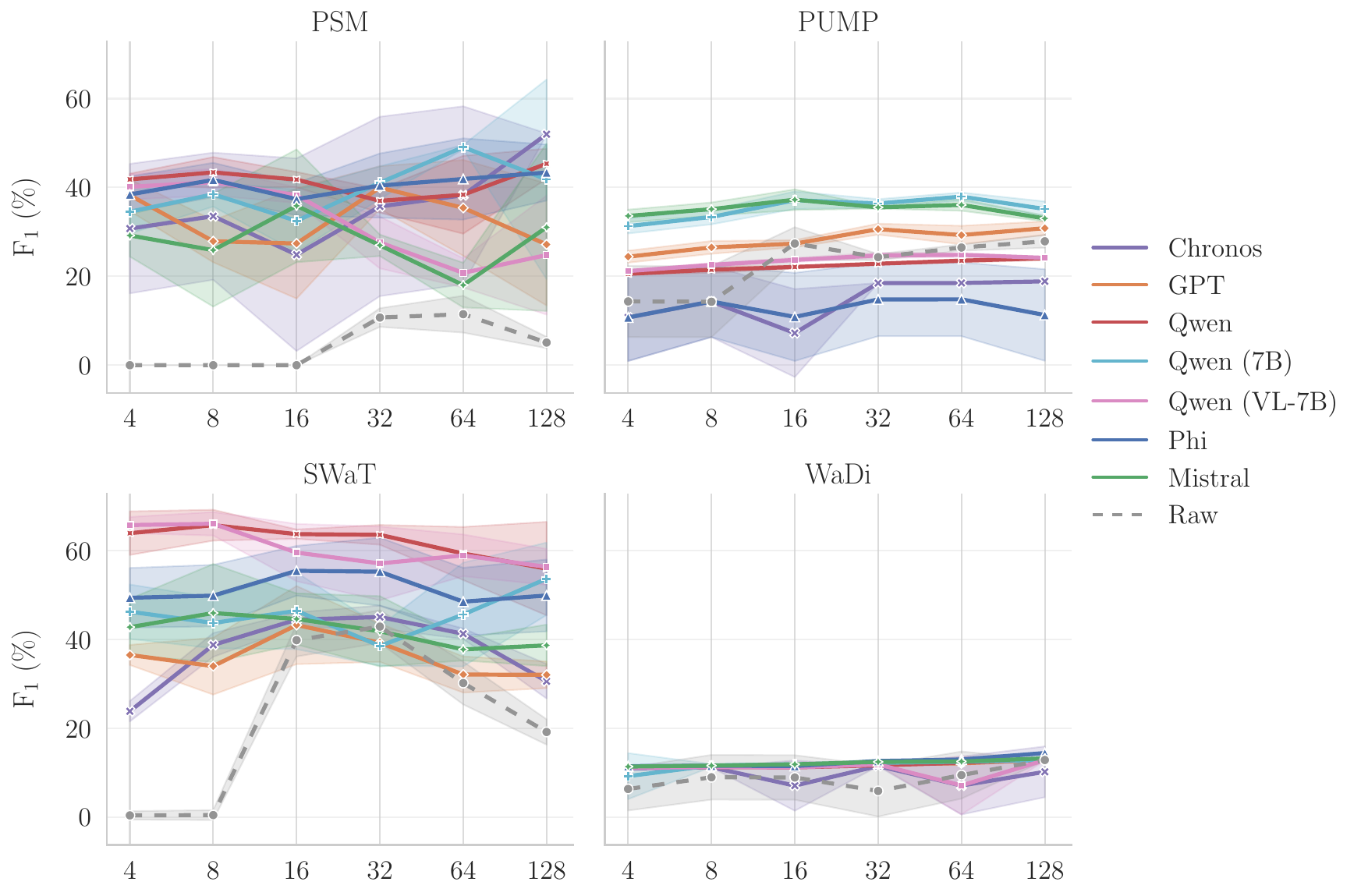}
    \caption{F$_1$ Padding}
    \label{fig:f1pad}
\end{figure*}

\begin{figure*}[ht]
    \centering
    \includegraphics[width=1\linewidth]{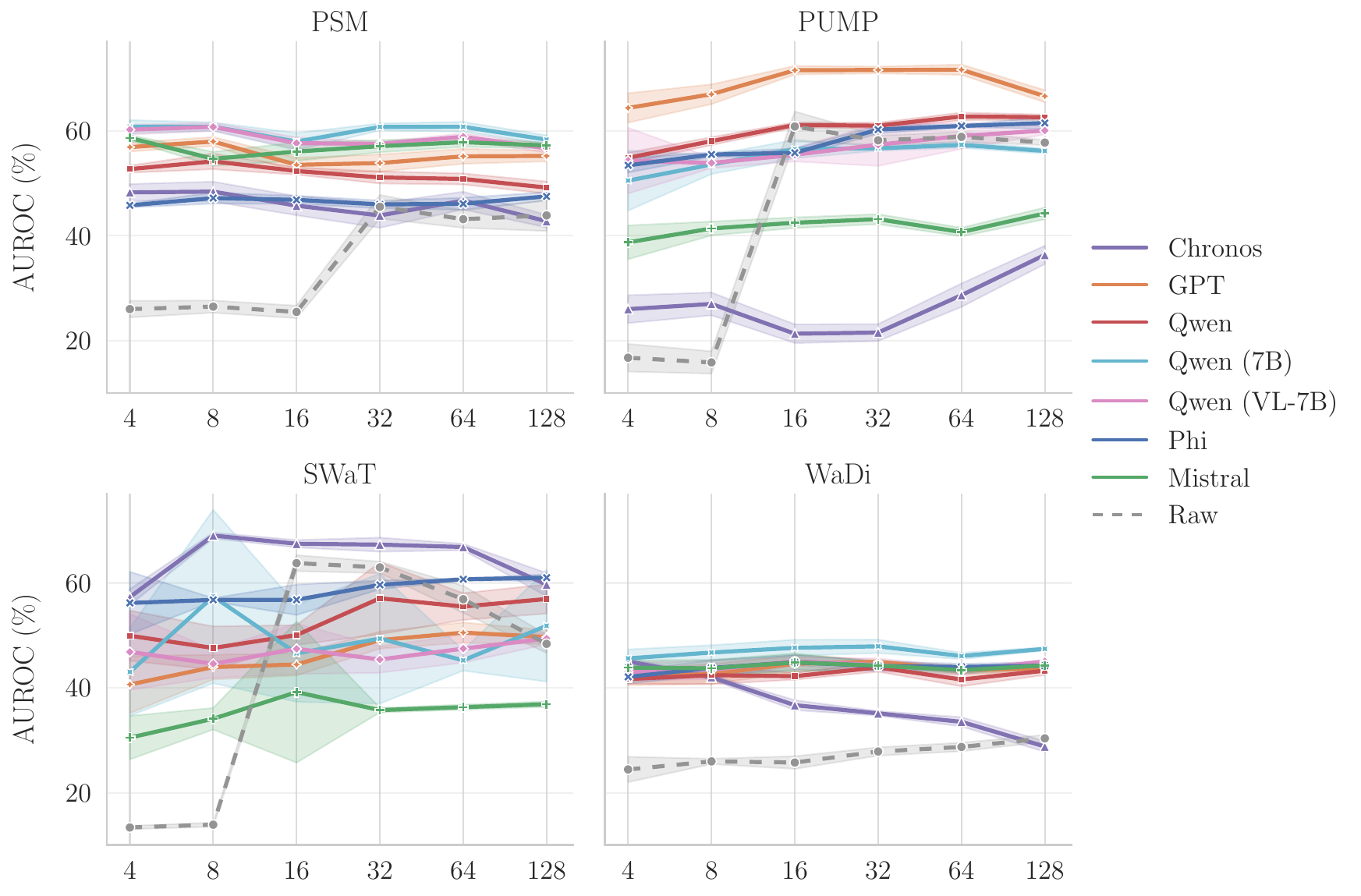}
    \caption{AUROC Closest Vocabulary}
    \label{fig:auroccv}
\end{figure*}

\begin{figure*}[ht]
    \centering
    \includegraphics[width=1\linewidth]{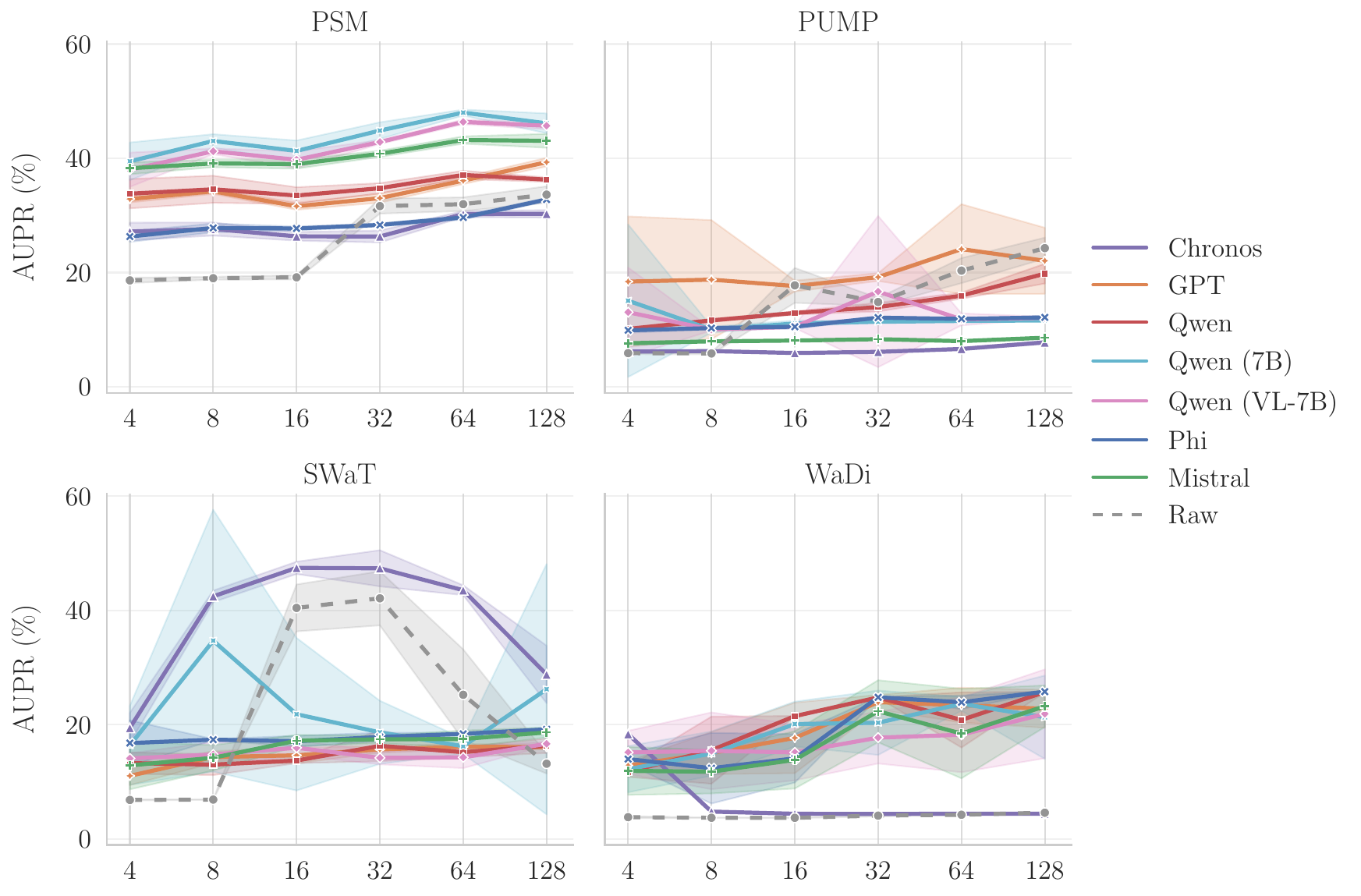}
    \caption{AUPR Closest Vocabulary}
    \label{fig:auprcv}
\end{figure*}

\begin{figure*}[ht]
    \centering
    \includegraphics[width=1\linewidth]{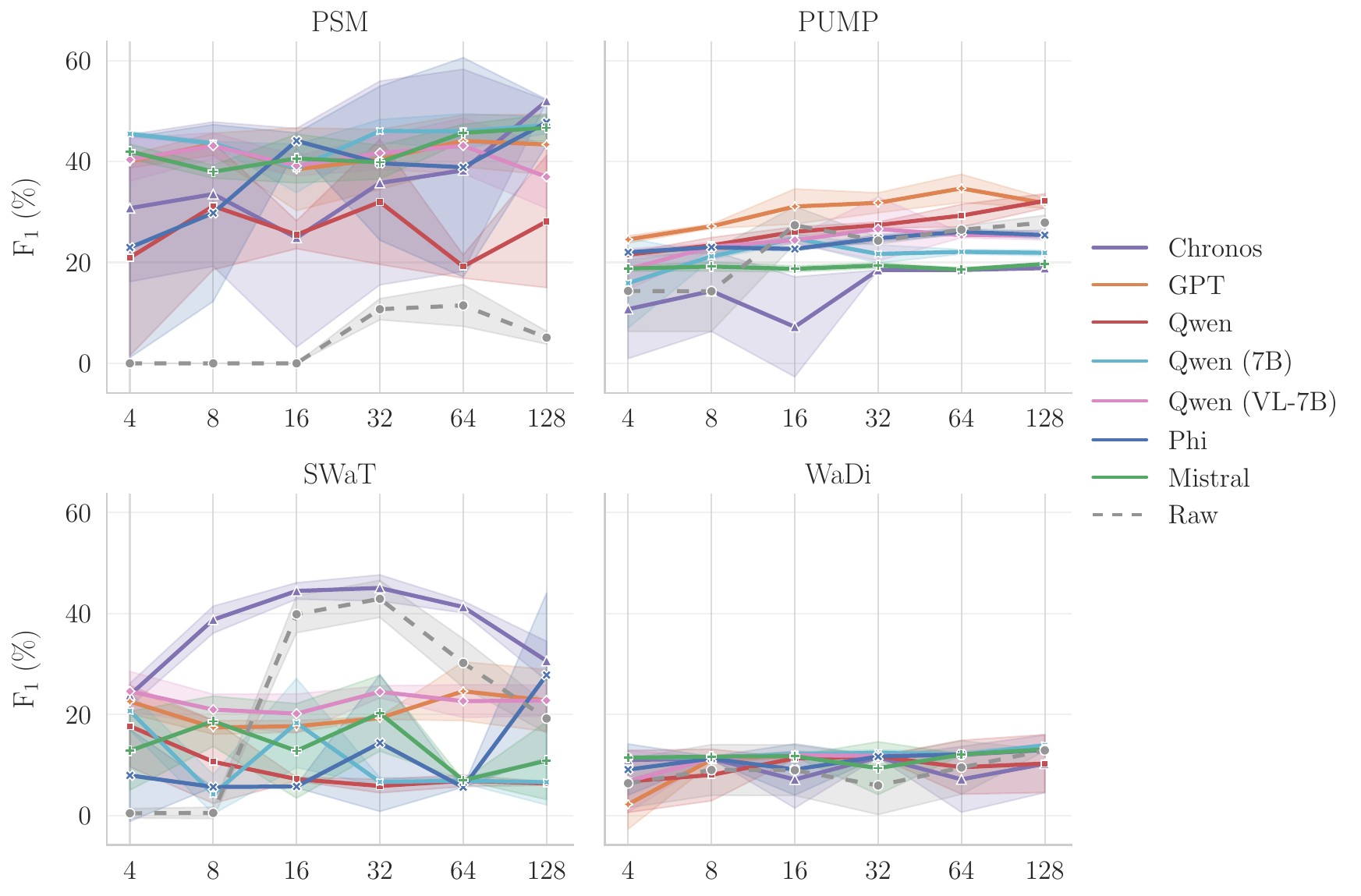}
    \caption{F$_1$ Closest Vocabulary}
    \label{fig:f1cv}
\end{figure*}

\begin{figure*}[ht]
    \centering
    \includegraphics[width=1\linewidth]{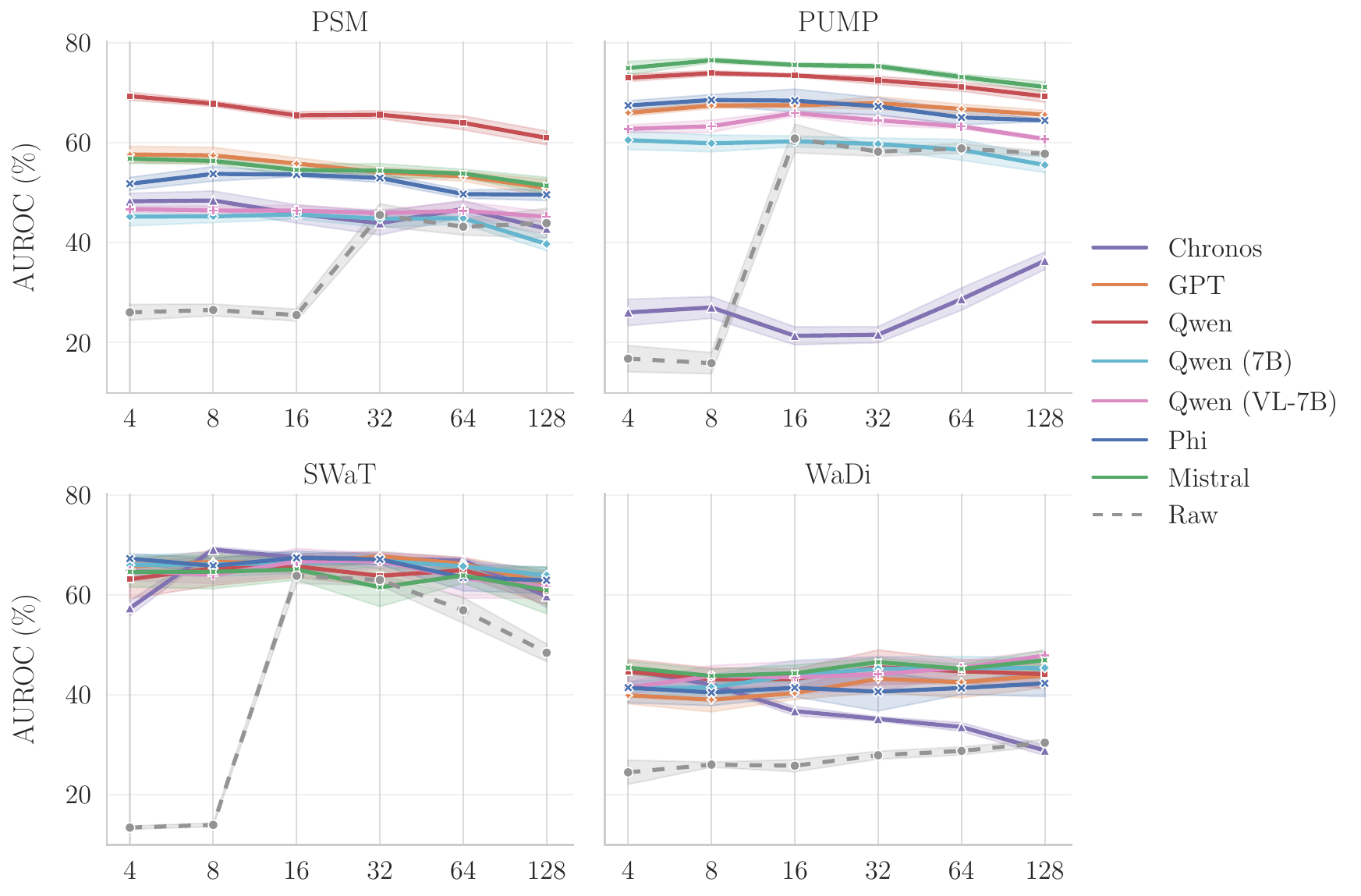}
    \caption{AUROC Interpolation}
    \label{fig:aurocavg}
\end{figure*}

\begin{figure*}[ht]
    \centering
    \includegraphics[width=1\linewidth]{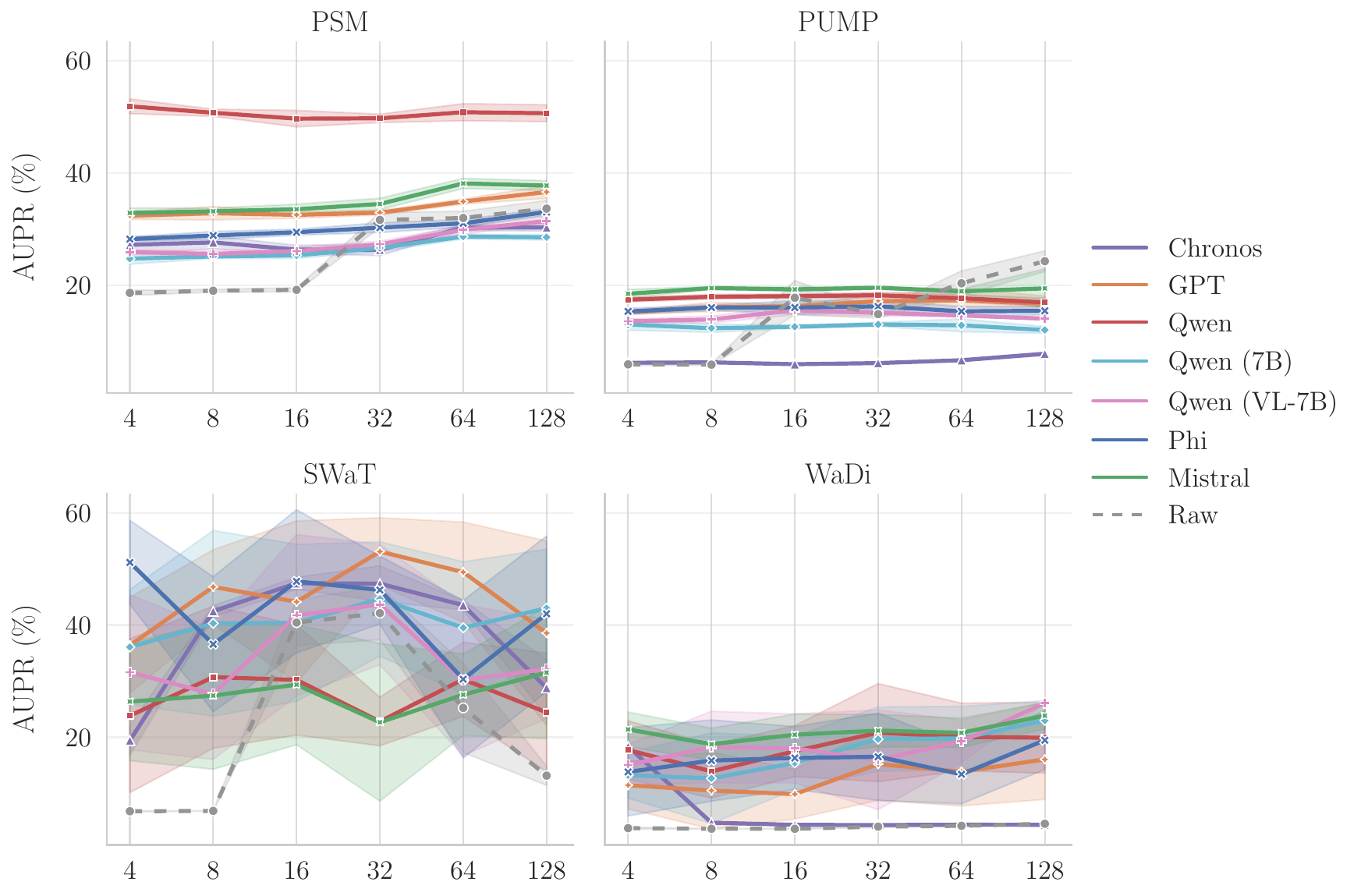}
    \caption{AUPR Interpolation}
    \label{fig:aupravg}
\end{figure*}

\begin{figure*}[ht]
    \centering
    \includegraphics[width=1\linewidth]{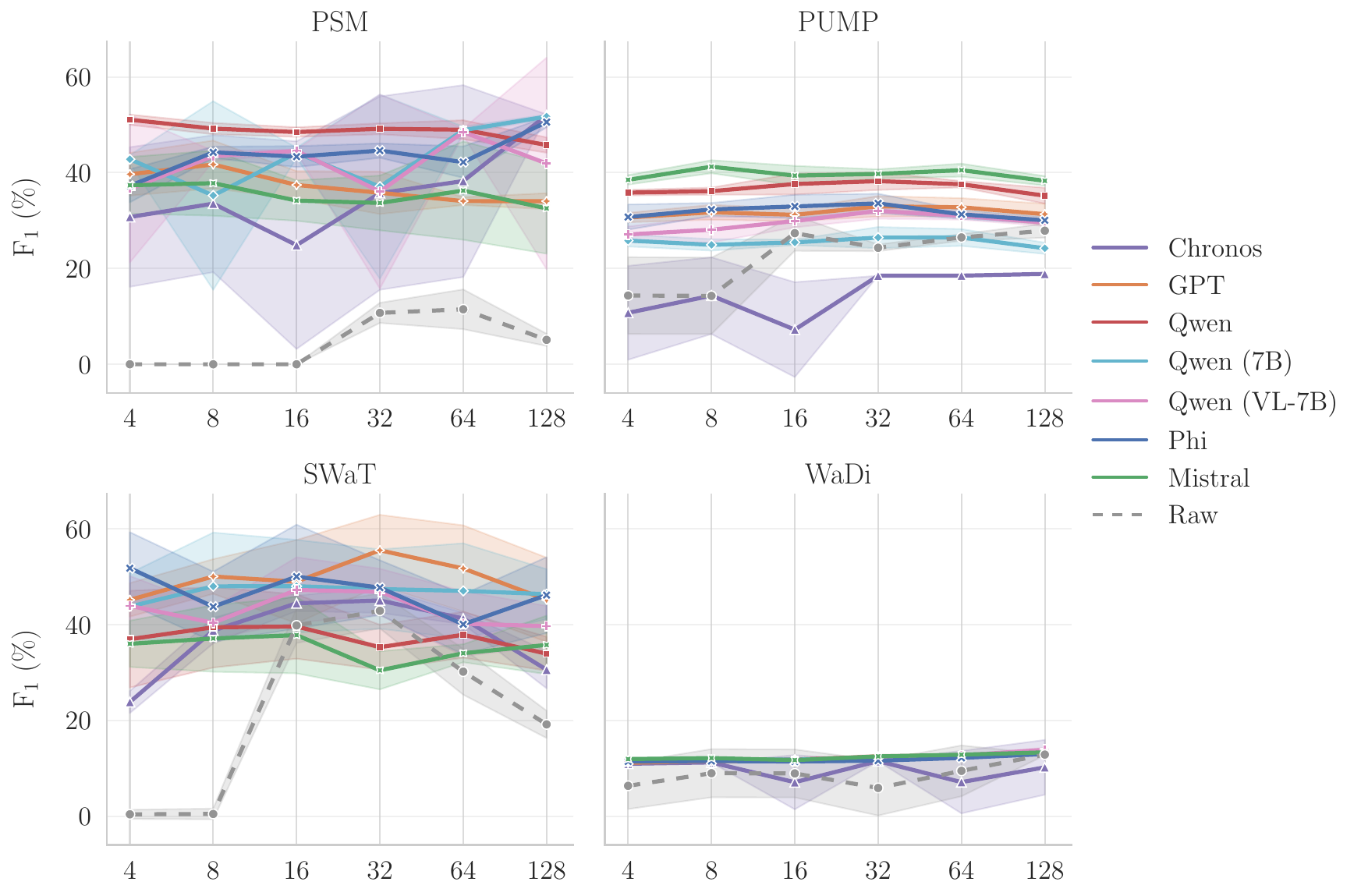}
    \caption{F$_1$ Interpolation}
    \label{fig:f1avg}
\end{figure*}

We provide a granular view of the results of all models, per window sizes (4, 8, 16, 32, 64, and 128), datasets (PSM, PUMP, SWaT and WaDi), and metrics (AUROC, AUPR and F$_1$) from Figure~\ref{fig:aurocpad} to Figure~\ref{fig:f1avg}.

Stable AUROC results can be observed throughout all window sizes and datasets.
AUPR is less stable throughout the datasets, with especially high or increased standard deviation on SWaT for most models.
F$_1$ acts as the most unstable metric, but it can be attributed to its inherent dependence on a threshold.
Applying the Isolation Forest directly on the data (\textit{Raw}) consistently performs worse on small window sizes; most of the time, results are bumped up by going from a window size of 8 to a window size of 16.

\begin{table*}[t]
    \centering
    \caption{Results of Isolation Forest on \textit{Raw} tokens, and on Chronos features (which required no tokenisation).
    The \textit{Control} path of the Padding strategy and the \textit{Raw} path have the same properties, with 0-variance dimensions ignored, and thus the same results. 
    }
    \label{tab:results-raw}
    \resizebox{1\linewidth}{!}{
\renewcommand{\arraystretch}{1.2}%
\begin{tabular}{cccccccccccccc}
\toprule
\multirow[c]{2}{*}{\textbf{Adapter}} & \multirow[c]{2}{*}{\textbf{Window Size}} & \multicolumn{3}{c}{\textbf{PSM}} & \multicolumn{3}{c}{\textbf{PUMP}} & \multicolumn{3}{c}{\textbf{SWaT}} & \multicolumn{3}{c}{\textbf{WaDi}} \\
\cline{3-14}
 & & \textbf{AUPR} & \textbf{AUROC} & \textbf{F$_1$} & \textbf{AUPR} & \textbf{AUROC} & \textbf{F$_1$} & \textbf{AUPR} & \textbf{AUROC} & \textbf{F$_1$} & \textbf{AUPR} & \textbf{AUROC} & \textbf{F$_1$} \\
\midrule\midrule
\multirow[c]{8}{*}{\textit{Raw}} & 4 & 18.63\stdv{0.47} & 26.03\stdv{1.58} & 0.00\stdv{0.00} & 5.90\stdv{0.18} & 16.76\stdv{2.64} & 14.35\stdv{8.02} & 6.82\stdv{0.09} & 13.40\stdv{0.32} & 0.43\stdv{0.96} & 3.81\stdv{0.18} & 24.45\stdv{2.42} & 6.38\stdv{4.85} \\
 & 8 & 19.01\stdv{0.24} & 26.50\stdv{1.17} & 0.00\stdv{0.00} & 5.84\stdv{0.13} & 15.86\stdv{2.13} & 14.28\stdv{7.98} & 6.88\stdv{0.05} & 13.93\stdv{0.39} & 0.50\stdv{1.12} & 3.70\stdv{0.03} & 25.99\stdv{0.52} & 9.01\stdv{5.04} \\
 & 16 & 19.17\stdv{0.37} & 25.49\stdv{1.19} & 0.00\stdv{0.00} & 17.76\stdv{3.06} & 60.83\stdv{2.87} & 27.38\stdv{3.69} & 40.44\stdv{4.11} & 63.77\stdv{1.52} & 39.87\stdv{3.67} & 3.69\stdv{0.06} & 25.77\stdv{1.21} & 8.97\stdv{5.01} \\
 & 32 & 31.64\stdv{1.32} & 45.53\stdv{2.25} & 10.73\stdv{2.10} & 14.84\stdv{0.72} & 58.19\stdv{0.92} & 24.32\stdv{0.70} & 42.12\stdv{4.74} & 62.96\stdv{1.11} & 42.92\stdv{3.68} & 4.08\stdv{0.14} & 27.87\stdv{0.80} & 5.93\stdv{5.76} \\
 & 64 & 31.97\stdv{1.20} & 43.17\stdv{1.65} & 11.48\stdv{4.15} & 20.34\stdv{2.19} & 58.85\stdv{0.95} & 26.49\stdv{0.70} & 25.23\stdv{7.95} & 56.89\stdv{2.58} & 30.22\stdv{4.81} & 4.23\stdv{0.10} & 28.74\stdv{0.81} & 9.50\stdv{5.31} \\
 & 128 & 33.65\stdv{1.46} & 43.89\stdv{2.99} & 5.10\stdv{1.28} & 24.28\stdv{1.84} & 57.76\stdv{0.50} & 27.89\stdv{1.46} & 13.17\stdv{1.73} & 48.39\stdv{1.76} & 19.21\stdv{2.86} & 4.59\stdv{0.08} & 30.38\stdv{0.66} & 12.90\stdv{0.15} \\
\bottomrule
\end{tabular}
}
\end{table*}

\begin{table*}[t]
    \centering
    \caption{Results of Isolation Forest on the Closest Vocabulary tokens (\textit{Control}).}
    \label{tab:results-control-cv}
    \resizebox{1\linewidth}{!}{
\renewcommand{\arraystretch}{1.2}%

\begin{tabular}{cccccccccccccc}
\toprule
 \multirow[c]{2}{*}{\textbf{Model}} & \multirow[c]{2}{*}{\textbf{\shortstack{Window\\Size}}} & \multicolumn{3}{c}{\textbf{PSM}} & \multicolumn{3}{c}{\textbf{PUMP}} & \multicolumn{3}{c}{\textbf{SWaT}} & \multicolumn{3}{c}{\textbf{WaDi}} \\
\cline{3-14}
 & & \textbf{AUPR} & \textbf{AUROC} & \textbf{F$_1$} & \textbf{AUPR} & \textbf{AUROC} & \textbf{F$_1$} & \textbf{AUPR} & \textbf{AUROC} & \textbf{F$_1$} & \textbf{AUPR} & \textbf{AUROC} & \textbf{F$_1$} \\
\midrule\midrule
\multirow[c]{8}{*}{GPT} & 4 & 33.32\stdv{0.91} & 60.38\stdv{0.66} & 45.18\stdv{1.77} & 13.59\stdv{1.02} & 64.02\stdv{1.76} & 22.49\stdv{4.88} & 10.04\stdv{1.62} & 32.15\stdv{2.98} & 20.83\stdv{2.94} & 12.25\stdv{8.15} & 41.68\stdv{1.68} & 9.03\stdv{5.05} \\
 & 8 & 34.65\stdv{0.67} & 58.35\stdv{1.24} & 44.22\stdv{2.29} & 18.60\stdv{10.30} & 65.87\stdv{1.59} & 24.75\stdv{2.63} & 11.73\stdv{2.04} & 42.58\stdv{5.04} & 18.86\stdv{1.97} & 14.60\stdv{3.59} & 42.36\stdv{1.68} & 11.63\stdv{0.35} \\
 & 16 & 33.37\stdv{0.65} & 56.59\stdv{1.30} & 45.04\stdv{1.13} & 16.70\stdv{0.80} & 70.37\stdv{0.68} & 27.79\stdv{1.13} & 21.90\stdv{13.48} & 53.24\stdv{8.58} & 11.98\stdv{8.21} & 18.81\stdv{5.59} & 44.69\stdv{1.65} & 12.02\stdv{0.50} \\
 & 32 & 33.17\stdv{1.60} & 54.03\stdv{1.23} & 41.75\stdv{3.72} & 29.65\stdv{10.34} & 72.64\stdv{1.36} & 36.04\stdv{1.70} & 14.94\stdv{0.86} & 46.94\stdv{2.28} & 18.49\stdv{1.61} & 23.54\stdv{3.15} & 45.88\stdv{0.92} & 12.15\stdv{0.39} \\
 & 64 & 35.49\stdv{1.03} & 54.51\stdv{0.63} & 40.73\stdv{1.63} & 22.00\stdv{1.80} & 72.69\stdv{0.95} & 37.77\stdv{4.44} & 15.45\stdv{0.50} & 47.84\stdv{2.35} & 26.44\stdv{3.51} & 24.87\stdv{0.05} & 44.75\stdv{0.60} & 12.20\stdv{0.20} \\
 & 128 & 37.55\stdv{1.19} & 54.53\stdv{1.48} & 44.62\stdv{3.72} & 25.16\stdv{7.17} & 73.34\stdv{1.24} & 34.14\stdv{2.74} & 15.77\stdv{0.47} & 49.03\stdv{1.87} & 23.79\stdv{4.63} & 24.49\stdv{3.03} & 44.76\stdv{0.72} & 13.13\stdv{0.33} \\
\bottomrule
\end{tabular}
}
\end{table*}

\begin{table*}[t]
    \centering
    \caption{Results of Isolation Forest on the Interpolated tokens (\textit{Control}).}
    \label{tab:results-control-pool}
    \resizebox{1\linewidth}{!}{
\renewcommand{\arraystretch}{1.2}%

\begin{tabular}{cccccccccccccc}
\toprule
 \multirow[c]{2}{*}{\textbf{Model}} & \multirow[c]{2}{*}{\textbf{\shortstack{Window\\Size}}} & \multicolumn{3}{c}{\textbf{PSM}} & \multicolumn{3}{c}{\textbf{PUMP}} & \multicolumn{3}{c}{\textbf{SWaT}} & \multicolumn{3}{c}{\textbf{WaDi}} \\
\cline{3-14}
 & & \textbf{AUPR} & \textbf{AUROC} & \textbf{F$_1$} & \textbf{AUPR} & \textbf{AUROC} & \textbf{F$_1$} & \textbf{AUPR} & \textbf{AUROC} & \textbf{F$_1$} & \textbf{AUPR} & \textbf{AUROC} & \textbf{F$_1$} \\
\midrule\midrule
\multirow[c]{8}{*}{GPT} & 4 & 20.84\stdv{0.99} & 33.81\stdv{2.50} & 0.00\stdv{0.00} & 7.63\stdv{0.84} & 39.36\stdv{6.42} & 17.94\stdv{0.09} & 6.75\stdv{0.15} & 10.44\stdv{0.95} & 8.79\stdv{11.83} & 3.66\stdv{0.08} & 24.65\stdv{1.11} & 4.43\stdv{6.07} \\
 & 8 & 22.17\stdv{1.01} & 36.18\stdv{2.68} & 0.40\stdv{0.89} & 13.36\stdv{2.04} & 57.36\stdv{2.34} & 21.96\stdv{1.53} & 10.96\stdv{1.57} & 44.60\stdv{4.46} & 19.46\stdv{4.43} & 3.76\stdv{0.01} & 27.22\stdv{0.47} & 11.25\stdv{0.06} \\
 & 16 & 27.33\stdv{1.50} & 42.40\stdv{1.81} & 5.77\stdv{1.73} & 17.82\stdv{2.98} & 61.13\stdv{2.05} & 27.54\stdv{3.17} & 41.87\stdv{2.22} & 63.81\stdv{0.96} & 40.37\stdv{2.29} & 3.69\stdv{0.06} & 25.69\stdv{1.31} & 8.97\stdv{5.02} \\
 & 32 & 29.86\stdv{0.96} & 44.66\stdv{1.49} & 6.91\stdv{2.21} & 14.87\stdv{1.35} & 57.44\stdv{1.55} & 24.31\stdv{1.02} & 42.42\stdv{8.58} & 62.46\stdv{2.16} & 42.18\stdv{7.82} & 4.12\stdv{0.12} & 28.29\stdv{0.74} & 5.25\stdv{5.83} \\
 & 64 & 31.83\stdv{1.24} & 43.02\stdv{2.22} & 13.04\stdv{5.52} & 20.87\stdv{1.52} & 59.42\stdv{1.45} & 28.30\stdv{1.91} & 28.34\stdv{11.60} & 57.24\stdv{4.18} & 33.88\stdv{8.84} & 4.20\stdv{0.08} & 28.48\stdv{0.51} & 7.12\stdv{6.50} \\
 & 128 & 33.97\stdv{1.15} & 43.72\stdv{1.77} & 11.45\stdv{7.16} & 27.40\stdv{2.62} & 59.76\stdv{0.87} & 30.92\stdv{1.30} & 14.62\stdv{4.68} & 47.62\stdv{5.49} & 22.98\stdv{3.87} & 4.62\stdv{0.08} & 30.40\stdv{0.63} & 12.99\stdv{0.17} \\
\bottomrule
\end{tabular}
}
\end{table*}

\begin{table*}[t]
    \centering
    \caption{Results of Isolation Forest on LLM features, pre-trained, with Padding tokenisation.}
    \label{tab:results-padding}
    \resizebox{1\linewidth}{!}{
\renewcommand{\arraystretch}{1.2}%

\begin{tabular}{cccccccccccccc}
\toprule
 \multirow[c]{2}{*}{\textbf{Model}} & \multirow[c]{2}{*}{\textbf{\shortstack{Window\\Size}}} & \multicolumn{3}{c}{\textbf{PSM}} & \multicolumn{3}{c}{\textbf{PUMP}} & \multicolumn{3}{c}{\textbf{SWaT}} & \multicolumn{3}{c}{\textbf{WaDi}} \\
\cline{3-14}
 & & \textbf{AUPR} & \textbf{AUROC} & \textbf{F$_1$} & \textbf{AUPR} & \textbf{AUROC} & \textbf{F$_1$} & \textbf{AUPR} & \textbf{AUROC} & \textbf{F$_1$} & \textbf{AUPR} & \textbf{AUROC} & \textbf{F$_1$} \\
\midrule\midrule
\multirow[c]{8}{*}{GPT} & 4 & 33.93\stdv{0.27} & 56.15\stdv{1.66} & 38.26\stdv{3.59} & 12.14\stdv{0.45} & 60.44\stdv{1.31} & 24.42\stdv{1.35} & 37.16\stdv{8.51} & 64.33\stdv{1.15} & 36.54\stdv{2.26} & 11.64\stdv{3.24} & 37.53\stdv{0.83} & 11.26\stdv{0.09} \\
 & 8 & 33.72\stdv{0.91} & 52.53\stdv{0.98} & 27.90\stdv{4.73} & 13.32\stdv{0.23} & 63.37\stdv{0.68} & 26.51\stdv{1.44} & 22.50\stdv{12.47} & 61.20\stdv{3.30} & 34.02\stdv{6.41} & 10.51\stdv{5.69} & 37.37\stdv{1.91} & 11.38\stdv{0.33} \\
 & 16 & 35.05\stdv{1.86} & 53.20\stdv{1.91} & 27.40\stdv{12.40} & 14.39\stdv{0.41} & 64.49\stdv{0.94} & 27.36\stdv{0.81} & 38.09\stdv{14.58} & 64.83\stdv{2.38} & 43.27\stdv{8.80} & 11.95\stdv{4.66} & 38.96\stdv{1.22} & 11.66\stdv{0.32} \\
 & 32 & 34.85\stdv{1.40} & 54.25\stdv{0.43} & 39.97\stdv{4.84} & 16.11\stdv{1.21} & 66.35\stdv{0.78} & 30.62\stdv{1.24} & 30.21\stdv{13.07} & 62.89\stdv{2.52} & 39.30\stdv{4.32} & 10.46\stdv{4.60} & 38.81\stdv{0.83} & 11.95\stdv{0.17} \\
 & 64 & 37.37\stdv{0.77} & 53.54\stdv{1.07} & 35.42\stdv{10.79} & 14.74\stdv{0.50} & 64.35\stdv{1.04} & 29.29\stdv{2.08} & 18.05\stdv{4.11} & 59.62\stdv{2.84} & 32.14\stdv{4.08} & 15.63\stdv{5.12} & 40.63\stdv{2.71} & 12.14\stdv{0.22} \\
 & 128 & 37.67\stdv{1.66} & 50.94\stdv{0.63} & 27.18\stdv{13.75} & 16.12\stdv{1.69} & 63.72\stdv{0.74} & 30.84\stdv{1.53} & 24.12\stdv{7.15} & 59.27\stdv{3.41} & 32.07\stdv{3.00} & 16.12\stdv{7.99} & 40.86\stdv{2.78} & 12.92\stdv{0.19} \\
\bottomrule
\end{tabular}
}
\end{table*}

\begin{table*}[t]
    \centering
    \caption{Results of Isolation Forest on LLM features, pre-trained, with Closest Vocabulary tokenisation.}
    \label{tab:results-cv}
    \resizebox{1\linewidth}{!}{
\renewcommand{\arraystretch}{1.2}%

\begin{tabular}{cccccccccccccc}
\toprule
 \multirow[c]{2}{*}{\textbf{Model}} & \multirow[c]{2}{*}{\textbf{\shortstack{Window\\Size}}} & \multicolumn{3}{c}{\textbf{PSM}} & \multicolumn{3}{c}{\textbf{PUMP}} & \multicolumn{3}{c}{\textbf{SWaT}} & \multicolumn{3}{c}{\textbf{WaDi}} \\
\cline{3-14}
 & & \textbf{AUPR} & \textbf{AUROC} & \textbf{F$_1$} & \textbf{AUPR} & \textbf{AUROC} & \textbf{F$_1$} & \textbf{AUPR} & \textbf{AUROC} & \textbf{F$_1$} & \textbf{AUPR} & \textbf{AUROC} & \textbf{F$_1$} \\
\midrule\midrule
\multirow[c]{8}{*}{GPT} & 4 & 32.88\stdv{0.60} & 56.90\stdv{0.84} & 40.03\stdv{1.31} & 18.42\stdv{11.41} & 64.37\stdv{2.81} & 24.55\stdv{0.73} & 11.04\stdv{1.62} & 40.64\stdv{5.46} & 22.58\stdv{2.44} & 12.84\stdv{0.76} & 42.10\stdv{1.60} & 2.20\stdv{4.92} \\
 & 8 & 34.22\stdv{0.45} & 57.96\stdv{0.86} & 43.42\stdv{2.20} & 18.76\stdv{10.42} & 66.97\stdv{1.88} & 27.15\stdv{0.43} & 14.32\stdv{1.05} & 43.96\stdv{2.36} & 17.44\stdv{1.37} & 15.03\stdv{3.70} & 42.64\stdv{1.88} & 11.07\stdv{0.81} \\
 & 16 & 31.61\stdv{0.62} & 53.52\stdv{1.35} & 38.46\stdv{8.20} & 17.64\stdv{0.95} & 71.54\stdv{0.82} & 31.08\stdv{3.49} & 14.64\stdv{1.42} & 44.46\stdv{2.10} & 17.67\stdv{1.17} & 17.68\stdv{6.18} & 44.60\stdv{1.67} & 11.92\stdv{0.35} \\
 & 32 & 33.03\stdv{0.80} & 53.86\stdv{1.63} & 40.42\stdv{5.97} & 19.21\stdv{0.68} & 71.59\stdv{0.64} & 31.81\stdv{1.98} & 15.67\stdv{0.58} & 49.11\stdv{1.63} & 19.30\stdv{0.26} & 23.93\stdv{1.28} & 44.95\stdv{0.34} & 11.73\stdv{0.09} \\
 & 64 & 36.10\stdv{0.60} & 55.15\stdv{1.42} & 44.08\stdv{5.15} & 24.11\stdv{7.86} & 71.65\stdv{0.98} & 34.67\stdv{2.80} & 16.13\stdv{0.50} & 50.53\stdv{1.94} & 24.62\stdv{5.83} & 23.43\stdv{3.04} & 43.42\stdv{0.78} & 12.05\stdv{0.15} \\
 & 128 & 39.33\stdv{0.75} & 55.22\stdv{1.04} & 43.36\stdv{6.01} & 22.07\stdv{5.78} & 66.60\stdv{1.14} & 31.76\stdv{0.99} & 16.08\stdv{0.41} & 49.70\stdv{1.35} & 22.78\stdv{6.16} & 22.77\stdv{3.32} & 43.66\stdv{0.55} & 13.05\stdv{0.16} \\
\bottomrule
\end{tabular}
}
\end{table*}

\begin{table*}[t]
    \centering
    \caption{Results of Isolation Forest on LLM features, pre-trained, with Interpolation tokenisation.}
    \label{tab:results-pool}
    \resizebox{1\linewidth}{!}{
\renewcommand{\arraystretch}{1.2}%

\begin{tabular}{cccccccccccccc}
\toprule
 \multirow[c]{2}{*}{\textbf{Model}} & \multirow[c]{2}{*}{\textbf{\shortstack{Window\\Size}}} & \multicolumn{3}{c}{\textbf{PSM}} & \multicolumn{3}{c}{\textbf{PUMP}} & \multicolumn{3}{c}{\textbf{SWaT}} & \multicolumn{3}{c}{\textbf{WaDi}} \\
\cline{3-14}
 & & \textbf{AUPR} & \textbf{AUROC} & \textbf{F$_1$} & \textbf{AUPR} & \textbf{AUROC} & \textbf{F$_1$} & \textbf{AUPR} & \textbf{AUROC} & \textbf{F$_1$} & \textbf{AUPR} & \textbf{AUROC} & \textbf{F$_1$} \\
\midrule\midrule
\multirow[c]{8}{*}{GPT} & 4 & 32.41\stdv{0.69} & 57.64\stdv{1.64} & 39.68\stdv{4.47} & 15.15\stdv{0.30} & 65.96\stdv{0.65} & 30.64\stdv{0.98} & 36.39\stdv{8.54} & 65.69\stdv{1.11} & 45.21\stdv{3.51} & 11.46\stdv{4.23} & 39.86\stdv{1.69} & 11.18\stdv{0.19} \\
 & 8 & 32.82\stdv{1.18} & 57.45\stdv{1.61} & 41.72\stdv{5.00} & 16.12\stdv{0.84} & 67.45\stdv{0.62} & 31.73\stdv{1.55} & 46.80\stdv{6.62} & 66.62\stdv{0.86} & 50.09\stdv{3.62} & 10.51\stdv{6.86} & 38.98\stdv{2.43} & 11.54\stdv{0.29} \\
 & 16 & 32.54\stdv{0.62} & 55.82\stdv{1.14} & 37.43\stdv{2.91} & 16.24\stdv{0.35} & 67.49\stdv{0.40} & 31.20\stdv{1.32} & 44.14\stdv{14.42} & 66.20\stdv{1.97} & 49.00\stdv{8.70} & 9.87\stdv{4.44} & 40.30\stdv{1.26} & 11.56\stdv{0.18} \\
 & 32 & 32.92\stdv{0.41} & 54.08\stdv{0.94} & 35.81\stdv{4.44} & 17.14\stdv{1.08} & 67.92\stdv{1.27} & 32.98\stdv{2.14} & 53.10\stdv{6.01} & 67.71\stdv{0.76} & 55.55\stdv{7.40} & 15.22\stdv{6.46} & 43.17\stdv{2.77} & 12.53\stdv{0.16} \\
 & 64 & 34.90\stdv{0.47} & 53.27\stdv{0.96} & 34.07\stdv{0.82} & 17.29\stdv{2.22} & 66.76\stdv{0.95} & 32.78\stdv{1.91} & 49.43\stdv{8.92} & 66.21\stdv{1.26} & 51.76\stdv{9.00} & 13.87\stdv{6.10} & 42.50\stdv{3.11} & 12.67\stdv{0.35} \\
 & 128 & 36.61\stdv{1.15} & 50.90\stdv{1.69} & 34.05\stdv{1.67} & 16.71\stdv{1.35} & 65.60\stdv{0.87} & 31.34\stdv{2.19} & 38.57\stdv{16.42} & 62.34\stdv{2.28} & 45.23\stdv{8.79} & 16.06\stdv{7.13} & 43.82\stdv{2.45} & 13.37\stdv{0.41} \\
\bottomrule
\end{tabular}
}
\end{table*}

\begin{table*}[t]
    \centering
    \caption{Results random 3-layers MLP model and random Linear projection with (embedding size of 768).}
    \label{tab:results-random-mlp}
    \resizebox{1\linewidth}{!}{
\renewcommand{\arraystretch}{1.2}%

\begin{tabular}{cccccccccccccc}
\toprule
 \multirow[c]{2}{*}{\textbf{Model}} & \multirow[c]{2}{*}{\textbf{\shortstack{Window\\Size}}} & \multicolumn{3}{c}{\textbf{PSM}} & \multicolumn{3}{c}{\textbf{PUMP}} & \multicolumn{3}{c}{\textbf{SWaT}} & \multicolumn{3}{c}{\textbf{WaDi}} \\
\cline{3-14}
 & & \textbf{AUPR} & \textbf{AUROC} & \textbf{F$_1$} & \textbf{AUPR} & \textbf{AUROC} & \textbf{F$_1$} & \textbf{AUPR} & \textbf{AUROC} & \textbf{F$_1$} & \textbf{AUPR} & \textbf{AUROC} & \textbf{F$_1$} \\
\midrule\midrule
\multirow[c]{8}{*}{Linear} & 4 & 22.18\stdv{1.16} & 36.07\stdv{2.22} & 1.95\stdv{2.45} & 9.14\stdv{0.97} & 46.68\stdv{2.60} & 10.23\stdv{9.40} & 7.16\stdv{0.12} & 17.95\stdv{1.30} & 13.06\stdv{11.92} & 3.77\stdv{0.04} & 27.31\stdv{1.01} & 7.04\stdv{5.60} \\
 & 8 & 22.53\stdv{0.60} & 37.73\stdv{1.49} & 2.26\stdv{2.34} & 14.24\stdv{2.00} & 58.57\stdv{2.68} & 25.34\stdv{4.16} & 22.26\stdv{4.34} & 56.25\stdv{2.44} & 27.68\stdv{3.31} & 3.66\stdv{0.05} & 24.96\stdv{1.14} & 4.50\stdv{6.16} \\
 & 16 & 27.12\stdv{0.65} & 42.40\stdv{0.85} & 5.73\stdv{2.05} & 13.03\stdv{0.49} & 55.13\stdv{0.57} & 22.59\stdv{1.06} & 47.51\stdv{3.31} & 65.68\stdv{0.58} & 47.66\stdv{2.30} & 3.70\stdv{0.07} & 26.01\stdv{1.30} & 11.22\stdv{0.03} \\
 & 32 & 31.51\stdv{1.52} & 45.80\stdv{3.95} & 12.36\stdv{5.66} & 11.95\stdv{0.62} & 51.44\stdv{1.69} & 20.53\stdv{0.74} & 58.92\stdv{0.81} & 68.46\stdv{0.47} & 59.20\stdv{1.97} & 4.28\stdv{0.30} & 28.62\stdv{0.62} & 6.47\stdv{3.61} \\
 & 64 & 35.38\stdv{1.32} & 45.99\stdv{1.88} & 15.15\stdv{4.23} & 10.49\stdv{0.69} & 47.83\stdv{1.70} & 17.13\stdv{2.06} & 59.35\stdv{1.82} & 68.20\stdv{0.63} & 63.28\stdv{3.83} & 4.78\stdv{0.33} & 32.69\stdv{0.98} & 4.76\stdv{2.12} \\
 & 128 & 38.02\stdv{2.57} & 46.38\stdv{2.16} & 14.56\stdv{6.50} & 16.01\stdv{2.13} & 50.12\stdv{1.13} & 22.75\stdv{1.95} & 55.83\stdv{2.23} & 65.86\stdv{0.21} & 58.64\stdv{5.84} & 5.13\stdv{0.11} & 35.06\stdv{0.66} & 7.58\stdv{5.08} \\
\bottomrule
\end{tabular}
}
\end{table*}

\begin{table*}[t]
    \centering
    \caption{Results of Isolation Forest on LLM features, randomly initialised, with Padding tokenisation.}
    \label{tab:results-random-padding}
    \resizebox{1\linewidth}{!}{
\renewcommand{\arraystretch}{1.2}%

\begin{tabular}{cccccccccccccc}
\toprule
 \multirow[c]{2}{*}{\textbf{Model}} & \multirow[c]{2}{*}{\textbf{\shortstack{Window\\Size}}} & \multicolumn{3}{c}{\textbf{PSM}} & \multicolumn{3}{c}{\textbf{PUMP}} & \multicolumn{3}{c}{\textbf{SWaT}} & \multicolumn{3}{c}{\textbf{WaDi}} \\
\cline{3-14}
 & & \textbf{AUPR} & \textbf{AUROC} & \textbf{F$_1$} & \textbf{AUPR} & \textbf{AUROC} & \textbf{F$_1$} & \textbf{AUPR} & \textbf{AUROC} & \textbf{F$_1$} & \textbf{AUPR} & \textbf{AUROC} & \textbf{F$_1$} \\
\midrule\midrule
\multirow[c]{8}{*}{GPT} & 4 & 32.95\stdv{1.22} & 57.44\stdv{1.80} & 38.27\stdv{5.03} & 18.42\stdv{0.46} & 74.88\stdv{0.76} & 38.92\stdv{1.29} & 34.47\stdv{8.91} & 65.79\stdv{1.80} & 38.29\stdv{3.90} & 16.55\stdv{0.40} & 41.37\stdv{1.01} & 11.55\stdv{0.43} \\
 & 8 & 34.33\stdv{0.95} & 58.02\stdv{1.08} & 41.70\stdv{5.20} & 19.47\stdv{0.34} & 76.99\stdv{0.45} & 40.80\stdv{1.50} & 17.63\stdv{3.93} & 60.50\stdv{2.55} & 30.11\stdv{2.50} & 17.64\stdv{2.32} & 42.84\stdv{1.21} & 11.91\stdv{0.17} \\
 & 16 & 35.23\stdv{1.82} & 55.55\stdv{0.61} & 34.54\stdv{6.59} & 19.75\stdv{0.18} & 77.05\stdv{0.32} & 40.58\stdv{1.87} & 18.69\stdv{3.51} & 61.58\stdv{2.68} & 31.92\stdv{4.57} & 15.98\stdv{4.28} & 43.35\stdv{1.05} & 12.25\stdv{0.37} \\
 & 32 & 35.36\stdv{1.15} & 54.04\stdv{0.45} & 30.18\stdv{1.52} & 20.09\stdv{0.35} & 76.72\stdv{0.58} & 42.22\stdv{1.11} & 30.66\stdv{14.25} & 63.94\stdv{3.27} & 36.35\stdv{7.92} & 18.28\stdv{2.91} & 44.77\stdv{0.88} & 12.66\stdv{0.31} \\
 & 64 & 37.58\stdv{1.40} & 52.75\stdv{0.84} & 30.10\stdv{6.15} & 19.13\stdv{0.45} & 74.86\stdv{0.76} & 40.74\stdv{1.11} & 26.53\stdv{12.91} & 62.40\stdv{3.41} & 33.70\stdv{7.38} & 22.22\stdv{2.90} & 46.10\stdv{1.54} & 12.99\stdv{0.28} \\
 & 128 & 38.29\stdv{2.07} & 50.69\stdv{1.73} & 25.17\stdv{2.66} & 18.80\stdv{0.18} & 73.97\stdv{0.28} & 38.48\stdv{0.62} & 17.40\stdv{2.29} & 58.15\stdv{1.51} & 31.55\stdv{2.63} & 24.09\stdv{4.39} & 46.92\stdv{0.73} & 13.58\stdv{0.27} \\
\bottomrule
\end{tabular}
}
\end{table*}

\begin{table*}[t]
    \centering
    \caption{Results of Isolation Forest on LLM features, randomly initialised, with Closest Vocabulary tokenisation.}
    \label{tab:results-random-cv}
    \resizebox{1\linewidth}{!}{
\renewcommand{\arraystretch}{1.2}%

\begin{tabular}{cccccccccccccc}
\toprule
 \multirow[c]{2}{*}{\textbf{Model}} & \multirow[c]{2}{*}{\textbf{\shortstack{Window\\Size}}} & \multicolumn{3}{c}{\textbf{PSM}} & \multicolumn{3}{c}{\textbf{PUMP}} & \multicolumn{3}{c}{\textbf{SWaT}} & \multicolumn{3}{c}{\textbf{WaDi}} \\
\cline{3-14}
 & & \textbf{AUPR} & \textbf{AUROC} & \textbf{F$_1$} & \textbf{AUPR} & \textbf{AUROC} & \textbf{F$_1$} & \textbf{AUPR} & \textbf{AUROC} & \textbf{F$_1$} & \textbf{AUPR} & \textbf{AUROC} & \textbf{F$_1$} \\
\midrule\midrule
\multirow[c]{8}{*}{GPT} & 4 & 30.53\stdv{2.59} & 53.35\stdv{2.89} & 39.19\stdv{4.85} & 9.52\stdv{1.75} & 49.58\stdv{9.81} & 22.27\stdv{2.50} & 12.73\stdv{3.29} & 42.56\stdv{11.03} & 18.05\stdv{5.01} & 9.10\stdv{4.18} & 42.32\stdv{3.89} & 11.39\stdv{0.35} \\
 & 8 & 31.46\stdv{2.89} & 51.02\stdv{5.31} & 32.91\stdv{10.05} & 12.58\stdv{2.26} & 61.43\stdv{6.35} & 26.20\stdv{5.40} & 13.73\stdv{1.80} & 48.18\stdv{6.49} & 6.58\stdv{2.07} & 11.89\stdv{6.45} & 44.39\stdv{1.24} & 11.24\stdv{0.98} \\
 & 16 & 30.44\stdv{0.87} & 51.57\stdv{1.56} & 35.87\stdv{4.82} & 12.22\stdv{3.17} & 58.33\stdv{7.25} & 25.86\stdv{6.19} & 13.55\stdv{2.63} & 49.45\stdv{6.07} & 12.29\stdv{7.65} & 22.57\stdv{3.24} & 44.18\stdv{1.60} & 11.55\stdv{0.50} \\
 & 32 & 30.22\stdv{2.80} & 51.03\stdv{3.87} & 43.02\stdv{4.61} & 10.54\stdv{1.77} & 53.27\stdv{7.56} & 22.22\stdv{3.28} & 25.43\stdv{24.03} & 56.79\stdv{11.12} & 3.53\stdv{2.38} & 22.09\stdv{3.77} & 44.87\stdv{2.16} & 12.11\stdv{0.55} \\
 & 64 & 34.94\stdv{4.14} & 51.78\stdv{2.92} & 43.06\stdv{6.66} & 11.72\stdv{2.37} & 57.75\stdv{6.17} & 24.25\stdv{4.18} & 26.01\stdv{21.81} & 58.94\stdv{8.98} & 14.76\stdv{17.49} & 24.80\stdv{0.09} & 43.76\stdv{1.10} & 12.11\stdv{0.15} \\
 & 128 & 41.51\stdv{3.08} & 55.83\stdv{2.19} & 44.22\stdv{4.44} & 12.97\stdv{3.95} & 59.55\stdv{10.55} & 28.23\stdv{10.78} & 17.63\stdv{1.31} & 54.73\stdv{8.31} & 14.36\stdv{14.28} & 25.74\stdv{0.30} & 44.86\stdv{1.81} & 10.51\stdv{5.90} \\
\bottomrule
\end{tabular}
}
\end{table*}

\begin{table*}[t]
    \centering
    \caption{Results of Isolation Forest on LLM features, randomly initialised, with Interpolation tokenisation.}
    \label{tab:results-random-pool}
    \resizebox{1\linewidth}{!}{
\renewcommand{\arraystretch}{1.2}%

\begin{tabular}{cccccccccccccc}
\toprule
 \multirow[c]{2}{*}{\textbf{Model}} & \multirow[c]{2}{*}{\textbf{\shortstack{Window\\Size}}} & \multicolumn{3}{c}{\textbf{PSM}} & \multicolumn{3}{c}{\textbf{PUMP}} & \multicolumn{3}{c}{\textbf{SWaT}} & \multicolumn{3}{c}{\textbf{WaDi}} \\
\cline{3-14}
 & & \textbf{AUPR} & \textbf{AUROC} & \textbf{F$_1$} & \textbf{AUPR} & \textbf{AUROC} & \textbf{F$_1$} & \textbf{AUPR} & \textbf{AUROC} & \textbf{F$_1$} & \textbf{AUPR} & \textbf{AUROC} & \textbf{F$_1$} \\
\midrule\midrule
\multirow[c]{8}{*}{GPT} & 4 & 34.64\stdv{0.63} & 60.26\stdv{1.54} & 44.23\stdv{3.25} & 17.78\stdv{1.40} & 73.21\stdv{3.40} & 36.86\stdv{2.89} & 48.45\stdv{5.62} & 68.18\stdv{0.95} & 48.31\stdv{4.69} & 13.40\stdv{2.90} & 40.60\stdv{1.45} & 11.70\stdv{0.42} \\
 & 8 & 34.88\stdv{1.52} & 58.75\stdv{1.34} & 42.95\stdv{2.76} & 19.17\stdv{1.64} & 75.42\stdv{3.86} & 39.33\stdv{3.96} & 44.93\stdv{9.10} & 67.64\stdv{0.85} & 47.71\stdv{6.05} & 14.45\stdv{2.39} & 42.04\stdv{1.38} & 11.85\stdv{0.31} \\
 & 16 & 34.28\stdv{1.26} & 57.51\stdv{0.90} & 40.89\stdv{3.72} & 19.29\stdv{1.09} & 74.84\stdv{3.21} & 39.13\stdv{3.66} & 46.24\stdv{13.83} & 67.37\stdv{1.99} & 48.22\stdv{12.17} & 16.33\stdv{3.62} & 43.40\stdv{1.50} & 11.97\stdv{0.30} \\
 & 32 & 35.88\stdv{1.72} & 56.53\stdv{2.03} & 37.08\stdv{5.26} & 19.11\stdv{1.33} & 73.88\stdv{3.54} & 38.94\stdv{3.43} & 55.69\stdv{3.88} & 69.31\stdv{0.71} & 58.58\stdv{7.57} & 13.92\stdv{4.30} & 44.44\stdv{1.93} & 12.51\stdv{0.14} \\
 & 64 & 37.63\stdv{2.06} & 54.28\stdv{2.15} & 34.93\stdv{2.97} & 18.85\stdv{1.37} & 72.54\stdv{3.62} & 39.37\stdv{4.32} & 46.59\stdv{5.49} & 66.95\stdv{0.71} & 47.91\stdv{5.72} & 19.71\stdv{3.10} & 43.82\stdv{1.72} & 12.59\stdv{0.25} \\
 & 128 & 39.48\stdv{1.51} & 53.69\stdv{0.76} & 35.38\stdv{1.95} & 18.14\stdv{1.29} & 70.54\stdv{2.81} & 37.00\stdv{3.32} & 38.06\stdv{9.78} & 62.96\stdv{1.59} & 42.70\stdv{7.89} & 22.62\stdv{2.98} & 45.07\stdv{1.53} & 13.63\stdv{0.35} \\
\bottomrule
\end{tabular}
}
\end{table*}

\subsection{Wilcoxon Test: Overview}

For each pair of LLMs, let $d_i = B_i - A_i$ ($i \in \llbracket  1,\ldots,N\rrbracket$) denote the paired difference in AUROC for the $i$-th configuration, where $A_i$ and $B_i$ are the AUROC scores, across datasets and window sizes, of the two LLMs being compared.
Zero differences are excluded from the analysis.
The absolute differences $|d_i|$ are ranked in ascending order, yielding ranks $\rho_i$.
The positive and negative rank sums are then defined as:
$$
\omega^+ = \sum_{i:d_i>0} \rho_i,
\qquad
\omega^- = \sum_{i:d_i<0} \rho_i.
$$

The Wilcoxon signed-rank test statistic is $
\omega = \min(\omega^+,\omega^-).$
We compute the signed rank-biserial effect size as:
$$
r =
\operatorname{sign}\left(\,\widetilde{d}\;\right)
\cdot\left(
1-\frac{\omega}{\omega_{\max}}
\right)
=
\operatorname{sign}\left(\,\widetilde{d}\;\right)
\cdot\left(
1-\frac{2\omega}{N(N+1)/2}
\right),
\qquad\omega_{\max} = \frac{N(N+1)}{4},
$$

where $\widetilde{d}$ denotes the median paired difference. The magnitude $|r|$ therefore reflects the degree to which the signed ranks are concentrated in one direction, while its sign indicates the direction of the difference. In our pairwise comparison matrix, a positive $r$ indicates that the LLM represented by the row achieves higher AUROC than the LLM represented by the column, whereas a negative $r$ indicates the opposite. Since all pairwise comparisons are performed over the same $N=24$ configurations, the magnitudes of $r$ are directly comparable across model pairs.

For each pairwise comparison, the Wilcoxon signed-rank test produces a $p$-value corresponding to the probability, under the null hypothesis, of observing a difference at least as extreme as the one obtained.
Since multiple pairwise Wilcoxon tests are conducted, considering each test independently at the $0.05$ significance level would inflate the probability of making at least one false rejection of the null hypothesis. We therefore apply the Holm procedure to the set of $p$-values obtained from the pairwise tests.
The resulting adjusted $p$-values, denoted by $p_{\mathrm{adj}}$, account for these multiple comparisons while controlling the group-wise error rate.
We consider a pairwise difference statistically significant when $p_{\mathrm{adj}} < 0.05$.

\section{Closest Vocabulary Overview}

\begin{figure*}[ht]
    \centering
    \includegraphics[width=1\linewidth]{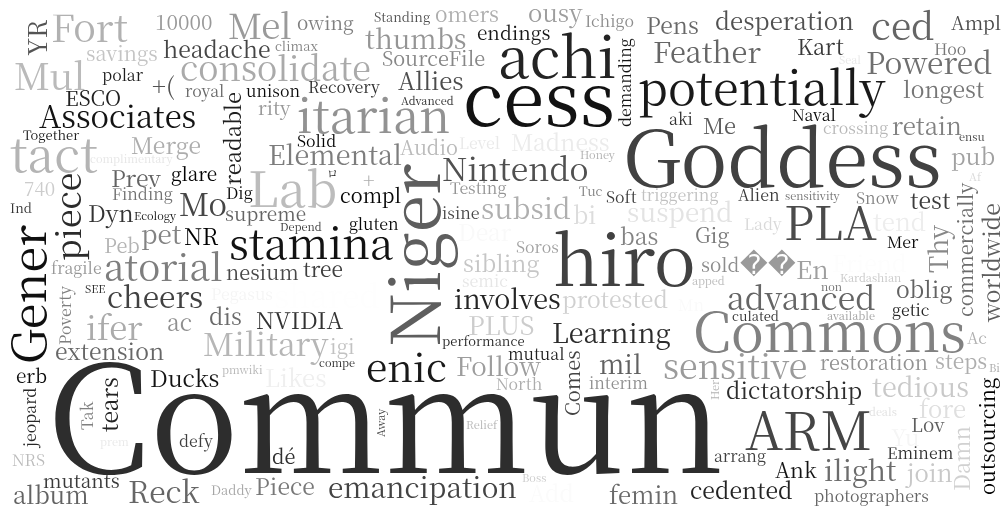}
    \caption{WaDi time-steps translated into GPT2 Vocabulary.}
    \label{fig:vocabgpt2}
\end{figure*}

\begin{figure*}[ht]
    \centering
    \includegraphics[width=1\linewidth]{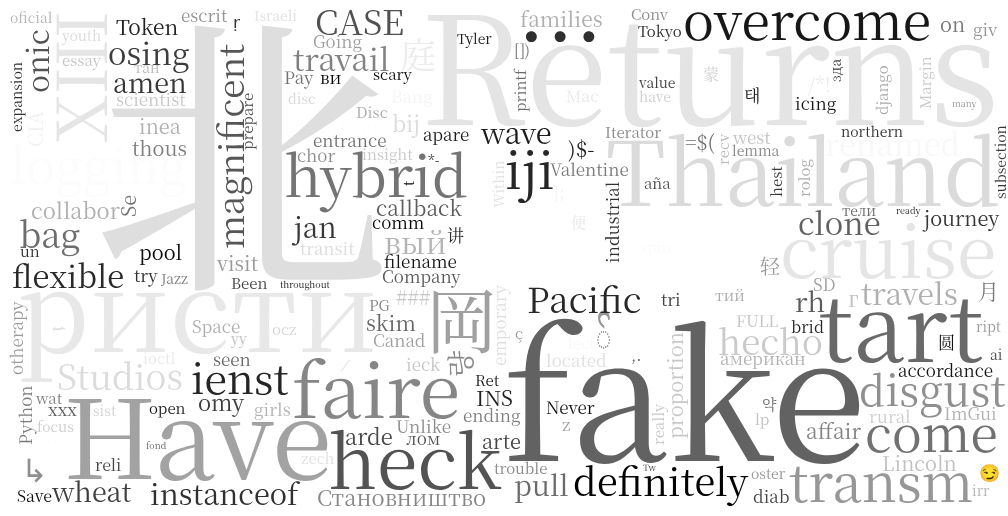}
    \caption{PUMP time-steps translated into Mistral-7B Vocabulary.}
    \label{fig:vocabmistral}
\end{figure*}

\begin{figure*}[ht]
    \centering
    \includegraphics[width=1\linewidth]{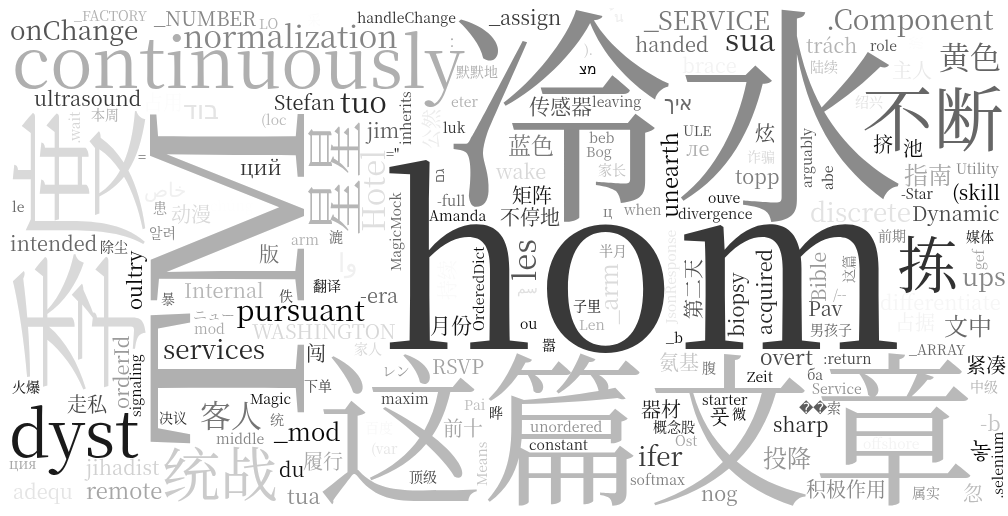}
    \caption{PSM time-steps translated into Qwen2 (0.5B) Vocabulary.}
    \label{fig:vocabqwen}
\end{figure*}

Out of interest, we visualised the resulting tokens when converting the time-series to tokens from a vocabulary (Closest Vocabulary).
Each model has a different vocabulary, and time-steps of a dataset will be converted to various token values and meanings.
We provide some examples from Figure~\ref{fig:vocabgpt2} to Figure~\ref{fig:vocabqwen}.




\end{document}